\documentclass[final,onefignum,onetabnum]{siam_template/siamonline190516}
\usepackage{subcaption}
\usepackage{color}

\usepackage{tikz}
\usetikzlibrary{arrows.meta}

\usepackage{multirow, tabularx, booktabs}
\DeclareUnicodeCharacter{2212}{-}

\usepackage{lipsum}
\usepackage{amsfonts}
\usepackage{graphicx}
\usepackage{epstopdf}
\usepackage{algorithmic}
\ifpdf
  \DeclareGraphicsExtensions{.eps,.pdf,.png,.jpg}
\else
  \DeclareGraphicsExtensions{.eps}
\fi

\usepackage{enumitem}
\setlist[enumerate]{leftmargin=.5in}
\setlist[itemize]{leftmargin=.5in}

\newsiamremark{remark}{Remark}
\newsiamremark{hypothesis}{Hypothesis}
\crefname{hypothesis}{Hypothesis}{Hypotheses}
\newsiamthm{claim}{Claim}

\headers{A Learning Framework for Diffeomorphic Mapping Problems}{Qiguang Chen, Zhiwen Li, and Lok Ming Lui}

\title{QCRegNet draft\thanks{Submitted to the editors DATE.
\funding{This work was funded by the Fog Research Institute under contract no.~FRI-454.}}}

\author{Qiguang Chen\thanks{Department of Mathematics, The Chinese University of Hong Kong, Shatin, Hong Kong  (\email{qgchen@math.cuhk.edu.hk}, \email{zwli@math.cuhk.edu.hk}, \email{lmlui@math.cuhk.edu.hk}).}
\and Zhiwen Li\footnotemark[1]
\and Lok Ming Lui\footnotemark[1]}

\usepackage{amsopn}

\title{A Deep Learning Framework for Diffeomorphic Mapping Problems\\ via Quasi-conformal Geometry applied to Imaging}

\begin{document}

\maketitle

\begin{abstract}
    Many imaging problems can be formulated as mapping problems. A general mapping problem aims to obtain an optimal mapping that minimizes an energy functional subject to the given constraints. Existing methods to solve the mapping problems are often inefficient and can sometimes get trapped in local minima. An extra challenge arises when the optimal mapping is required to be diffeomorphic. In this work, we address the problem by proposing a deep-learning framework based on the Quasiconformal (QC) Teichm\"uller theories. The main strategy is to learn the Beltrami coefficient (BC) that represents a mapping as the latent feature vector in the deep neural network. The BC measures the local geometric distortion under the mapping, with which the interpretability of the deep neural network can be enhanced. Under this framework, the diffeomorphic property of the mapping can be controlled via a simple activation function within the network. The optimal mapping can also be easily regularized by integrating the BC into the loss function. A crucial advantage of the proposed framework is that once the network is successfully trained, the optimized mapping corresponding to each input data information can be obtained in real time. To examine the efficacy of the proposed framework, we apply the method to the diffeomorphic image registration problem. Experimental results outperform other state-of-the-art registration algorithms in both efficiency and accuracy, which demonstrate the effectiveness of our proposed framework to solve the mapping problem.
    
\end{abstract}


\section{Introduction} Mapping problems are ubiquitous in the field of imaging sciences. Generally speaking, a mapping problem is referred to as optimizing a mapping between two corresponding domains that satisfies prescribed conditions. Many imaging tasks can be formulated as mapping problems. For instance, image registration aims to align one image to another. It can be formulated as optimizing a mapping between two images that satisfies certain image intensity matching and regularization constraints \cite{chen2019image,lam2014landmark, lam2015quasi,zhang2019new}. In image segmentation, a suitable mapping can be optimized to deform a prior template shape to extract the important object in an image \cite{chan2018topology,siu2020image,DaopingZhang2021Taci, zhang2021topology_3d}. In addition, mapping problems are also considered in the area of image analysis to study the deformation pattern of shapes in an image \cite{choi2020shape,lui2013shape}. It thus calls for the need of developing an efficient and accurate framework to solve the mapping problem.

Mathematically, a mapping problem can be described as an optimization problem to find an optimal mapping $f:\Omega_1\to \Omega_2$ such that
\begin{equation}
    f = \textbf{argmin}_{g:\Omega_1\to \Omega_2} \mathcal{L}(g) \text{\ \ \  subject to } g\in \mathcal{C},
\end{equation}
\noindent where $\mathcal{L}$ is usually called the energy functional or loss function. By minimizing it, the desired properties of the mapping can be achieved. $\mathcal{C}$ is the constraint space, where the desired mapping should be situated. For example, in image registration, the loss function $\mathcal{L}$ can be a combination of the intensity mismatching error as the fidelity term and a smoothness regularization term. 

Besides, in many practical applications, the mapping is desired to be diffeomorphic. Examples include the diffeomorphic image registration in medical imaging, topology-preserving image segmentation and so on. In this case, the constraint space $\mathcal{C}$ is chosen as the space of diffeomorphisms. This category of mapping problems is called the Diffeomorphism Optimization Problem (DOP). The DOP is generally challenging due to the non-linearity and complexity of the constraint space. Most of the existing methods to solve the problem are based on optimization techniques, the efficiency of which is limited due to the iterative nature of these algorithms. Real-time computation for practical applications is an arduous task. On the other hand, the accuracy is another concern since the methods can sometimes get trapped into local minima. We are therefore motivated to develop a framework to solve the DOP, which is both fast and accurate.

Recently, deep learning have shown tremendous success in various fields. Once successfully trained, deep-learning based models can commonly yield results in real time. Empirically, the stochastic gradient descent (SGD) or related optimization strategies to train the deep neural network can allow the model to avoid undesirable local minima or the so-called premature convergence. Deep-learning techniques have opened up a promising direction to solve the diffeomorphic mapping problem. However, it is generally hard to control the properties of the mapping within a deep learning framework. In particular, enforcing geometric constraints, such as the diffeomorphic property, of the deformation field within the deep neural network is challenging. As a result, topological destruction, namely foldings, can often be seen and violates the constraints of the DOP. 

In this work, we propose a deep-learning framework to solve the DOP based on Quasiconformal (QC) Teichm\"uller theories. To the best of our knowledge, it is the first work to incorporate QC theories into the field of deep learning to enhance the interpretability of the deep neural network. The strategy is to learn the Beltrami coefficient (BC) that represents a mapping as the main latent feature vector in the deep neural network. The BC measures the local geometric distortion of a mapping. By considering the BC as the latent feature, the network structure can be carefully designed according to the prescribed geometric properties, and the neural network can be easily interpreted. Besides, the diffeomorphic constraint of the mapping can be imposed by controlling the magnitude of the BC. As such, the constraint in the DOP can easily be coped with using an appropriate activation function. To incorporate the BC, a task-independent network, called {\it BSNet}, is trained, which reconstructs the quasiconformal mapping from its associated BC. It allows the framework to go back and forth between the mapping and its BC. To train the network efficiently and accurately, the number of parameters in BSNet is reduced using the Fourier approximation of the BC. Under this framework, a flexible loss function combining with the BC can be designed to control the geometric properties. Instead of solving an optimization problem each time, the trained deep neural network can output the optimal mapping in real time given the input data. As such, our proposed framework can solve the mapping problem in real time. The proposed deep-learning based framework for solving the mapping problem is very general, which can be applied to various imaging tasks. In this paper, we examine the efficacy of our proposed method by applying it to an imaging problem, namely the diffeomorphic image registration. A Quasiconformal Registration Network ({called \it QCRegNet}) is developed in an unsupervised manner, which is a novel fast network to obtain diffeomorphic image registration even with large deformations between the image pair. Experiments have been carried out on different data, such as underwater and medical images. Results outperform other state-of-the-art registration models in terms of both accuracy and efficiency. This demonstrates the efficacy of our proposed deep-learning based framework to solve the mapping problem.

The rest of the paper is organized as follows. \Cref{sec:contributions} lists the contributions of this paper. \Cref{sec:related_works} reviews previous works related to our proposed framework. We propose and analyse a learning framework for solving general mapping problems in \cref{sec:proposed_model}. In \cref{sec:QCRegNet}, the diffeomorphic image registration problem is explored and the Quasiconformal Registration Network (QCRegNet) is proposed based on our proposed framework to solve the registration problem. \Cref{sec:implementation} presents the details of the experimental setup. We present the experimental results and discuss insights of the results in \cref{sec:exp_results}. Finally, the conclusion is drawn in \cref{sec:conclusion}.

\section{Contributions}\label{sec:contributions} The main contributions of this paper are outlined as follows. 

\medskip

\begin{enumerate}
    \item A deep-learning based framework using Quasiconformal (QC) theories to solve general diffeomorphism optimization problems is proposed in this paper. To the best of our knowledge, it is the first work to integrate QC theories to control geometric properties within a deep learning framework. A deep neural network with an explainable architecture can be constructed.
    \item The Beltrami Solver Network (BSNet) is proposed and trained in this paper, which computes the corresponding quasiconformal map associated to a prescribed Beltrami coefficient (BC). A mapping can then be represented by the BC, which measures the local geometric distortion under the mapping. The diffeomorphic constraint in the DOP can also be enforced by a simple activation function in the network.
    \item To avoid over-fitting and enhance the training efficiency, the number of parameters in BSNet is carefully reduced using the discrete Fourier transform (DFT) of the BCs. 
    \item The learning based framework together with the BSNet allows us to design a flexible loss function that is incorporated with the BC to control the geometric properties of the output mapping from the network. This allows us to train the network in an unsupervised manner with prescribed geometric constraints to solve the mapping problem, which is especially important when there is not enough labeled data.
    \item Using the proposed framework, a Quasiconformal Registration Network, called QCRegNet, is developed. It is a novel fast network to obtain diffeomorphic image registration results even with large deformations between the image pair. Experimental results obtained by the proposed network show outstanding performance when compared with other state-of-the-art methods.
    
\end{enumerate}

\section{Related works}\label{sec:related_works}
In this work, our goal is to develop a deep learning framework to solve the mapping problems. The main strategy of our proposed framework is to integrate quasiconformal theories into the deep neural network. In this section, we review some existing works on the mapping problems, which are closely related to this paper. Besides, quasiconformal theories are applied in this work. Previous works related to computational quasiconformal geometry are also discussed in this section.

\subsection{Mapping problems} Mapping problems have been widely studied. Many imaging problems, such as image segmentation and registration, can be formulated as mapping problems. Various techniques have been developed to solve the problems. Image segmentation, a process of extracting important object in an image, can be formulated as mapping problems. Numerous models solve the segmentation problem by searching for an optimal mapping that deforms a template binary image to segment the image \cite{chan2018topology,ibrahim2016improved,le2011combined,sdika2010combining,siu2020image,warfield1999nonlinear,yezzi2001variational,DaopingZhang2021Taci, zhang2021topology_3d}. To solve the related mapping problems for image segmentation, different algorithms have been proposed. In \cite{ibrahim2016improved}, a simple gradient descent algorithm together with a multi-level strategy are used to solve the mapping problems. In \cite{chan2018topology,siu2020image,DaopingZhang2021Taci, zhang2021topology_3d}, alternating minimization algorithms are used to solve the optimization problems. Besides, image registration can naturally be regarded as a mapping problem to find an optimal registration map between two corresponding images. Different algorithms have been proposed to solve the associated mapping problems. For example, Horn et al. \cite{horn1981determining} proposed an optical flow method that solves the registration problem. Joshi et al. \cite{joshi2000landmark} proposed the Large Diffeomorphic Distance Metric Mapping (LDDMM) framework to generate diffeomorphic mapping for landmark-matching registration. The corresponding Euler-Lagrange equations are carefully investigated and solved for the solution of the LDDMM problem in \cite{beg2005computing}. Ashburner et al. \cite{AshburnerJohn2007Afdi} developed the DARTEL algorithm, which considers the registration mapping problem as a local optimization problem and solves it using a Levenberg-Marquardt strategy. Vercauteren et al. \cite{vercauteren2009diffeomorphic} proposed the DDemons algorithm for diffeomorphic image registration, which is an iterative gradient descent approach. Burger et al. \cite{burger2013hyperelastic} introduced an image registration model with the incorporation of the hyperelastic regularization term. A numerical scheme is developed to solve the optimization problem, which is based on a nodal-based discretization with a special tetrahedral partitioning. Glocker et al. \cite{glocker2011deformable} proposed an image registration method by computing an optimal registration mapping using a Markov random field formulation. In \cite{lui2015splitting}, Lui et al. introduced an efficient splitting method based on Alternating Direction Method with Multipliers(ADMM) to solve the Diffeomorphism Optimization Problem (DOP). Similarly, Qiu et al. \cite{qiu2020inconsistent} proposed an optimization model to solve the inconsistent surface registration problem. The mapping problem is solved through an iterative algorithm, which involves the projection into the constraint space and the free boundary quasiconformal deformation algorithm. The above methods solve the mapping problem with satisfactory accuracies. Nevertheless, the efficiency of the existing approaches is limited due to the iterative nature of these algorithms.

Learning-based methods, notably those based on deep convolutional neural networks (CNNs), show outstanding performance in various fields, including imaging sciences \cite{chak2018subsampled,unet_bib}, computer vision \cite{guo2016deep,NIPS2015_33ceb07b,meng2021structure,voulodimos2018deep,zhao2019object} and scientific computing \cite{lu2021learning,raissi2019physics}. Attempts has been made to introduce neural network to solve mapping problems. The DLIR framework \cite{de2019deep} makes use of image similarity by minimizing a dissimilarity metric to register images. In \cite{balakrishnan2019voxelmorph}, a velocity field generated by a network, called VoxelMorph, is integrated to obtain the registration mapping. \cite{law2022quasiconformal} uses the feature vectors generated by a truncated classification network to guide large deformation registration.

\subsection{Quasiconformal models} Quasiconformal theories are integrated in proposed framework to solve the mapping problems. Different computational algorithms have been developed to compute the quasiconformal mappings \cite{choi2015flash,lui2013texture,lui2014teichmuller,lui2015splitting,lui2012optimization,zeng2012computing}. For example, an iterative scheme, called the Beltrami holomorphic flow, is proposed in \cite{lui2012optimization} to compute the quasiconformal mapping. Linear Beltrami Solver (LBS) is proposed in \cite{lui2013texture} to obtain the quasiconformal mapping through solving a linear system that discretize the elliptic PDEs. Besides, Quasi-Yamabe flow is developed in \cite{zeng2012computing} to iteratively compute the quasiconformal mapping based on discrete Yamabe flow with an auxiliary metric. Quasiconformal theories have been applied in diverse imaging problems ranging from shape analysis \cite{CHAN2016177,chan2020quasi,choi2020tooth,choi2020shape,lui2013shape,meng2016tempo} and map compression \cite{lui2013texture,lui2010compression} to registration \cite{chen2019image,lam2014genus,lam2015landmark,lam2014landmark,lam2015quasi,lui2014teichmuller,lui2014geometric,lui2010shape,lui2012optimization,qiu2020inconsistent,wen2015landmark,yung2018efficient,zeng2014surface,zhang2019new,zhang2014automatic} and segmentation with a topological prior \cite{chan2018topology,siu2020image,DaopingZhang2021Taci,zhang2021topology_3d}.

\section{Proposed model}\label{sec:proposed_model} In this section, we describe our proposed learning framework for solving general mapping problems in detail. Our ultimate goal is to train a deep neural network that outputs the desired mapping with prescribed constraints optimizing the energy functional of the mapping problem, given the input data.

\subsection{Problem setting} In practical applications, a general mapping problem can be formulated as an optimization problem over the space of mappings $f$ that minimizes a certain energy functional $\mathcal{L}(f;d_1,d_2,...,d_n)$, subject to the constraint that the mapping $f$ is located in a certain constraint space $\mathcal{C}$. Here, ${\bf d} = (d_1, d_2$,..., $d_n)$ is the input data information. For instance, for image registration, the loss function can be defined as: $\mathcal{L}(f;I_1,I_2) = ||I_1 - I_2(f)||^2_2 + \alpha ||\mu(f)||_2^2$, where $\mu(f)$ refers to the Beltrami coefficient of $f$ and ${\bf d}$ is given by the two corresponding images $I_1$ and $I_2$. Minimizing $||\mu(f)||_2^2$ can control the local geometric distortion under the mapping $f$.

Generally speaking, given the input data ${\bf d}$, an optimization problem is solved to obtain the optimal mapping $f$ associated to ${\bf d}$. Ideally, it is desirable to learn a mapping that outputs the optimal mapping $f$ from the input data information ${\bf d}$. Our strategy is to train a deep neural network $N_{\theta}$ with network parameters $\theta = (\theta_1,...,\theta_m)$, which takes ${\bf d}$ as the input and yields a mapping $N_{\theta}({\bf d})$ such that
\begin{equation}
    N_{\theta}({\bf d}) = \text{argmin}_{g:\Omega_1\to \Omega_2} \mathcal{L}(g;{\bf d}) \text{ and } N_{\theta}({\bf d}) \in \mathcal{C}.
\end{equation}

The advantage of this framework is that once the network is successfully trained, the optimized mapping corresponding to each input data information ${\bf d}$ can be obtained in real time. It will be much more applicable in practice than conventional approaches, in which an optimization problem has to be solved for each input data. The main challenge is the control of $N_{\theta}({\bf d})$ to lie in the constraint space $\mathcal{C}$. It requires the incorporation of an effective geometric quantity that measures the geometric properties of the mapping. In this work, we consider the incorporation of the Beltrami coefficient, which is a geometric quantity from QC theories that measures the geometric distortion of a deformation.

\subsection{Overall framework} 

\begin{figure*}[t]
	\centering
	\includegraphics[width=5in]{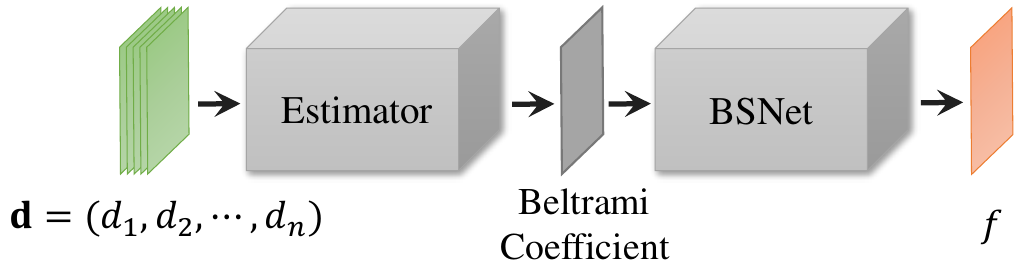}
	\caption{An overview of our proposed learning framework to solve the mapping problem.}
	\label{Overall_framework} 
\end{figure*}

The proposed overall framework to solve the mapping problem via a deep neural network is depicted in \cref{Overall_framework}. We propose to construct a deep neural network, which takes the data information ${\bf d} = (d_1,...,d_n)$ as the input and outputs a mapping. The output mapping should satisfy certain prescribed constraints and it can be controlled by a suitable loss function. The overall network, denoted by $N_{\theta}$, consists of two sub-networks, which are: (1) the {\it estimator} $E_{\phi}$ and (2) the {\it Beltrami Solver Network (BSNet)} $B_{\varphi}$. The estimator estimates the Beltrami coefficient $\mu$ associated to the desired mapping $f$ solving the mapping problem given ${\bf d}$, where $\phi = (\phi_1,...,\phi_k)$ are the network parameters. In other words, $E_{\phi}({\bf d}):\Omega_1\to \mathbb{C}$ is a Beltrami coefficient, which is a complex-valued function on $\Omega_1$. $E_{\phi}({\bf d})$ comprises of the real part $R_{\phi}({\bf d})$ and the imaginary part $Im_{\phi}({\bf d})$. Hence, $E_{\phi}({\bf d})=R_{\phi}({\bf d}) + i Im_{\phi}({\bf d})$. $R_{\phi}({\bf d})$ and $Im_{\phi}({\bf d})$ can be regarded as two images on the domain $\Omega_1$. The main strategy in our proposed framework is to represent a mapping $f = N_{\theta}({\bf d})$ by its Beltrami coefficient $\mu = E_{\phi}({\bf d})$, which measures the geometric distortion under the mapping. As such, the latent representation within the deep neural network is explainable and reflects the geometric properties of the mapping. The incorporation of the Beltrami coefficient into the network is advantageous since the geometric properties of the mapping can be easily controlled by manipulating $\mu$ directly. For instance, if the mapping problem is a diffeomorphism optimization problem, an activation function that rescales the modulus of $\mu$ to $[0, 1)$ can ensure the associated mapping $f$ is diffeomorphic.

On the other hand, in order to incorporate BC into the network, a transformation to recover the corresponding mapping from a given BC is necessary. In this work, we propose to represent this transformation by another deep neural network $B_{\varphi}(\mu)$. In other words, $B_{\varphi}(\mu)$ satisfies the following partial differential equation:
\begin{equation}\label{BSnetBeltramieqt}
    \frac{\partial B_{\varphi}(\mu)}{\partial \bar{z}} = \mu(z) \frac{\partial B_{\varphi}(\mu)}{\partial z},
\end{equation}

\noindent subject to some conditions imposed on $B_{\varphi}({\bf \mu})$. A commonly used condition is the boundary condition on $B_{\varphi}({\bf \mu})$. $B_{\varphi}({\bf \mu})$ can be pre-trained independently and integrated into $N_\theta$. The detailed procedure to train $B_{\varphi}({\bf \mu})$ is explained in \cref{BSNetsubsection}. The parameters $\varphi$ can also be further updated during the training process of $N_{\theta}$, so that the BSNet can be tuned to obtain mappings associated to a particular problem more accurately.

To train a network to obtain the desired mapping, a suitable loss function needs to be designed. Our framework facilitates the incorporation of geometric information in the loss function conveniently. More specifically, the BC that measures the geometric distortion under the associated mapping can be integrated into the loss function. The loss function is usually designed to comprise of the fidelity term and the regularization term. The fidelity term, which depends on the mapping $B_{\varphi}(\mu)$, aims to drive the mapping obtained from the network to fit into the data. The regularization term that depends on the BC $E_{\phi}({\bf d})$ aims to enforce the mapping to satisfy certain prescribed conditions. Hence, the loss function can be formulated as follows:
\begin{equation}
    \mathcal{L}(\phi,\varphi) = E_{data}(B_{\varphi}({\mu})) + E_{reg}(E_{\phi}({\bf d})) = E_1(\phi, \varphi) + E_2(\phi),
\end{equation}
\noindent where $E_{data}$ and $E_{reg}$ denote the data fidelity and regularization terms respectively. $E_1$ and $E_2$ are functions depending on the network parameters, which correspond to $E_{data}$ and $E_{reg}$ respectively. Note that even though $E_{data}$ depends on $B_{\varphi}({\mu})$, $\mu$ is given by $E_{\phi}({\bf d})$. Thus, $E_1$ relies on both $\phi$ and $\varphi$. In case the BSNet is pre-trained and no more updates are considered, the parameter $\varphi$ can be omitted. Thus, the loss function solely depends on the parameter $\phi$.

\subsection{Beltrami Solver Network (BSNet)} \label{BSNetsubsection} The incorporation of the BC into the network facilitates the process to control geometric properties of the mapping. To do so, a network that outputs the quasiconformal mapping associated to an input BC is necessary. In this subsection, we explain in detail how this network, called the {\it Beltrami Solver Network (BSNet)}, is built.

Given an input $\mu$, the BSNet aims to recover the associated mapping $f = B_{\varphi}(\mu)$. $B_{\varphi}(\mu)$ satisfies the Beltrami's equation \cref{BSnetBeltramieqt}. We first discuss some characteristics of the Beltrami's equation, so that a suitable loss function can be designed to train the BSNet.

Let $\mu(z) = \rho(z) + i\tau(z)$ and $f = B_{\varphi}(\mu) = u(z) + iv(z)$ for every $z\in \Omega_1$. Consider the case when $|\mu(z)|<1$ for all $z\in \Omega_1$. Then, the Beltrami's equation \cref{BSnetBeltramieqt} is related to the following energy functional:
\begin{equation}
    E_{LSQC}(u,v;\mu) = \frac{1}{2}\int_{\Omega_1} ||P\nabla u + JP\nabla v||^2 dxdy,
\end{equation}
\noindent where $P(z)=\frac{1}{\sqrt{1-|\mu(z)|^2}}\begin{pmatrix} 1-\rho(z) & -\tau(z) \\
-\tau(z) & 1+\rho(z)
\end{pmatrix}$, $J = \begin{pmatrix} 0 & -1\\1 & 0
\end{pmatrix}$ and $||\cdot||^2 = <\cdot, \cdot>$ is the usual Euclidean norm. $E_{LSQC}$ can be rewritten as:
\begin{equation}
    E_{LSQC}(u,v;\mu) = \left( \frac{1}{2}\int_{\Omega_1}(||A^{1/2}\nabla u||^2 +  ||A^{1/2}\nabla v||^2) dxdy \right) - \mathcal{A}(u,v),
\end{equation}
\noindent where the area distortion term is given by $\mathcal{A}(u,v) =\int_{\Omega_1} (u_y v_x - u_x v_y) dxdy$ and $A = P^t P$. Classical quasiconformal theories assert that $ \frac{1}{2}\int_{\Omega_1}(||A^{1/2}\nabla u||^2 +  ||A^{1/2}\nabla v||^2) dxdy$ is bounded from below by $\mathcal{A}(u,v)$. Hence, $E_{LSQC}(u,v;\mu)\geq 0$ and the global minimum is attained when $u+iv$ satisfies the Beltrami's equation \cref{BSnetBeltramieqt}. The optimization of $E_{LSQC}$ is equivalent to solving the following system of partial differential equations:
\begin{equation}\label{eq:qcpde}
\begin{cases}
-\nabla\cdot(A\nabla u)  =0 \text{ in } \Omega_1 \\
-\nabla\cdot(A\nabla v)  =0 \text{ in } \Omega_1 \\
\partial_{A{\bf n}}u +\nabla v \times {\bf n} =0 \text{ on } \partial\Omega_1\\
\partial_{A{\bf n}}v -\nabla u \times {\bf n} =0 \text{ on } \partial\Omega_1
\end{cases},
\end{equation}
where as before ${\bf n}(z)$ is the outer unit normal vector. In the discrete case, the differential operator $\nabla\cdot(A)$ is discretized using the linear finite element method on the triangular mesh. The system of PDEs can be approximated by a sparse linear system.

\bigskip

\noindent {\it Remark\cite{qiu2019computing}: In case $||\mu||_{\infty}$ is not strictly less than 1, we partition $\Omega_1$ into two domains, namely, $\Omega_+ =\{z\in \Omega_1:|\mu(z)|<1\}$} and $\Omega_- =\{z\in \Omega_1:|\mu(z)|>1\}$. Then, the solution of the Beltrami's equation can also be considered as the critical point of the following energy functional:
\begin{equation}
    E_{generalized}(u,v;\mu) = \frac{1}{2}\int_{\Omega_+} ||P\nabla u + J P\nabla v ||^2dxdy - \frac{1}{2}\int_{\Omega_-} ||P\nabla u + J P\nabla v ||^2dxdy.
\end{equation}
\noindent {\it A system of PDEs similar to equation \cref{eq:qcpde} can also be derived and a sparse linear system can be solved in the discrete case to approximate the solution of the Beltrami's equation. As such, our framework is designed to handle general mapping problems, instead of restricting our attention to homeomorphic mappings only.}

\bigskip

In practical imaging applications, mappings from one rectangle to another rectangle are commonly considered. Suppose $\Omega_1 = [0,1]\times [0,1]$ is a square and assume that $f$ maps $\Omega_1$ onto a rectangle $[0,1]\times [0,h]$. In this case, the boundary conditions on $\partial \Omega_1$ can be imposed as follows:
\begin{equation}\label{eq:qcpdeboudaryrect}
\begin{cases}
u(z) = 0 \text{ if } \text{Re}(z) = 0;\ \ u(z) = 1 \text{ if } \text{Re}(z) = 1\\
v(z) = 0 \text{ if } \text{Im}(z) = 0;\ \ v(z) = 1 \text{ if } \text{Im}(z) = h\\
\end{cases},
\end{equation}
\noindent for some suitable positive $h\in \mathbb{R}$ depending on $\mu$. In fact, we can check that once $u$ satisfies the PDE in equation \cref{BSnetBeltramieqt} subject to the boundary conditions in \cref{eq:qcpdeboudaryrect}, $h$ can be determined by:
\begin{equation}
    h = \frac{1}{2}\int_{\Omega_1}||A^{1/2}\nabla u||^2 dxdy.
\end{equation}

\begin{figure*}[t]
	\centering
	\includegraphics[width=0.6\textwidth]{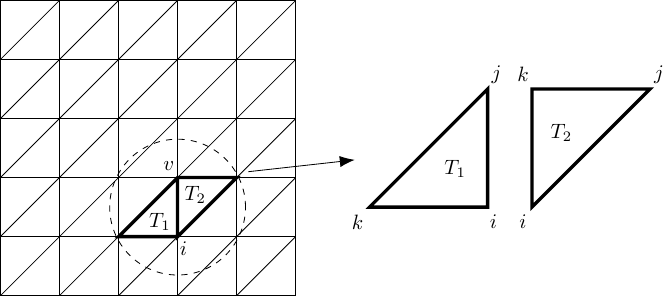}
	\caption{Triangular discretization of $\Omega_1$.}
	\label{triangular_discretization} 
\end{figure*}

To numerically solve the equations, the domain $\Omega_1$ is discretized by a regular triangulation as shown in \cref{triangular_discretization}. We assume the restriction on each triangular face $T$ is linear. Suppose $T=[\vec{v}_i, \vec{v}_j, \vec{v}_k]$ consists of three vertices $\vec{v}_i, \vec{v}_j,$ and $\vec{v}_k$. Let $\vec{w}_I=f(\vec{v}_I)$, $\vec{v}_I = g_I+ih_I$ and $\vec{w}_I = s_I+it_I$, where $I = i,j,k$. The differential of $f$ on $T$ is then a constant $2\times 2$ matrix given by:
\begin{equation}
   \begin{pmatrix}\frac{\partial u}{\partial x}(T) & \frac{\partial u}{\partial y}(T)\\
    \frac{\partial v}{\partial x}(T) & \frac{\partial v}{\partial y}(T)\end{pmatrix} = \begin{pmatrix}
 D^1_{i}(T)s_{i}+D^1_{j}(T)s_{j}+D^1_{k}(T)s_{k} &  D^2_{i}(T)s_{i}+D^2_{j}(T)s_{j}+D^2_{k}(T)s_{k}\\
 D^1_{i}(T)t_{i}+D^1_{j}(T)t_{j}+D^1_{k}(T)t_{k} &  D^2_{i}(T)t_{i}+D^2_{j}(T)t_{j}+D^2_{k}(T)t_{k} 
 \end{pmatrix}
\end{equation}
where
\begin{equation}
    \begin{pmatrix}
    D^1_{i}(T) & D^1_{j}(T) & D^1_{k}(T)\\
    D^2_{i}(T) & D^2_{j}(T) & D^2_{k}(T)
    \end{pmatrix} = \frac{1}{2Area(T)} \begin{pmatrix}
    h_{j}-h_{k} & h_{k}-h_{i} & h_{i}-h_{j}\\
    g_{k}-g_{j} & g_{i}-g_{k} & g_{j}-g_{i}
    \end{pmatrix},
\end{equation}
\noindent where $Area(T)$ refers to the area of the triangular face $T$. As such, the BC $\mu$ is considered as a piecewise constant function on every triangular face. The matrix $A = P^t P$ is a constant $2\times 2$ matrix on every triangular face $T$. More specifically,
\begin{equation}
    A(T) = P(T)^t P(T) = \begin{pmatrix}
        \frac{(\rho(T)-1)^2+\tau(T)^2}{1-\rho(T)^2-\tau(T)^2} &
        -\frac{2\tau(T)}{1-\rho(T)^2-\tau(T)^2} \\
        -\frac{2\tau(T)}{1-\rho(T)^2-\tau(T)^2} &
        \frac{(\rho(T)+1)^2+\tau(T)^2}{1-\rho(T)^2-\tau(T)^2}
    \end{pmatrix}
\end{equation}

For each vertex $\vec{v}_i$, we let $N_i$ be the collection of all neighborhood faces attached to $\vec{v}_i$. The system of PDEs in \cref{BSnetBeltramieqt} can be discretized to obtain a sparse linear system. More specifically, at each vertex $\vec{v}_i$, the following linear equation is satisfied:
\begin{equation}\label{LBSlinear}
    c_{\vec{v}_i} \vec{w}_i +\sum_{\vec{v}\in V_i} c_{\vec{v}} f(\vec{v}) = {\bf 0},
\end{equation}
\noindent where $V_i$ is the set of all adjacent vertices attached to $\vec{v}_i$. The coefficients $c_{\vec{v}_i}$ and $c_{\vec{v}}$ are defined as follows. For $c_{\vec{v}_i}$ associated to the central vertex $\vec{v}_i$, 
\begin{equation} \label{coef:central_vertex}
c_{\vec{v}_i} = \sum_{T\in N_{i}} [ \alpha_1(T) (D_i^1(T))^2 + 2 \alpha_2(T) D_i^1(T) D_i^2(T) + \alpha_3(T)(D_i^2(T))^2 ]
\end{equation}
\noindent As for the rest of the coefficients,
\begin{equation} \label{coef:adjacent_vertex}
\begin{array}{rl}
c_{\vec{v}}&=\alpha_1(T_1) D_{i}^1(T_1) D_{j}^1(T_1) + \alpha_2(T_1) (D_{i}^1(T_1) D_{j}^2(T_1) + D_{j}^1(T_1) D_{i}^2(T_1)) + \alpha_3(T_1) D_{i}^2(T_1) D_{j}^2(T_1) \\
&+\alpha_1(T_2) D_{i}^1(T_2) D_{k}^1(T_2) + \alpha_2(T_2) (D_{i}^1(T_2) D_{k}^2(T_2) + D_{k}^1(T_2) D_{i}^2(T_2)) + \alpha_3(T_2) D_{i}^2(T_2) D_{k}^2(T_2)
\end{array}
\end{equation}

\noindent where $\alpha_1(T) = \frac{(\rho(T)-1)^2+\tau(T)^2}{1-\rho(T)^2-\tau(T)^2}$, $\alpha_2(T)=-\frac{2\tau(T)}{1-\rho(T)^2-\tau(T)^2}$ and $\alpha_3(T)=\frac{(\rho(T)+1)^2+\tau(T)^2}{1-\rho(T)^2-\tau(T)^2}$.

For an $N \times N$ mesh, equation \cref{LBSlinear} can be rewritten as
\begin{equation} \label{linear_system}
\begin{array}{rcl}
    C_{s} \boldsymbol{s} &=& \boldsymbol{0}\\
    C_{t} \boldsymbol{t} &=& \boldsymbol{0}
\end{array}
\end{equation}
\noindent where $\boldsymbol{s}$ and $\boldsymbol{t}$ are vectors whose entries are given by $s_i$'s and $t_i$'s on each vertex respectively. Here, the entries corresponding to boundary vertices are set to be their actual boundary constraints. $C_{s}$ and $C_{t}$ are both $N^{2} \times N^{2}$ sparse matrices, each row of which contains $c_{i}$ and $c_{v}$ for a vertex in the mesh grid. The rows of $C_{s}$ and $C_{t}$, which correspond to the boundary constraints of the two coordinates, are set to 0.

\begin{figure*}[t]
	\centering
	\includegraphics[width=5in]{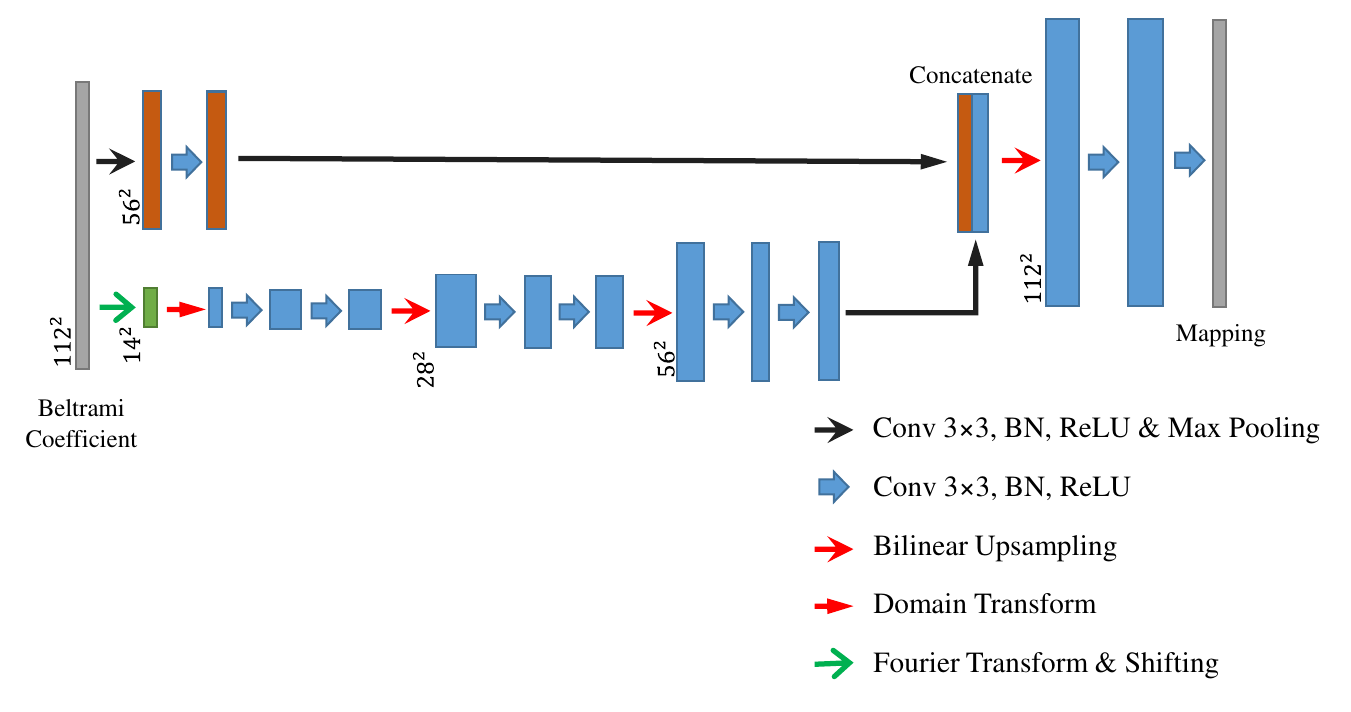}
	\caption{The architecture of the Beltrami Solver Network (BSNet).}
	\label{BSNet} 
\end{figure*}

To build the Beltrami Solver Network, we can train a deep neural network that minimizes the following loss function $\mathcal{L}_{BS}$:
\begin{equation}
\mathcal{L}_{BS} = \frac{1}{2N^{2}}(\|C_{s} \boldsymbol{s}\|_{1} + \|C_{t} \boldsymbol{t}\|_{1})
\end{equation}

Note that each row in $C_{s}$ and $C_{t}$ represents the relationship between a certain pixel and the pixels adjacent to it. In the context of a triangular mesh, a pixel is adjacent to at most six pixels. As such, each row in $C_{s}$ and $C_{t}$ has at most seven nonzero elements. Although the two $N^2 \times N^2$ matrices $C_{s}$ and $C_{t}$ are sparse, both matrices can be rewritten as two dense arrays in the implementation of our method. Thus, the computation of $\mathcal{L}_{BS}$ is memory saving and efficient.

\begin{figure}[t]
    \centering
    \begin{tabularx}{\textwidth}{cXX}
        \multirow{2}{0.33\textwidth}{\includegraphics[width=\linewidth]{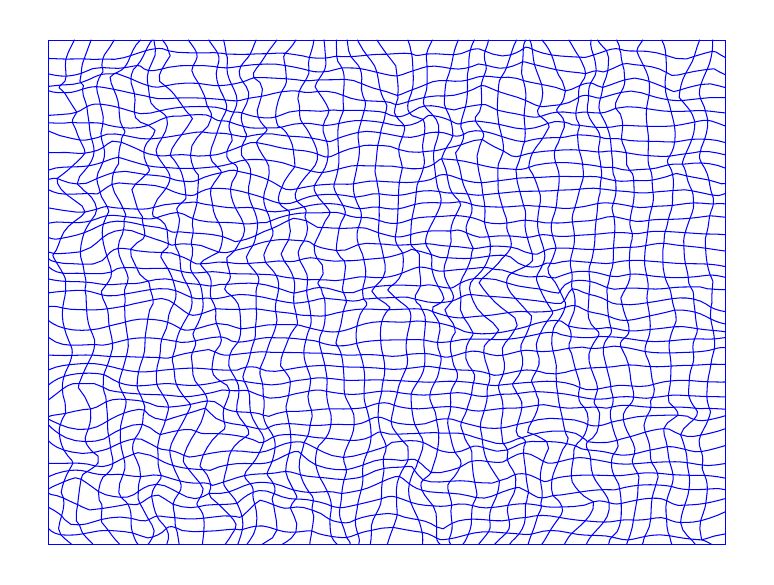}
        \subcaption{Ground truth mapping}}
    &   \includegraphics[width=\linewidth]{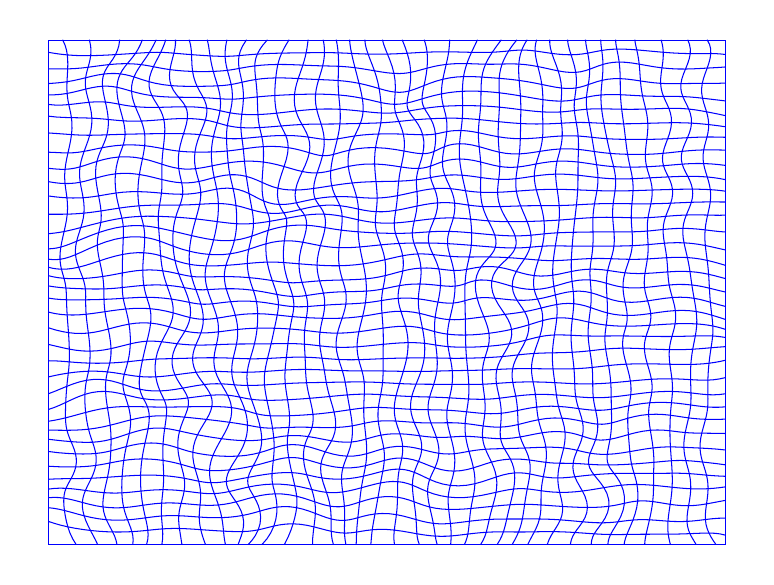}
        \subcaption{2\% of the Fourier coefficient of the Beltrami coefficient}
    &   \includegraphics[width=\linewidth]{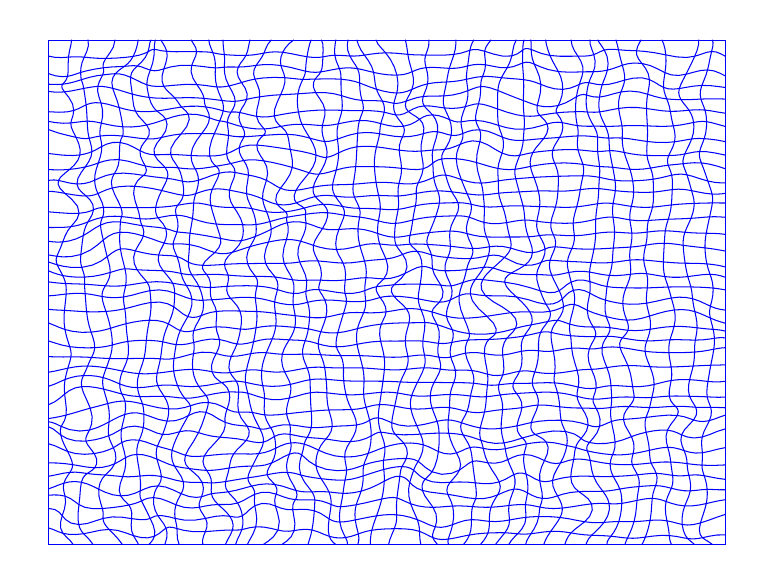}
        \subcaption{8\% of the Fourier coefficient of the Beltrami coefficient}\\
    &   \includegraphics[width=\linewidth]{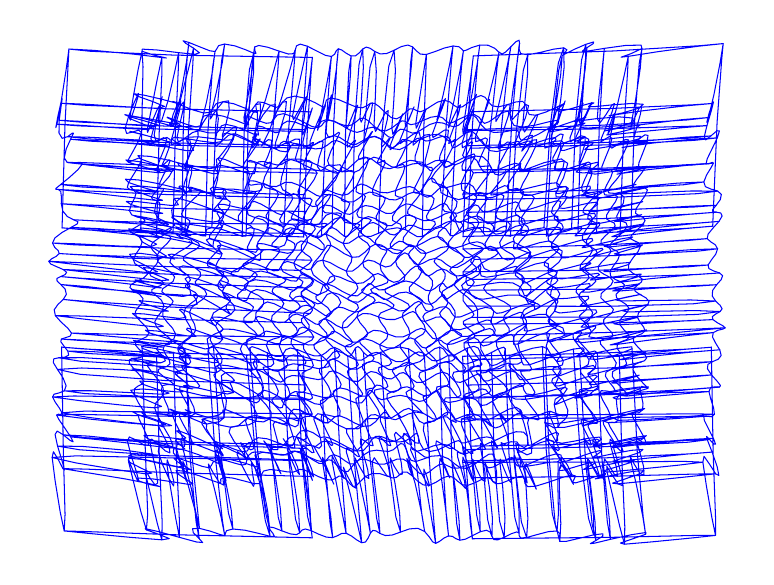}
        \subcaption{8\% of the Fourier coefficient of the coordinate function}
    &   \includegraphics[width=\linewidth]{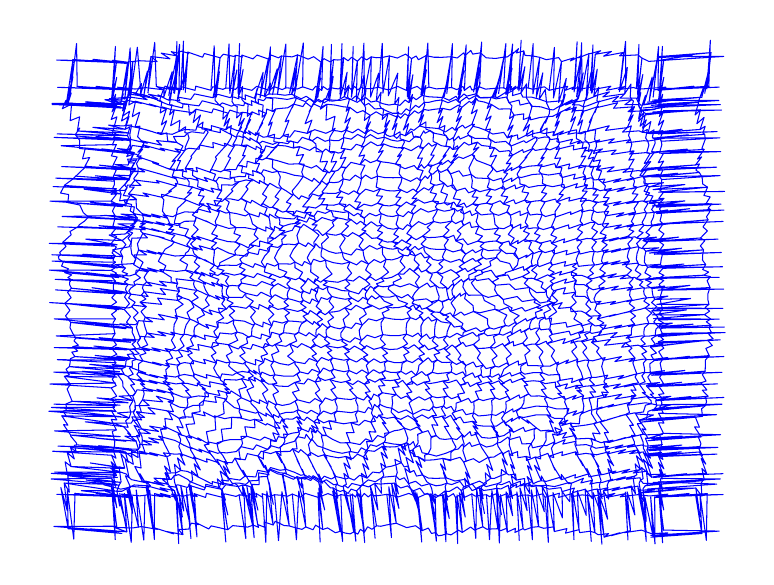}
        \subcaption{80\% of the Fourier coefficient of the coordinate function}
    \end{tabularx}
    \caption{Comparison of Fourier compression of Beltrami coefficient and coordinate function. From these figures, we notice that a small proportion of the Beltrami coefficient can store the major information of mapping, while the coordinate function does not have this property.}
    \label{comparison_Fourier_compression}
\end{figure}

In order to train the BSNet efficiently and accurately, a smaller and simpler deep neural network is preferable. In \cite{lui2013texture}, it is observed that the truncated Fourier decomposition of the BC captures the overall geometry of the corresponding quasiconformal deformation effectively, as shown in \cref{comparison_Fourier_compression}. The low frequency components of the Fourier transform contains most of the geometric distortion information under the associated deformation. Therefore, to simplify the network, we first transform the input BC by discrete Fourier transform (DFT). Given the input BC $\mu = \mu_r + i\mu_i \in M_{N\times N}(\mathbb{C})$, the DFT of $\mu$ is given by $\hat{\mu} =\hat{\mu}_r + i \hat{\mu}_i$, where $\hat{\mu}_r = U\mu_rU$, $\hat{\mu}_i = U\mu_iU$ and $U = (\frac{1}{N} e^{-i \frac{2\pi kl}{N}})_{0\leq k,l \leq N-1}$. We then take the $H\times W$ Fourier coefficients $\hat{\mu}_{tc}$ corresponding to the lowest frequency components as the feature map of the next layer of the network. Note that the DFT has no trainable parameters. The generated low frequency component has only two channels, which is much fewer than the deep features extracted by conventional deep neural network. Nevertheless, it captures the overall geometric distortion information.

Before passing the truncated Fourier coefficients to the next layer, it is noted that common operations, such as convolution and interpolation, in a deep neural network are spatial operators. Directly applying these operators on the Fourier or frequency domain is unreasonable and will hinder the effectiveness of the network to solve the Beltrami's equation. As such, a special layer that transforms the feature in the frequency domain to the spatial domain with a given dimensions $K\times L$ is necessary. For this purpose, we propose the {\it Domain Transform Layer (DTL)}, which takes $\hat{\mu}_{tc} \in M_{H\times W}(\mathbb{C})$ from the center of the shifted Fourier coefficient as the input and yields $\tilde{\mu}^{coarse} \in M_{K\times L}(\mathbb{C})$ as the output. More specifically, consider a complex matrix $\hat{\mu}_{tc} = (\hat{\mu}(m,n))_{0\leq m\leq H-1, 0\leq n\leq W-1}$ of size $H\times W$ containing the selected Fourier coefficients. $\hat{\mu}_{tc}$ is associated to a BC $\tilde{\mu}$ given by:
\begin{equation}
\tilde{\mu}(p,q) = \sum_{m=0}^{H-1} \sum_{n=0}^{W-1} \hat{\mu}_{tc}(m, n)\phi_{m,n}(p,q) 
\end{equation}
\noindent where $\phi_{m,n}(p,q) = e^{2\pi i \frac{(m-H/2)p+(n-W/2) q}{N}}$, with the assumption that $H$ and $W$ are even numbers. $\tilde{\mu}$ is a Fourier approximation of the original BC $\mu$. Nevertheless, $\tilde{\mu}$ corresponds to a discrete deformation map $\tilde{f}$ on a fine $N\times N$ grid. As such, the dimension of $\tilde{\mu}$ is $N\times N$. 

To simplify the network, we construct the DTL that outputs a BC $\tilde{\mu}^{coarse} \in M_{K\times L}(\mathbb{C})$, which corresponds to a coarse deformation map $\tilde{f}^{coarse}$ on a coarse $K\times L$ grid. $\tilde{f}^{coarse}$ can be regarded as a `low-resolution' deformation map of $\tilde{f}$. To obtain $\tilde{\mu}^{coarse}$, we append another layer that imitates the inverse DFT, the process of coarsening $\tilde{f}$ and the computation of the BC $\tilde{\mu}^{coarse}\in M_{K\times L}(\mathbb{C})$ of $\tilde{f}$. Mathematically, this layer can be formulated as
\begin{equation}
    \tilde{\mu}^{coarse} = M\hat{\mu}_{tc}N = (\hat{\mu}_{tc}^T M^T)^T N
\end{equation}
\noindent where $M$ and $N$ are trainable complex matrix of sizes $K\times H$ and $W\times L$ respectively. 
In our experiments, we set $H=W=K=L = 14$. With DTL, the features in frequency domain can be transformed to a proper domain where spatial operations work. We can then perform convolution and upsampling on the features to obtain the mappings.

Nevertheless, the truncated Fourier approximation of $\mu$ causes an information loss for the subtle detail of the deformation map $f$. As such, we introduce a second path in the network to add back local details of the deformation map. As in \cref{BSNet}, the upper path is the second path. In this path, a convolution layer and a down sampling are performed on the input $\mu$, after which there are two more convolution layers. The output features are concatenated to the output from the first path. Note that the number of parameters in this path is small and thus it does not cause much burden to the overall network structure. Experimental results in the following sections show the necessity of this path.

The BSNet is trained in an unsupervised manner. Countless $N \times N$ Beltrami coefficients $\mu_{sqr}$ are generated by stacking pairs of images in the ILSVRC2012 dataset, which are augmented with some data augmentation tricks, such as random crop and flipping. $\mu_{sqr}$ serves as the input of BSNet. The parameters are optimized such that the expected value of the loss function over the training dataset is minimized. 

\begin{figure*}[t]
	\centering
	\includegraphics[width=5in]{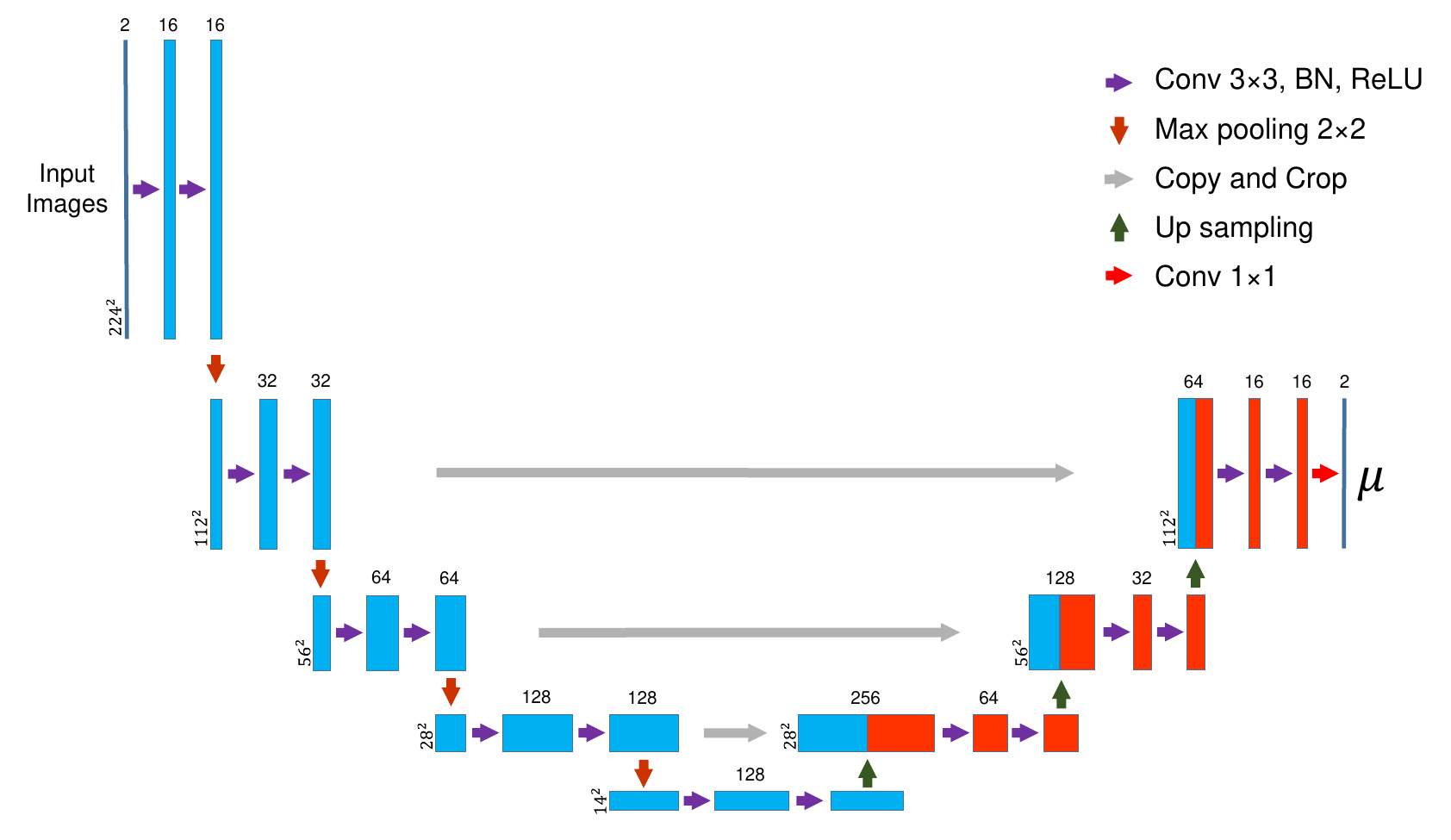}
	\caption{The architecture of the estimator network.}
	\label{Estimator} 
\end{figure*}

\subsection{Estimator Network} The main feature of our proposed framework is that the BC, which measures the local geometric distortions of the mapping, is used to represent the mapping. Incorporating BC into the network allows us to control the geometric distortion of the mapping effectively and have a better geometric understanding of the network. The incorporation of BC into the network is feasible with the BSNet as introduced in the last subsection. In order to learn the mapping $f$ that solves a given mapping problem, the {\it estimator network} $E_{\phi}$ is introduced, where $\phi = (\phi_1,...,\phi_k)$ are the network parameters. The estimator network takes the data information ${\bf d}$ as the input and yields a suitable BC $\mu$ as the output. Once successfully trained, the output $\mu$ corresponds to an associated quasiconformal mapping $f$ that solves the mapping problem with a given ${\bf d}$. The estimator network is designed based on U-Net \cite{unet_bib} with an activation function added to its output. The overall framework is shown in \cref{Estimator}. The estimator network performs convolution and pooling operations to extract the deep features of the input data. The deep features are then up-sampled to yield the BC $\mu$, which can be regarded as a two channel image representing the real and imaginary parts of $\mu$. When diffeomorphic property of the mapping is required, $\mu$ should satisfy the constraint that $||\mu||_{\infty}<1$. It is guaranteed by the following theorem. 

\begin{theorem}\label{bijectivity}
Let $f:M_1\to M_2$ be a piecewise linear mapping between two meshes $M_1$ and $M_2$, which are both simply-connected. In other words, $f$ is defined on every vertices of $M_1$ and linearly interpolated on every triangulation faces. Denote the BC of $f$ by $\mu:M_1\to \mathbb{C}$, which is piecewise constant on each face.  If $|\mu(T)|<1$ for every face $T$, then $f$ is globally bijective ({\it foldover-free globally}).
\end{theorem}

\begin{proof}
We consider the mapping $f|_T := u|_T + i v|_T$ restricted to a triangular face $T$. $f|_T$ is linear. The BC of $f|_T$, $\mu(T)$, is a complex constant. Since $f|_T$ is linear, its partial derivatives are constants. Denote the derivatives $\frac{\partial u|_T}{\partial x}$, $\frac{\partial u|_T}{\partial y}$, $\frac{\partial v|_T}{\partial x}$ and $\frac{\partial v|_T}{\partial y}$ by $a_T$, $b_T$, $c_T$ and $d_T$ respectively. The Jacobian determinant of $f|_T$ is given by
\begin{equation}
\begin{split}
    J(f|_T) = a_T d_T - b_T c_T &= \frac{1}{4} |(a_T+d_T) - i (b_T-c_T)|^2 - \frac{1}{4}|(a_T-d_T) + i(b_T+c_T)|^2\\
    &= \kappa \left(1 - \frac{|(a_T-d_T) + i (b_T+c_T)|^2}{|(a_T+d_T) - i (b_T-c_T)|^2}\right)\\
    &=\kappa(1-|\mu(T)|^2),
\end{split}
\end{equation}
\noindent where $\mu(T) = \frac{(a_T-d_T) + i (b_T+c_T)}{(a_T+d_T) - i (b_T-c_T)}$ and $\kappa = |\frac{\partial f|_T}{\partial z}|^2 >0 $ for a well-defined BC. If $|\mu(T)| <1$ for every face $T$, then $J(f|_T) >0$ and so $f|_T$ is orientation preserving on every face $T$. Thus, $f|_T$ is locally injective foldover-free. By inverse function theorem, $f$ is a local homeomorphism. Besides, $f$ is proper. Recall that for simply-connected $M_1$ and $M_2$, a local homeomorphism $f:M_1\to M_2$ is a global homeomorphism of $M_1$ onto $M_2$ if and only if $f$ is proper. Thus, $f$ is actually a global bijection (foldover-free globally).
\end{proof}

To enforce this constraint, we add an activation function in the last layer. The following activation function $\mathcal{T}$ is used:
\begin{equation}\label{activation}
    \mathcal{T}(\nu)(T) = \left(\frac{e^{|\nu(T)|} - e^{-|\nu(T)|}}{e^{|\nu(T)|} + e^{-|\nu(T)|}}\right) e^{i\text{arg}(\nu(T))}
\end{equation}
\noindent where $\nu$ is the output of the last layer, $\nu(T)$ is a complex number corresponding to triangle $T$. As $|\nu(T)|\geq 0$, we must have $0 \leq |\mathcal{T}(\nu)(T)|<1 $ for all $T$. Thus, $||\mathcal{T}(\nu)||_{\infty}<1$.

Together with other criteria of the mapping problem, an overall loss function can be considered to train the deep neural network. The overall loss function $\mathcal{L}$ usually consists of the regularization and fidelity terms: 
\begin{equation}
    \mathcal{L}(\mu,f) = \mathcal{L}_{reg}(\mu) + \mathcal{L}_{fidelity}(f).
\end{equation} 

\noindent The regularization term depends on the BC $\mu$. For instance, to suppress the size of the BC, the following energy term can be added to $\mathcal{L}_{reg}$:
\begin{equation}
    \mathcal{L}_{BC}= \frac{1}{|\Omega_1|} \sum_{T\in \Omega_1} |\mu(T)|^2.
\end{equation}
\noindent where $|\Omega_1|$ means the number of faces in the discretized domain $\Omega_1$. This term helps to reduce the local geometric distortion under the mapping and control the bijectivity of the mapping. In addition, the smoothness of the mapping can be enhanced by adding the smoothness term to $\mathcal{L}_{reg}$:
\begin{equation}
\mathcal{L}_{smooth}= \frac{1}{|\Omega_1|} \sum_{T\in \Omega_1} |\nabla \mu(T)|^{2},
\end{equation}
\noindent where $\nabla$ refers to the discrete gradient, which can be approximated by finite-difference scheme.

\noindent The fidelity term often depends on the quasiconformal mapping $f$ corresponding to the BC $\mu$. The neural network is then trained by minimizing $\mathcal{L}(\mu,f)$ subject to the constraint that $\mu$ is the BC of $f$ satisfying Equation \cref{linear_system}. The optimization problem can be solved easily, with the embedded BSNet that gives the mapping $f$ from the BC $\mu$. More specifically, $\mathcal{L}$ can be considered as a functional depending on the parameters $\phi$ and $\varphi$ of the estimator network and BSNet respectively. To further simplify the problem, the BSNet is pre-trained and the training of parameters $\varphi$ can be omitted when training parameters $\phi$ of the estimator network. The optimization problem is solved by the RMSprop gradient descent method \cite{rmsprop}. 

\subsection{Analysis of the proposed framework} In this subsection, we analyze the theoretical aspects of our proposed framework.

A useful feature of our proposed framework is that it can control the bijectivity of the mapping returned by the network. It can be achieved simply by truncating the BC feature obtained from the estimator network using a suitable activation function. 

To analyze this, recall that our BSNet aims to output a quasiconformal map $f$ that minimizes $\mathcal{E}_{LSQC}(f)$. In the discrete setting, $f$ is piecewise linear on every triangular face $T\in \mathcal{F}$, where $\mathcal{F}$ is the collection of triangular faces. The integrand in $\mathcal{E}_{LSQC}(f)$ restricted to each face $T$ can be rewritten as:

\begin{equation}
    \mathcal{E}_{BC}(T) = \frac{1}{2}\left(\alpha_1(T) (a_T^2 + c_T^2) + 2\alpha_2(T) (a_Tb_T +c_T d_T) + \alpha_3(T) (b_T^2 + d_T^2)\right) - \mathcal{A}(T),
\end{equation}
\noindent where $a_T$, $b_T$, $c_T$ and $c_T$ denote $\frac{\partial u}{\partial x}(T)$, $\frac{\partial u}{\partial y}(T)$, $\frac{\partial v}{\partial x}(T)$ and $\frac{\partial v}{\partial y}(T)$ respectively. $\alpha_1$, $\alpha_2$ and $\alpha_3$ are as defined in \cref{BSNetsubsection}. $\mathcal{A}(T)$ is the Jacobian determinant of $f$ on $T$. The following theorem discusses the bijectivity of the mapping returned by the BSNet under certain conditions.

\begin{theorem}[Bijective network]
For any input data ${\bf d}$, suppose there exists $\epsilon=\epsilon({\bf d})>0$ such that the activation function $\mathcal{T}$ rescales the output feature $\nu({\bf d})$ to obtain $\mu({\bf d})$ that satisfies $||\mu({\bf d})||^2_{\infty} \leq 1-\epsilon$. Then there exists $M=M({\bf d})>0$ such that whenever $\mathcal{E}_{BC}(T) <M\epsilon$ for all $T\in \mathcal{F}$, the proposed network outputs a bijective mapping $f$. 
\end{theorem}

\begin{figure}[h]
    \centering
    \includegraphics[width=.9\textwidth]{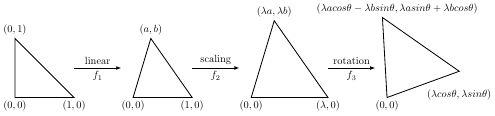}
    \caption{Illustration of $f|_T$}
    \label{fig:f_T}
\end{figure}

\begin{proof}
Given a BC $\mu$ of a piecewise linear map $f:= u+ iv$, $\mu(T)$ is a complex constant. Under the discretization of the rectangular domain as shown in \cref{triangular_discretization}, each $T\in \mathcal{F}$ is a right-angled isosceles triangle. 

Consider the transformations as shown in \cref{fig:f_T}. Without loss of generality, $f|_T$ can be regarded as a linear map that maps $(0,0)\mapsto (0,0)$, $(1,0)\mapsto (\lambda \cos\theta,\lambda \sin\theta)$ and $(0,1)\mapsto (\lambda a \cos\theta - \lambda b \sin\theta, \lambda a \sin\theta + \lambda b \cos\theta)$. Scaling and rotation are conformal transformations, thus $f_1$, $f_2\circ f_1$ and $f_3\circ f_2\circ f_1$ share the same $\mu(T)$. 

Under our framework, $\mu(T)$ is well-defined for all $T\in\mathcal{F}$. Thus, there exists $C_1>0$ such that $\lambda \geq C_1$ for all $T\in \mathcal{F}$. 

Let $f^{\mu(T)}=f_1$, then there exists a constant $C_2 >0$ such that $|\frac{\partial f^{\mu(T)}}{\partial z}|\geq C_2$ for all $T\in \mathcal{F}$ since $\mu(T) = \frac{\partial f^{\mu(T)}}{\partial \bar{z}} / \frac{\partial f^{\mu(T)}}{\partial z}$ is well-defined. 

Let $M = C_1^2 C_2^2/2$. Now if $\mathcal{E}_{BC}(T)< M\epsilon$ for all $T\in \mathcal{F}$, then
\begin{equation}
\begin{split}
    M\epsilon & > \mathcal{E}_{BC}(T)\\
    & = \frac{\lambda^2}{2}(\alpha_1(T) + 2\alpha_2(T) a + \alpha_3(T) a^2) + \frac{\lambda^2 \alpha_3(T)b^2}{2} - \lambda^2 b\\
    & = \frac{\lambda^2}{2}|\frac{\partial f^{\mu(T)}}{\partial z}|^2 (1-|\mu(T)|^2) + \frac{\lambda^2 \alpha_3(T)b^2}{2} - \lambda^2 b\\
    & \geq M\epsilon + \frac{\lambda^2 \alpha_3(T)b^2}{2} - \lambda^2 b \text{, using the fact that } ||\mu||_{\infty}^2 \leq 1-\epsilon.
\end{split}
\end{equation}
\noindent Hence, $\frac{\lambda^2 \alpha_3(T)b^2}{2} - \lambda^2 b < 0$. This implies $b>0$. We can conclude that $f|_T$ is orientation-preserving for all $T\in \mathcal{F}$. Since $f$ is proper, we conclude that $f$ is a global bijection.
\end{proof}

Therefore, as the loss function decreases to a certain level during the training process, the output mapping from the network is bijective.

\medskip

Next, given the complicated loss function derived from the mapping problem, a crucial question is whether our proposed network has an optimal solution during the training process. The following theorem gives an affirmative answer.

\begin{theorem}[Existence of minimizer]
Let the admissible set of parameters be:
\[
\mathcal{S} = \{ \phi \in \overline{B(0,r)}\subset\mathbb{R}^{m}: E_{\phi}({\bf d})\in \mathcal{C}_1; B_{\varphi}(E_{\phi}({\bf d}))\in \mathcal{C}_2 \text{ for all } {\bf d}\in \mathcal{D} \}.
\]
\noindent where $\epsilon >0$ is a positive constant and $\mathcal{D}$ denotes the training dataset. $\mathcal{C}_1$ and $\mathcal{C}_2$ are the constraint spaces according to the mapping problem. Suppose $\mathcal{C}_1$ and $\mathcal{C}_2$ are complete. $\mathcal{L}(\phi;{\bf d})$ is continuous in $\phi$ for every ${\bf d}\in \mathcal{D}$. Then, the proposed network has an optimal solution in $\mathcal{S}$ during the training process.
\end{theorem}

\begin{proof}
$\mathcal{S}$ is non-empty by assumption. Consider the parameter in the network $\phi$ as a vector in $\mathbb{R}^{m}$, where $m$ is the number of parameters in the estimator network. We proceed to show that $\mathcal{S}$ is compact under the standard Euclidean norm. Take any Cauchy sequence $\{\phi_n\}_{n=1}^{\infty}$ in $\mathcal{S}$. Since $\mathbb{R}^{m}$ is complete, $\phi_n \to \phi^*$ in  $\mathbb{R}^{m}$. Since the estimator network is continuous in $\phi$, $E_{\phi_n}({\bf d})\to E_{\phi^*}({\bf d})$. Since $\mathcal{C}$ is complete, $E_{\phi^*}({\bf d})\in \mathcal{C}$. Thus, $\phi^* \in \mathcal{S}$.

On the other hand, $\mathcal{S}$ is totally bounded since $\mathcal{S}\subset\overline{B(0,r)}$. We conclude that $\mathcal{S}$ is compact.

Now, the parameters $\phi$ is trained by optimizing the expectation of $\mathcal{L}(\phi;{\bf d})$ over the available training dataset $\mathcal{D}$. Suppose $\mathcal{D} =\{{\bf d}_1,{\bf d}_2,...,{\bf d}_N\}$ The expectation can be written as:
\begin{equation}
    \mathbb{E}_{{\bf d}\in \mathcal{D}}\ ( \mathcal{L}(\phi;{\bf d})) = \frac{1}{N} \sum_{i=1}^N \mathcal{L}(\phi;{\bf d}_i). 
\end{equation}
\noindent Since $\mathcal{L}(\phi;{\bf d}_i)$ is continuous in $\theta$ for all $1\leq i\leq N$, $\mathbb{E}_{{\bf d}\in \mathcal{D}}\ ( \mathcal{L}(\phi;{\bf d}))$ is continuous in $\phi$. Thus, there exist a global minimizer $\phi^*$ of $\mathbb{E}_{{\bf d}\in \mathcal{D}}\ ( \mathcal{L}(\phi;{\bf d}))$ in the compact set $\mathcal{S}$.
\end{proof}

We remark that the conditions on the constraint spaces $\mathcal{C}_1$ and $\mathcal{C}_2$ are realistic. These can often be fulfilled in most of the mapping problems. For instance, in the setting of the Quasiconformal Registration model \cite{lam2014landmark}, the constraint spaces for $E_{\phi}({\bf d})$ and $B_{\varphi}(E_{\phi}({\bf d}))$ are designed to satisfy: $E_{\phi}({\bf d})(p_i)\leq C_1 +\epsilon$ and $B_{\varphi}(E_{\phi}({\bf d}))(p_i) = q_i$ for $i=1,2,...,n$. The associated constrained spaces are both complete.

\section{Applications in Imaging: Diffeomorphic Image Registration}\label{sec:QCRegNet} In this section, we describe applications of our proposed framework to an imaging problem. More specifically, we illustrate the idea of our proposed deep learning framework by applying it to solve the image registration problem. 

Image registration aims to establish a meaningful dense correspondence between two related images. Registration is necessary for comparing and integrating data obtained from different measurements. Applications can be found in various fields, such as  remote sensing, medical imaging and computer vision. Transformation models, which deform one (moving) image into another (template) image, are commonly used technique to solve the registration problem. In many practical applications, the transformation between two images should be diffeomorphic. For instance, the transformation between two medical images capturing a corresponding anatomical structure should be a diffeomorphism. In this scenario, the diffeomorphic image registration problem has to be solved.

Mathematically, the diffeomorphic image registration problem can be formulated as a mapping problem to find an optimal diffeomorphic registration map between two corresponding images. Denote the two corresponding images to be registered by $I_1:\Omega \to \mathbb{R}$ and $I_2:\Omega \to \mathbb{R}$, where $\Omega$ is the image domain. Our goal is to find a diffeomorphic registration map $f:\Omega\to \Omega$ with the following conditions:
\begin{enumerate}
    \item $I_2(f(z)) = I_1(z)$ for $z\in \Omega$;
    \item $f$ is regularized;
    \item $f$ is in a suitable constrained space $\mathcal{C}$.
\end{enumerate}

The registration problem is usually formulated as a mapping problem to search for an optimal mapping $f$ that minimizes an energy functional:
\begin{equation}
    \mathcal{E}_{registration} = E_{fidelity}(f) + E_{reg}(f),
\end{equation}

\noindent where $E_{fidelity}$ is the fidelity term for matching the intensities of the two images to satisfy condition 1. $E_{reg}$ is the regularization term to enhance the smoothness of the mapping. In addition, $f$ should lie in certain constraint space $\mathcal{C}$, such as the space of diffeomorphisms for diffeomorphic image registration.

In this work, we adopt the quasiconformal registration model \cite{lam2014landmark} to design the loss function to train the deep neural network for the diffeomorphic image registration. In the quasiconformal registration framework, the fidelity term is chosen as the intensity mismatching error:
\begin{equation}
    E_{fidelity}(f) = \int_{\Omega} (I_1 - I_2\circ f)^2
\end{equation}
\noindent The regularization term is defined in terms of the Beltrami coefficient of $f$:
\begin{equation}
    E_{reg}(f) = \int_{\Omega} \alpha |\mu(f) |^2 + \beta |\nabla(\mu(f))|^2
\end{equation}
\noindent where $\mu(f) = \frac{\partial f}{\partial \bar{z}}/\frac{\partial f}{\partial z}$ is the Beltrami coefficient of $f$. The first term minimizes the local geometric distortion under $f$. The second term further enhances the smoothness of the mapping $f$. To ensure that $f$ is a diffeomorphism, we further require that $|\mu(f)(z)|<1$ for all $z\in \Omega$. In other words, $||\mu(f)||_{\infty}<1$.

Since the Beltrami coefficient is an effective geometric quantity to represent the mapping, the above mapping problem can be formulated in terms of the Beltrami coefficient. More precisely, the mapping problem for diffeomorphic image registration can be written as finding an optimal mapping $f:\Omega\to \Omega$ that minimizes
\begin{equation}
    \mathcal{E}_{registration} (\mu) = \int_{\Omega} \alpha |\mu|^2 + \beta |\nabla \mu|^2 + (I_1 - I_2\circ f^{\mu})^2
\end{equation}

\noindent subject to the conditions that $||\mu||_{\infty}<1$ and $\frac{\partial f^{\mu}}{\partial \bar{z}} = \mu \frac{\partial f^{\mu}}{\partial z}$.

\begin{figure*}[t!]
	\centering
	\includegraphics[width=5in]{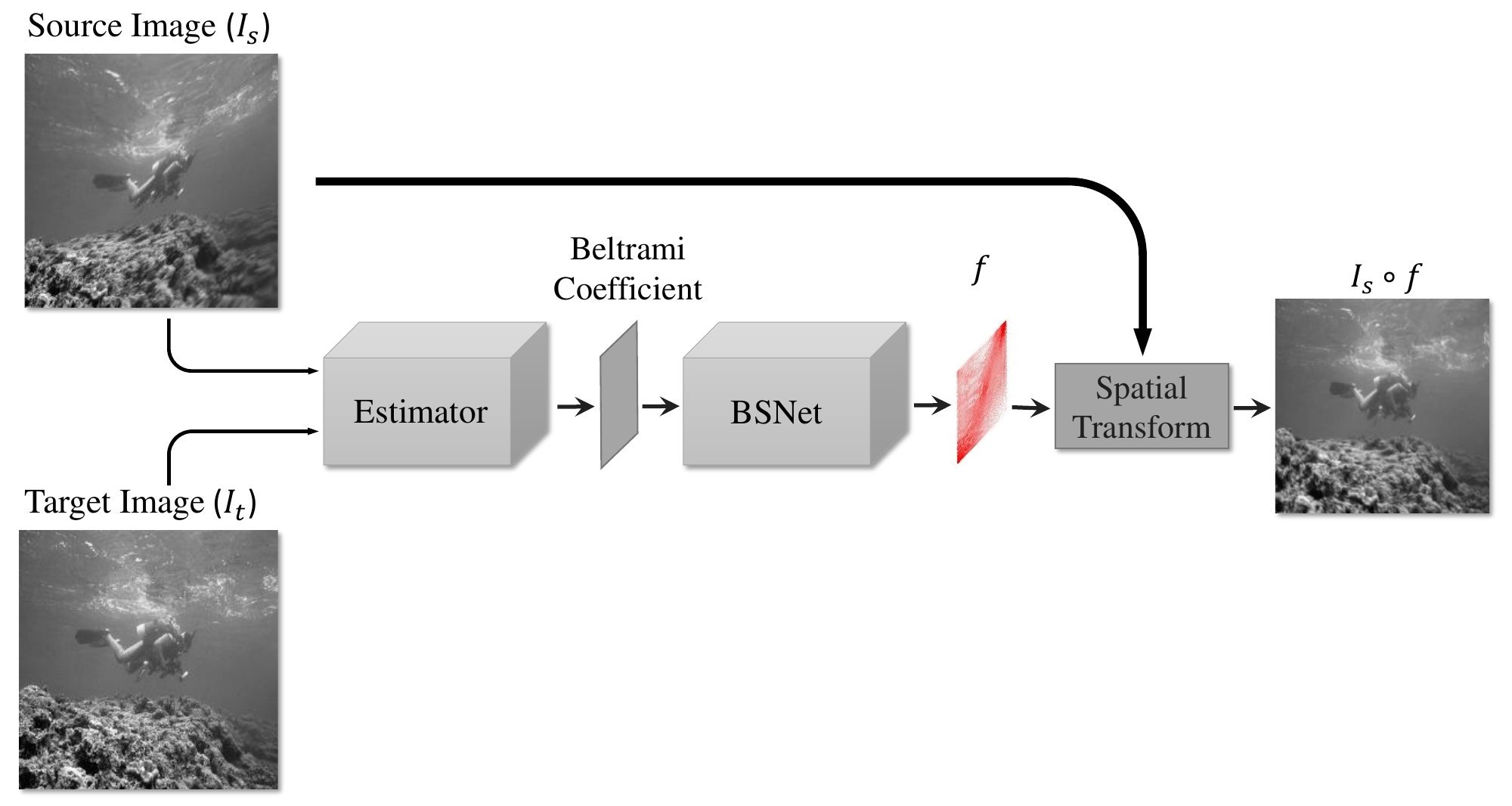}
	\caption{The Network Structure of the Quasiconformal Registration Network (QCRegNet).}
	\label{QCRegNet_Structure} 
\end{figure*}

We apply the proposed deep learning framework to solve the above mapping problem. More specifically, we propose the deep neural network, called the {\it Quasiconformal Registration Network (QCRegNet)}, to solve the diffeomorphic registration problem. The overall network structure is shown in \cref{QCRegNet_Structure}. As shown in \cref{QCRegNet_Structure}, QCRegNet takes two corresponding images $I_1$ and $I_2$ as the input data. They are passed to the estimator network $E_\phi$ to get the Beltrami coefficient $\mu$ associated to the optimal quasiconformal map $f^{\mu}$. To ensure the constraint that $||\mu||_{\infty}<1$, the activation function $\mathcal{T}$ as described in Equation \cref{activation} is used. The BC $\mu = E_{\phi}(I_1,I_2)$ is then passed to the BSNet to obtain the quasiconformal map $f^{\mu}$. The moving image $I_2$ is then deformed by $f^{\mu}$ to get a deformed image $\tilde{I} = I_2\circ f^{\mu}$. Upon successful training, $\tilde{I}$ should closely resemble $I_1$. The proposed QCRegNet can be trained in an unsupervised manner. Assuming the parameters in the BSNet is pretrained, given a collection of image pairs $\mathcal{I}=\{I_1^k, I_2^k\}_{k=1}^N$, we propose to find the optimal parameters $\phi$ that minimizes the expectation $\mathbb{E}_{(I_1,I_2)\in \mathcal{I}} (\mathcal{L}_{QCRegNet}(I_1,I_2;\phi))$, where
\begin{equation}\label{Estimator_loss}
    \mathcal{L}_{QCRegNet}(I_1,I_2;\phi) = \alpha ||E_{\phi}(I_1,I_2)||_2^2 + \beta ||\nabla E_{\phi}(I_1,I_2)||_2^2 + \eta  ||\tilde{I}-I_1||_2^2.
\end{equation}

\noindent The optimization problem is solved using the RMSprop gradient descent method \cite{rmsprop}.

\section{Implementation}\label{sec:implementation}
The experiments are conducted with PyTorch framework \cite{NEURIPS2019_9015}. In our experiments, we adopt a two-stage training scheme. The BSNet is trained at the first stage, followed by the training process of the estimator network with the trained BSNet fixed and serving as the Beltrami equation solver that converts the Beltrami coefficient $\mu$ to its corresponding QC map. 

Both two models are trained independently with 2000 epochs using the RMSprop optimizer \cite{rmsprop} with a momentum of 0.9. For the training of BSNet, the learning rate is set to 0.0002. For the training of the estimator, the hyper-parameters $\alpha$, $\beta$ and $\eta$ in \cref{Estimator_loss} are set to 1, 1 and 400 respectively and the learning rate is set to 0.00002.

Note that the BSNet plays the role of Beltrami equation solver that maps any given Beltrami coefficient to its corresponding deformation map. Hence the BSNet is indeed irrelevant to any specific task. On top of that, it is unnecessary to retrain the BSNet whenever we encounter a new task that deals with data of the same size.

The BSNet is trained in an unsupervised way since the ground-truth QC maps which can serve as training labels are not available. In the loss function we solve the discretized second-order elliptic PDE, i.e. the linear system \cref{linear_system}, within which $C_s$ and $C_t$ are computed directly from the input Beltrami coefficient $\mu$. We synthesize all kinds of Beltrami coefficients as training data using ILSVRC2012 \cite{ILSVRC15}, a dataset consisting of large amounts of realistic images. We preprocess these images by data augmentation techniques, then use them to generate smooth $\mu$ and noisy $\mu$. By combining the smooth $\mu$ and noisy $\mu$, we generate acceptable $\mu$ associated to `realistic' mappings which possess both low and high frequency information. 

To demonstrate the effectiveness of our proposed algorithm, for the underwater image registration task, we train for comparison a VoxelMorph \cite{balakrishnan2019voxelmorph} model with the same dataset, preprocessing methods and its default hyper-parameters for 2000 epochs. We also show some experimental results generated by the Hyperelastic methods \cite{burger2013hyperelastic}. 

\section{Experimental Results}\label{sec:exp_results}

To investigate the efficacy of our proposed framework, numerous experiments have been carried out. The experimental results are reported in this section. Our proposed framework relies heavily on the BSNet, which reconstructs the quasiconformal map from a given Beltrami coefficient. The accuracy and efficiency of the BSNet will be examined. To test the effectiveness of the proposed learning framework for mapping problems, we apply the framework to solve the diffeomorphic image registration problem. More specifically, the QCRegNet is built, which are tested on real images, such as medical images and underwater images.

In order to quantitatively measure the accuracy, the following two indicators are considered:
\begin{itemize}
    \item We consider the number of triangular faces on which Jacobian determinant $\text{det}(J)$ of the map is less than 0. We denote this indicator by $N_J$. As described in \cref{BSNetsubsection}, the image domain is discretized by a triangular mesh. A discrete map is piecewise linear on each triangular face. $\text{det}(J) <0$ means the map has an orientation change and thus a folding occurs. $N_J$ shows the number of folded triangular faces.
    \item We also consider the following indicator:
    \[
    S_J = \sum_{det(J) <0} |det(J)|.
    \]
    $S_J$ is the sum of the absolute value of negative $det(J)$.
    This indicator shows the extent of the folding. As shown in \cref{sec:uw_reg}, the larger this indicator is, the more the severity of the folding.
\end{itemize}

\subsection{Performance of the BSNet}
A key component of our proposed framework is the BSNet, which allows us to reconstruct the quasiconformal map from a prescribed Beltrami coefficient. In this subsection, we demonstrate the effectiveness of the BSNet by the experimental results. An ablation study is also carried out to illustrate the importance of each structure of the BSNet.

\subsubsection{Reconstruction of quasiconformal maps}
\begin{figure}[t]
    \centering
    \begin{tabularx}{\textwidth}{cX}
        \multirow{2}{0.24\textwidth}{\includegraphics[width=\linewidth]{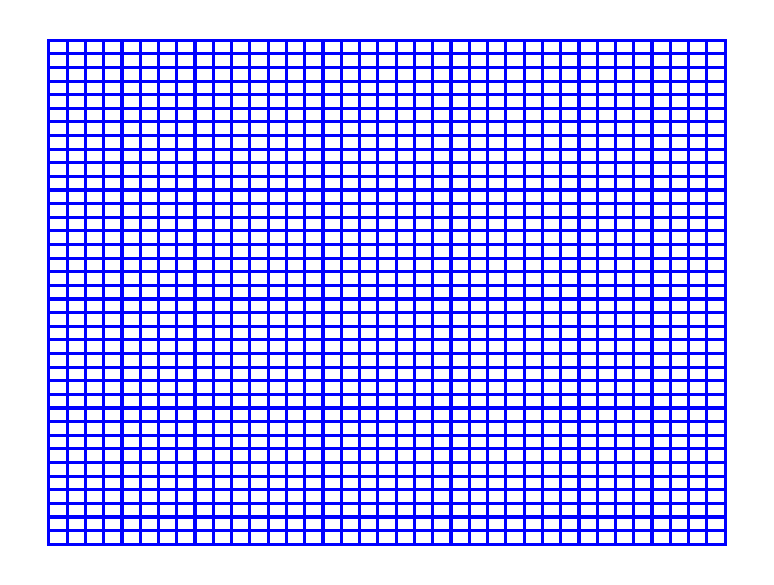}
        \subcaption{Source grid}}
    &   
            \hspace{0.25em}
            \includegraphics[width=0.30\linewidth]{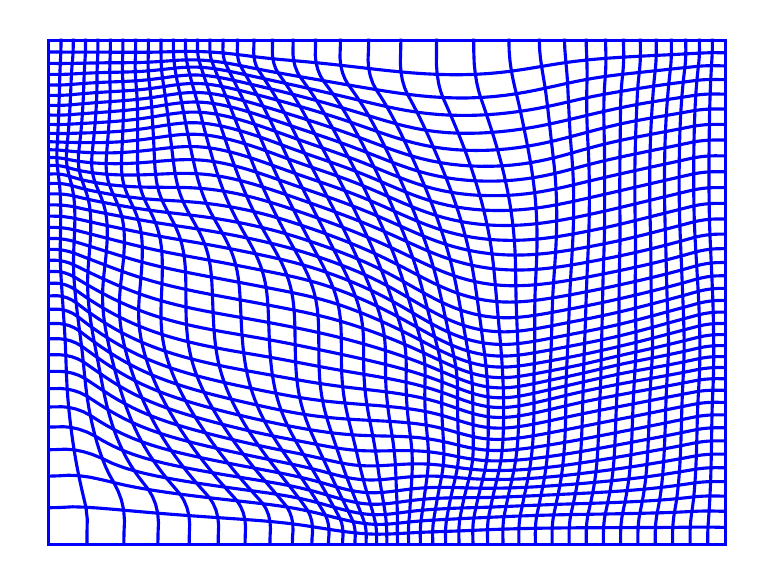}
            \hspace{0.5em}
            \includegraphics[width=0.30\linewidth]{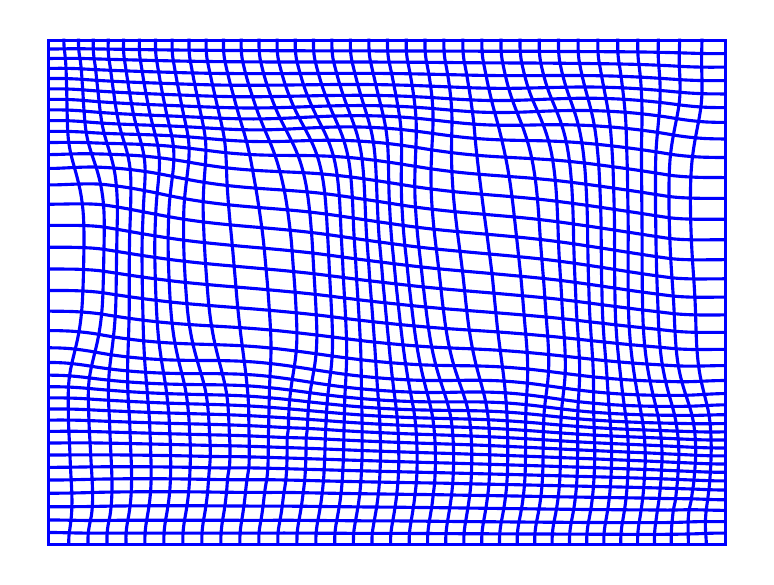}
            \hspace{0.5em}
            \includegraphics[width=0.30\linewidth]{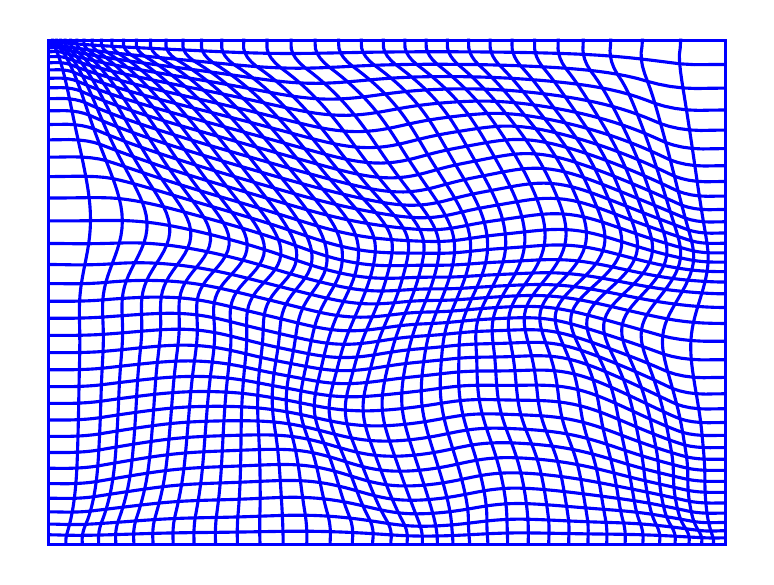}
            \subcaption{Mappings generated by Linear Beltrami Solver (LBS)}
        \\
    &   
            \hspace{0.25em}
            \includegraphics[width=0.30\linewidth]{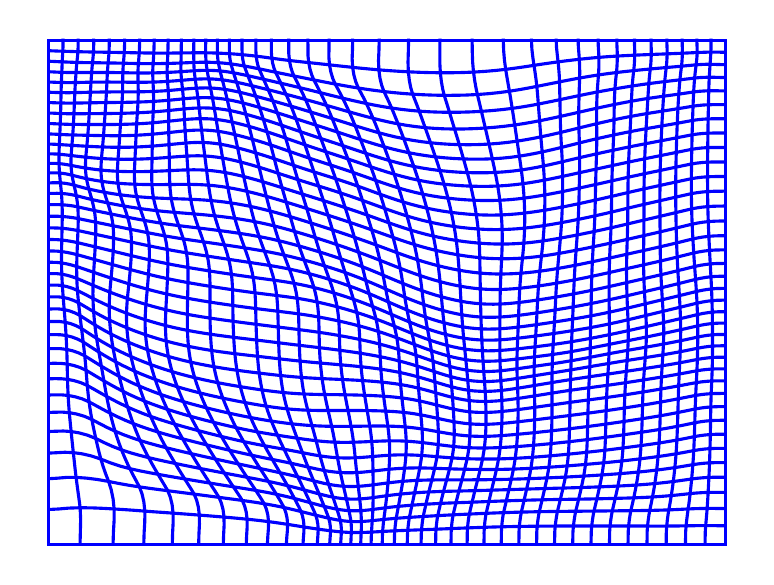}
            \hspace{0.5em}
            \includegraphics[width=0.30\linewidth]{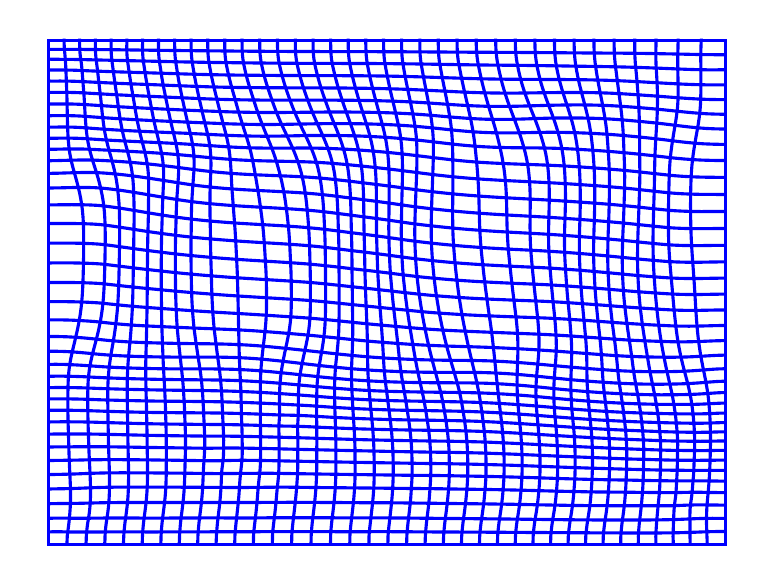}
            \hspace{0.5em}
            \includegraphics[width=0.30\linewidth]{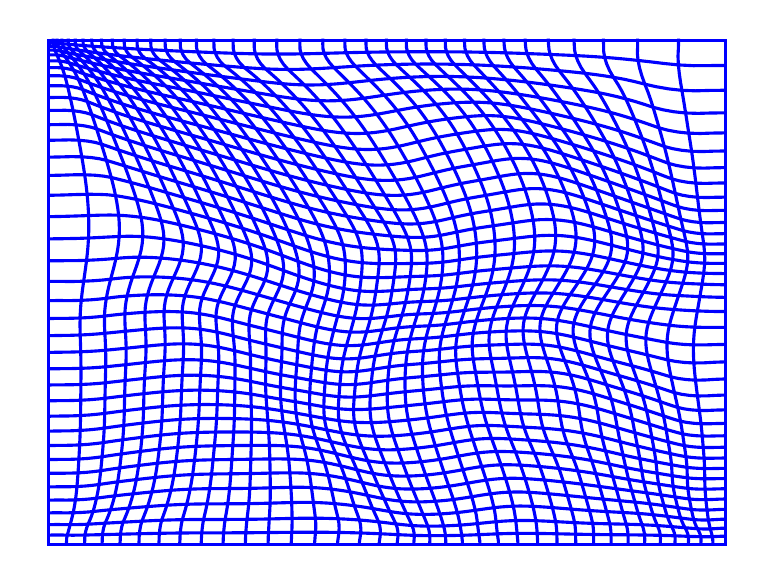}
            \subcaption{Mappings generated by Beltrami Solver Network (BSNet)}
        
    \end{tabularx}
    \caption{Illustration of mappings generated by LBS and BSNet. Each column shows the mappings generated by LBS and BSNet given the same Beltrami coefficient. The error of the three columns (the Frobenius norm of the difference between the resulting coefficients of the two mappings) are $9.7622$, $9.8390$ and $7.3868$ respectively.}
    \label{BSNetReconstruction}
    \end{figure}

We first investigate the effectiveness of reconstructing the quasiconformal map from a given Beltrami coefficient using the BSNet, which obeys the Beltrami's equation. \cref{BSNetReconstruction} shows the reconstruction of quasiconformal maps from prescribed Beltrami coefficients using the Linear Beltrami Solver (LBS) \cite{lui2013texture} and BSNet. (a) shows a grid on the source domain. It is deformed by quasiconformal maps associated to different Beltrami coefficients. (b) shows the quasiconformal maps with respect to different Beltrami coefficients, which are reconstructed by LBS. The quasiconformal maps are visualized as deformations from a regular grid on the domain. (c) shows quasiconformal maps reconstructed by the BSNet. It is observed that the reconstructed maps from the LBS and BSNet closely resemble each other. We further compare the difference between the maps reconstructed by the LBS and BSNet by computing the difference between their resulting Beltrami coefficients. Here we use the Frobenius norm of the difference between the resulting coefficients of the two mappings to indicate the error. For the three Beltrami coefficients, the errors are only 9.7622, 9.8390 and 7.3868 respectively, which are very small, demonstrating the fact that the maps reconstructed by LBS and BSNet are almost the same. 

\subsubsection{Ablation Study} In this subsection, we further study the necessity of different components in the BSNet. An ablation study is carried out and the results are reported below.

\smallskip

\begin{figure}[!htp]
    	\centering
    	\begin{tabularx}{\textwidth}{cXX}
    	    \multirow{2}{0.33\textwidth}{\includegraphics[width=\linewidth]{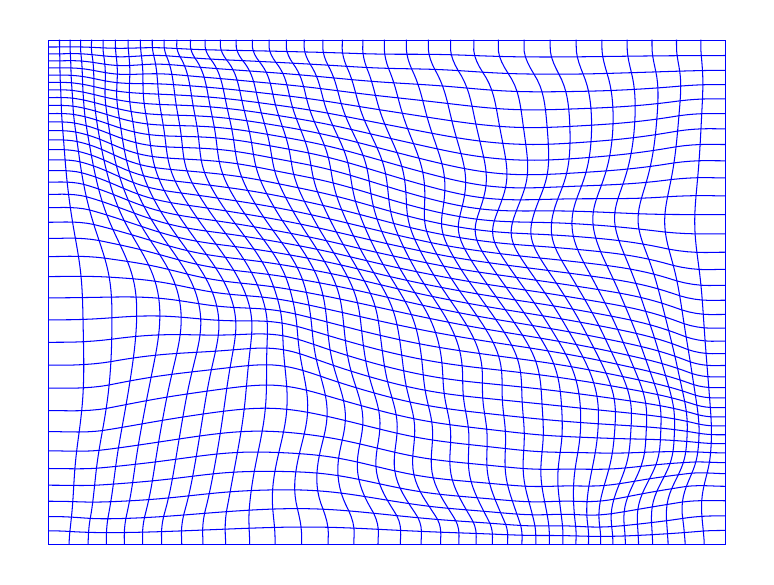}
            \subcaption{Target mapping}}
        &   \includegraphics[width=\linewidth]{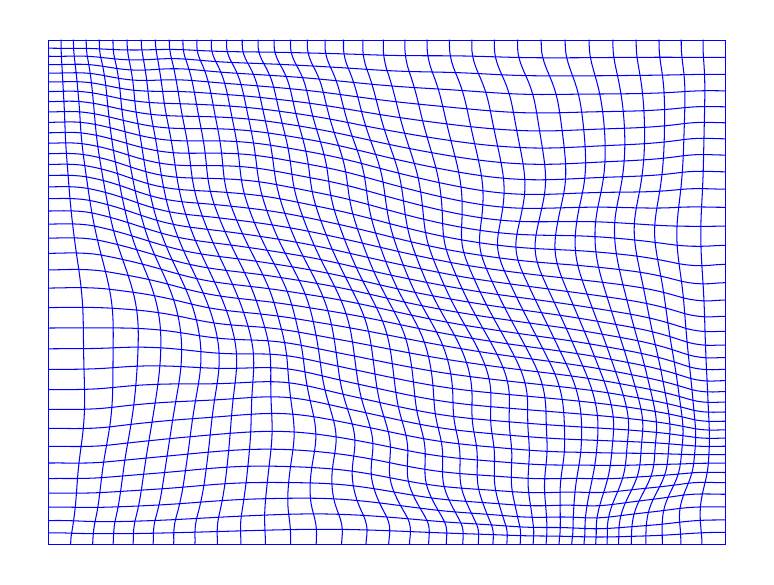} 
            \subcaption{BSNet} 
        &   \includegraphics[width=\linewidth]{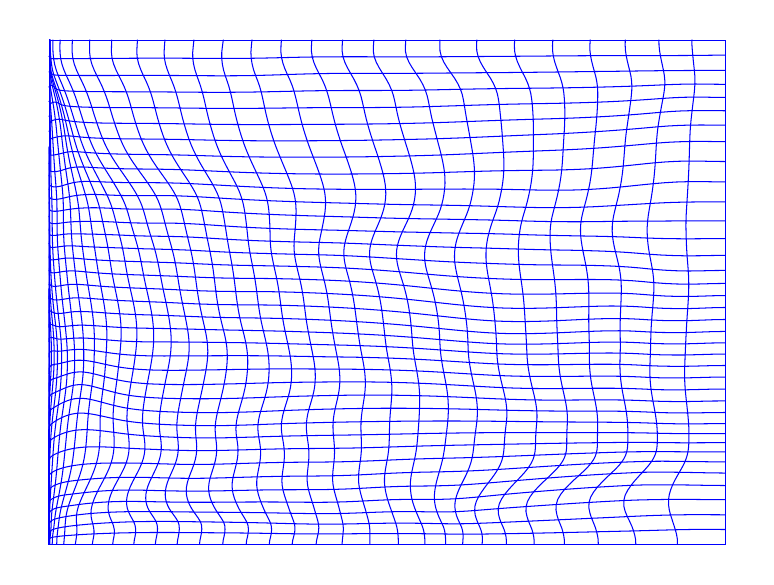}
            \subcaption{BSNet-noDTL}\\
        &   \includegraphics[width=\linewidth]{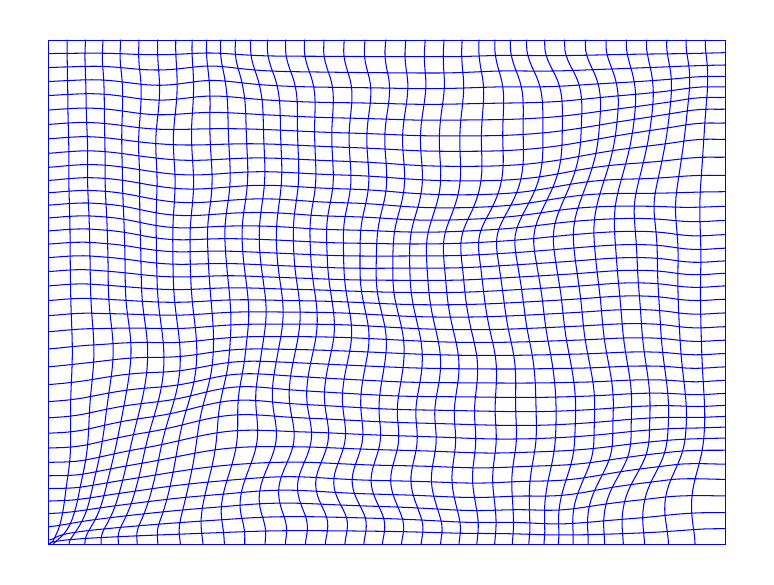}
            \subcaption{BSNet-conv1}
        &   \includegraphics[width=\linewidth]{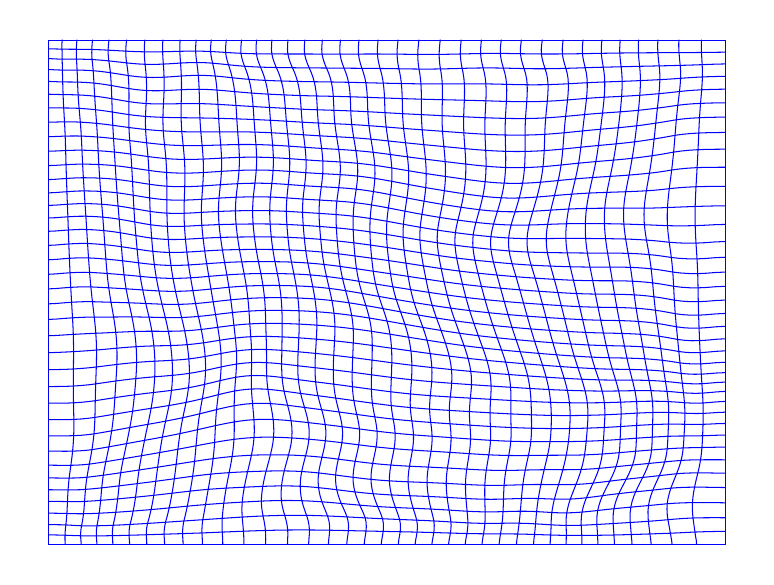}
            \subcaption{BSNet-conv2}                                      
        \end{tabularx}
        \caption{Examples generated by experiment of replacing DTL with different structure.}\label{fig:nodtl1}

    	\begin{tabularx}{\textwidth}{cXX}
    	    \multirow{2}{0.33\textwidth}{\includegraphics[width=\linewidth]{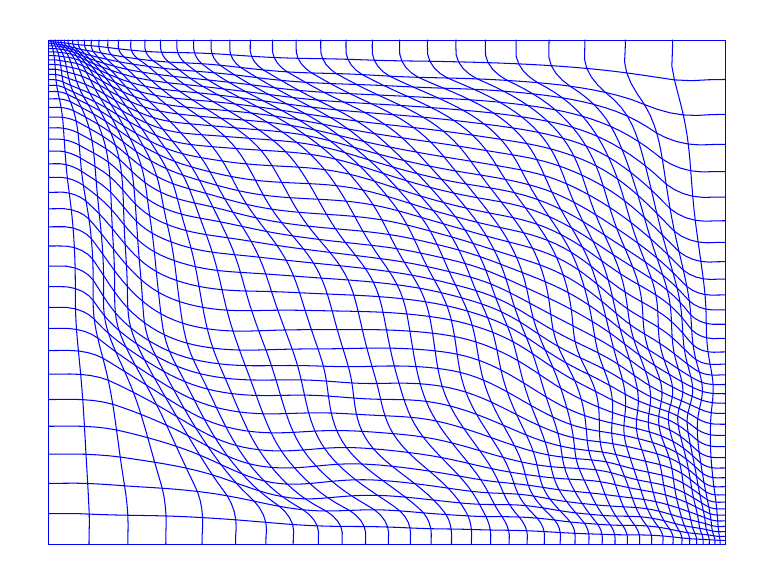}
            \subcaption{Target mapping}}
        &   \includegraphics[width=\linewidth]{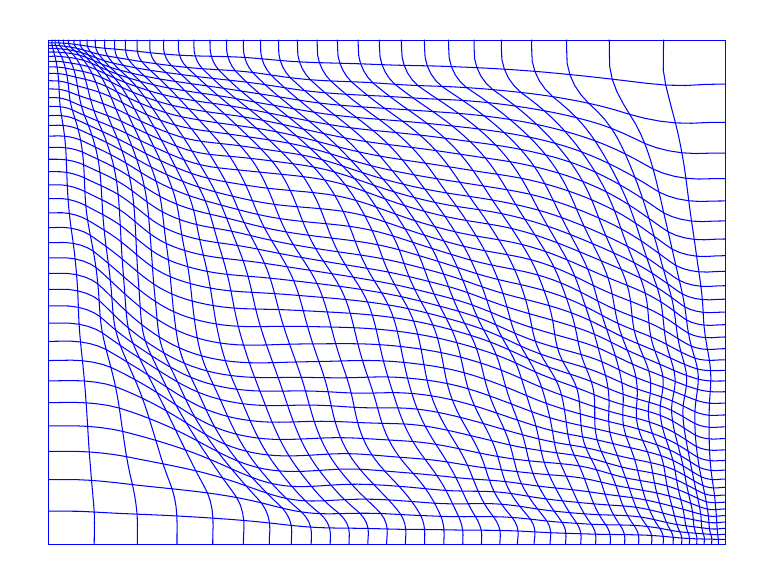} 
            \subcaption{BSNet} 
        &   \includegraphics[width=\linewidth]{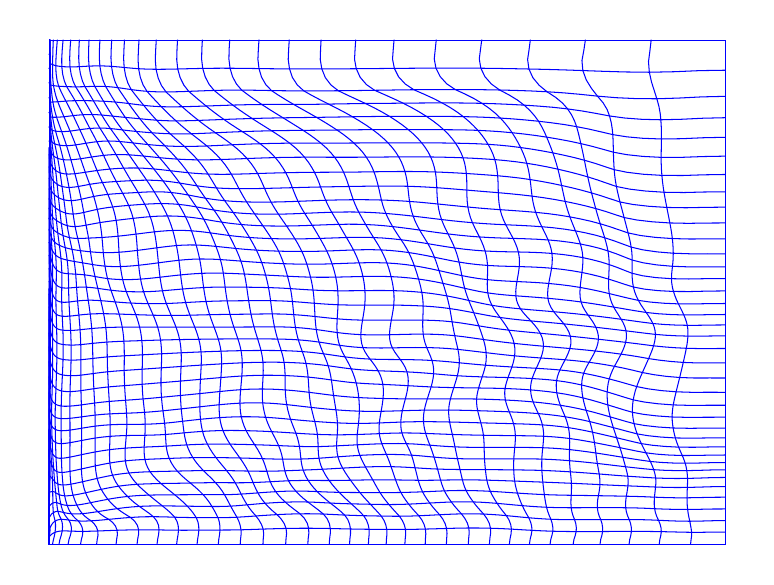}
            \subcaption{BSNet-noDTL}\\
        &   \includegraphics[width=\linewidth]{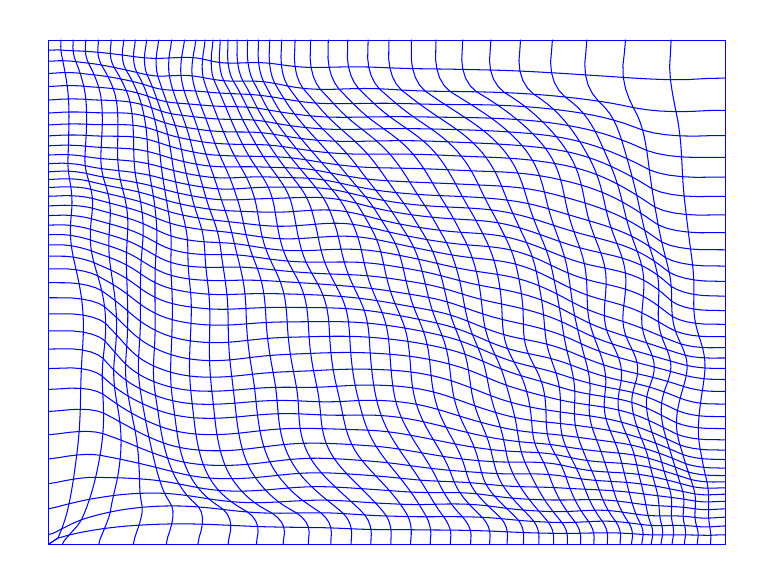}
            \subcaption{BSNet-conv1}
        &   \includegraphics[width=\linewidth]{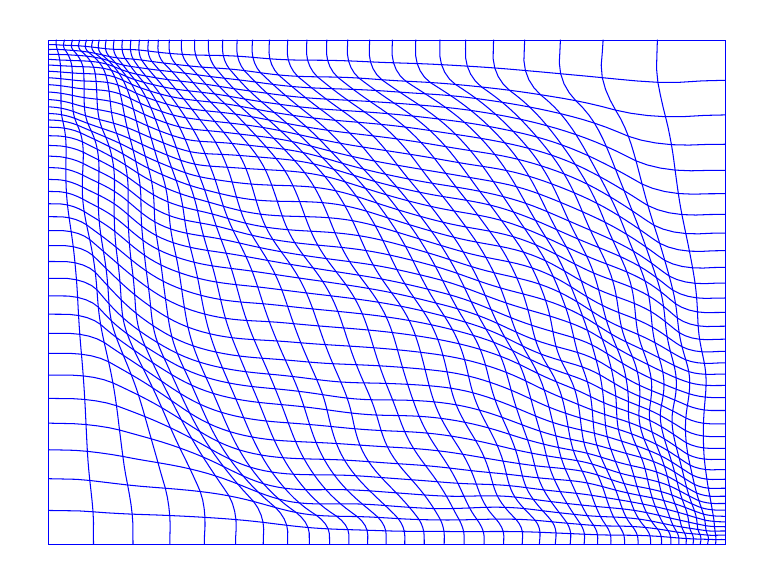}
            \subcaption{BSNet-conv2}                                      
        \end{tabularx}
        \caption{Examples generated by experiment of replacing DTL with different structure.}\label{fig:nodtl2}
    \end{figure}

    \paragraph{Necessity of Domain Transform Layer} \label{BSNet-noDTL}
    We first study the importance of the domain transform layer (DTL). The DTL transforms information from the frequency domain to the coarser spatial domain.    
    \cref{fig:nodtl1} and \cref{fig:nodtl2} show the reconstruction results with and without the DTL. In both \cref{fig:nodtl1} and \cref{fig:nodtl2}, (a) shows the target mapping visualized as the deformation of a regular grid on the domain. Each target mapping is associated to a Beltrami coefficient, which is passed to the BSNet as the input. (b) shows the reconstructed map by the BSNet. The reconstructed map closely resemble the target mapping as shown in (a). To investigate the necessity of the DTL, we remove the DTL component in the BSNet. We denote the BSNet with the DTL removed by BSNet-noDTL. (c) shows the reconstructed map by BSNet-noDTL. The reconstructed map is significantly deviated from the target mapping in (a). This illustrates the DTL component is necessary in the BSNet. To further analyze, we consider two modified BSNets, in which the DTL is replaced by one and two convolutional layers. We denote them by BSNet-conv1 and BSNet-conv2 respectively. The convolutional layers in both BSNet-conv1 and BSNet-conv2 have 64 $3\times 3$ kernels. Both networks were trained with the same hyper-parameters for 2000 epochs.
    
    Numerically, we compared $\mathcal{L}_{BS}$ of the four models on a 10000 Beltrami coefficients dataset. $\mathcal{L}_{BS}$ of BSNet is 0.000291, which is the lowest loss. BSNet-conv2 got 0.000395, which means that DTL performs better in converting information from the frequency domain to the spatial domain when compared with 2 convolutional layers. After removing a convolutional layer, the loss of this model increased to 0.000470. The BSNet-noDTL got the highest loss, which is 0.000654. These experimental results show the necessity of the DTL.
    
    \paragraph{Necessity of the Short Paths in BSNet}
    
    \begin{figure}[h]
        \centering
        \begin{subfigure}[t]{0.45\textwidth}
            \centering
    	    \includegraphics[width=0.9\textwidth]{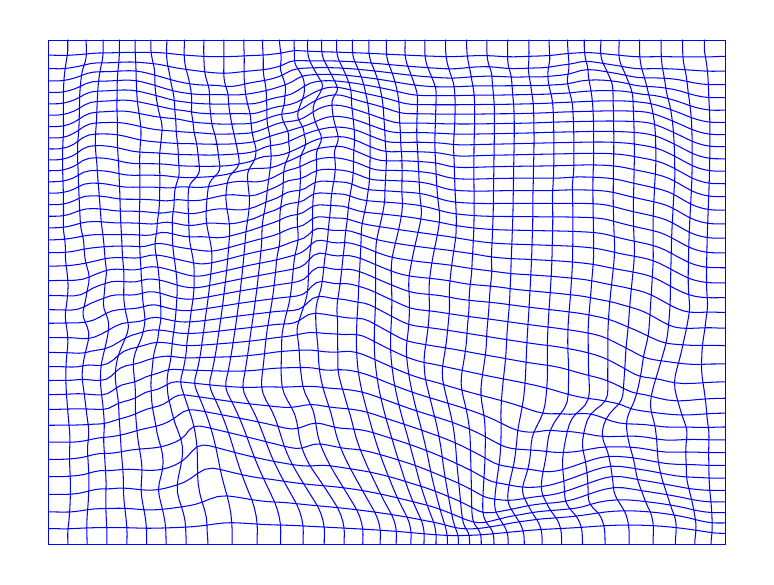}
            \caption{Target mapping}
        \end{subfigure}
        \begin{subfigure}[t]{0.45\textwidth}
            \centering
            \includegraphics[width=0.9\textwidth]{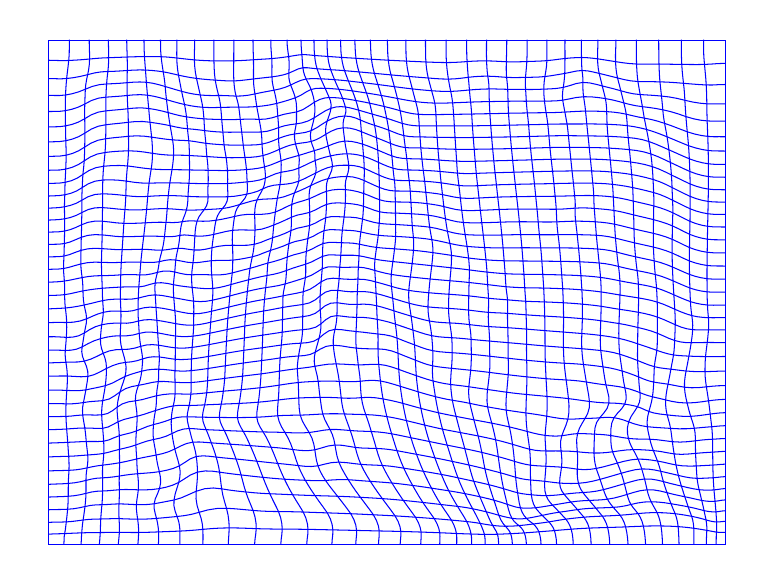} 
            \caption{BSNet}
        \end{subfigure}
        \begin{subfigure}[t]{0.45\textwidth}
            \centering
            \includegraphics[width=0.9\textwidth]{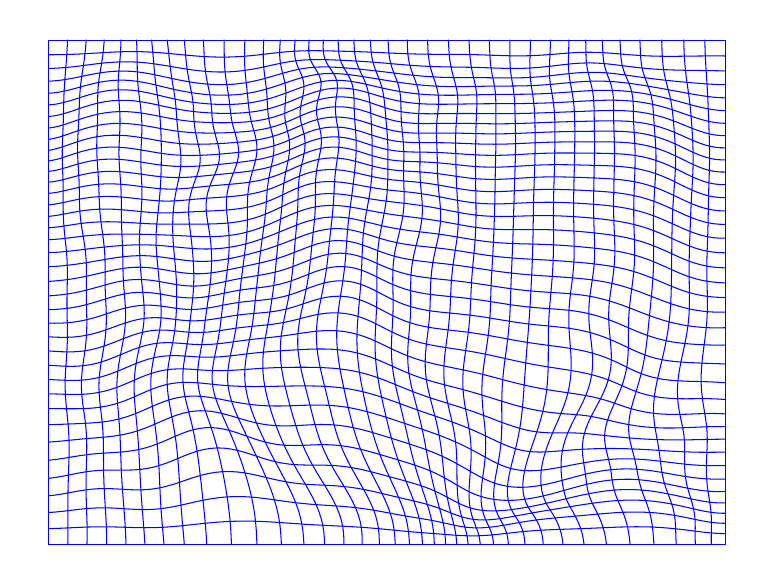}
            \caption{Reconstruction from the Fourier approximation Using LBS}
        \end{subfigure}
        \begin{subfigure}[t]{0.45\textwidth}
            \centering
            \includegraphics[width=0.9\textwidth]{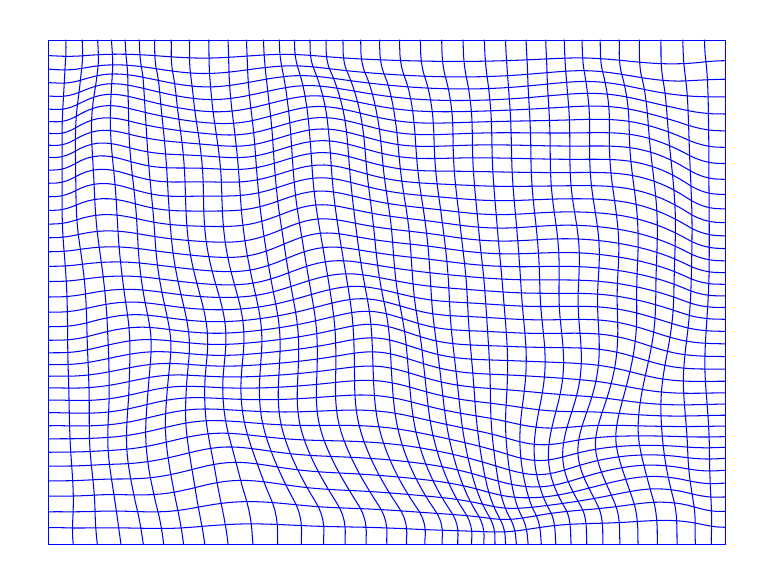}
            \caption{BSNet without the short path}
        \end{subfigure}
        \caption{Examples generated by experiment of removing the short path.}
        \label{fig:compare_path1}
    \end{figure}
    
    \begin{figure}[h]
        \centering
        \begin{subfigure}[t]{0.45\textwidth}
            \centering
    	    \includegraphics[width=0.9\textwidth]{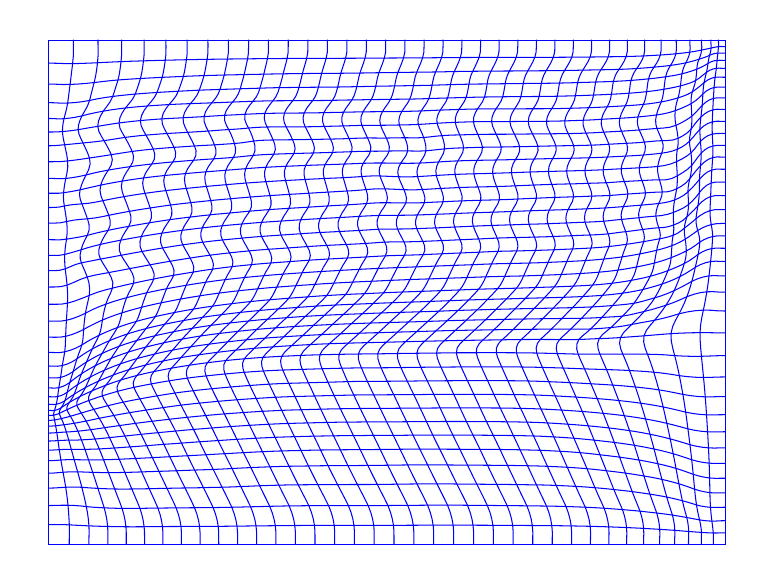}
            \caption{Target mapping}
        \end{subfigure}
        \begin{subfigure}[t]{0.45\textwidth}
            \centering
            \includegraphics[width=0.9\textwidth]{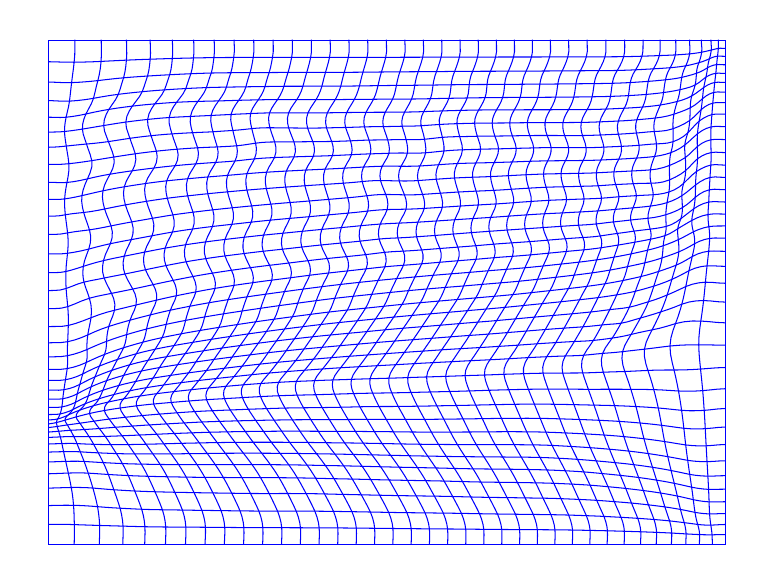} 
            \caption{BSNet}
        \end{subfigure}
        \begin{subfigure}[t]{0.45\textwidth}
            \centering
            \includegraphics[width=0.9\textwidth]{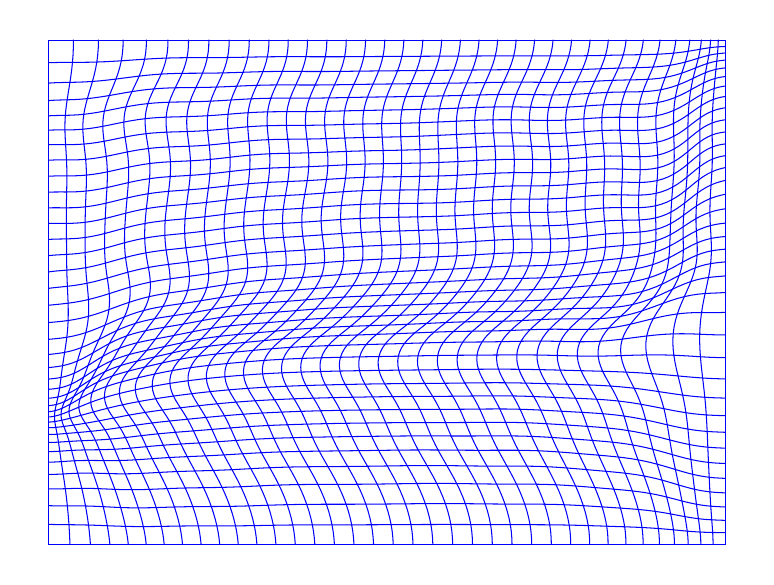}
            \caption{Reconstruction from the Fourier approximation Using LBS}
        \end{subfigure}
        \begin{subfigure}[t]{0.45\textwidth}
            \centering
            \includegraphics[width=0.9\textwidth]{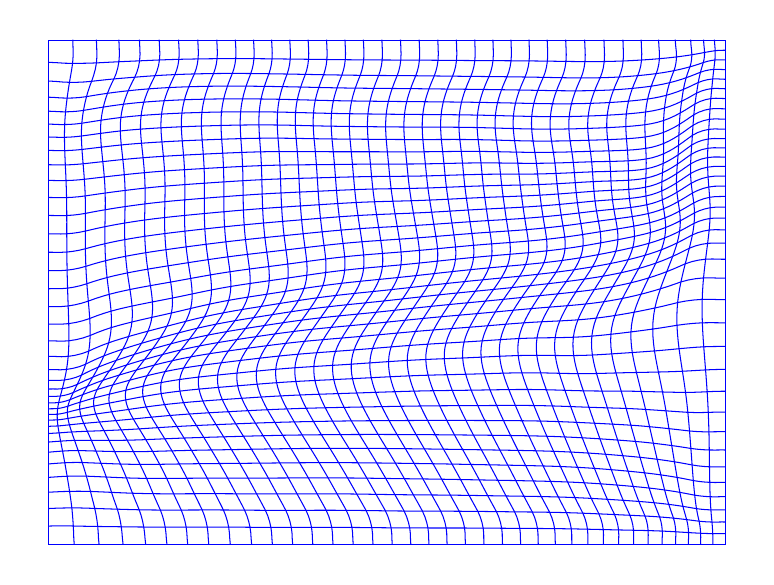}
            \caption{BSNet without the short path}
        \end{subfigure}
        \caption{Examples generated by experiment of removing the short path.}
        \label{fig:compare_path2}
    \end{figure}
    
    Recall that the BSNet has two paths. The upper path consists of the down-sampling, convolution and concatenation operations. We call this upper path the short path. In this subsection, we investigate the necessity of the short path in the BSNet. 
    
    \cref{fig:compare_path1} and \cref{fig:compare_path2} report the results. In each figure, (a) is the target mapping generated by the LBS. Again, the mapping is visualized by the deformation of a regular grid on the domain. (b) shows the reconstructed mapping by the BSNet. (c) shows the reconstructed mapping associated to a Fourier approximation of the Beltrami coefficient using the LBS; (d) shows the reconstructed mapping using the modified BSNet without the short path.
    
    Comparing (b) and (d), it is observed that the mapping in (b) has more details than (d). This indicates that the reconstructed mapping may lose subtle details if the short path is omitted in the BSNet. The short path provides high-frequency information of the mapping that is ignored by the Fourier approximation of the Beltrami coefficient. Furthermore, the mapping in (c) evidently contains more details than that in (d). It means that the mapping reconstructed by the long path alone cannot preserve all the geometric information captured by the Fourier approximation of the Beltrami coefficient. By integrating the short path in the BSNet, the missing geometric information of the mapping can be reconstructed. The comparison supports that the short path is necessary in generating the target mapping of a given Beltrami coefficient.
    
    We also compared $\mathcal{L}_{BS}$ of the three models on the same Beltrami coefficients dataset used in \cref{BSNet-noDTL}. $\mathcal{L}_{BS}$ of the modified BSNet without the short path is 0.000528, which is higher than that of the BSNet (0.000291). 
    These quantitative measurements indicates that the short path in the BSNet is necessary.
    
\subsubsection{Convergence and inference time of BSNet}

One of the superiority of BSNet against LBS is that, the inference time of BSNet is trivial. To see this, we apply the two methods to some Beltrami coefficients, and compute the mean and standard deviation of inference time. This experiment is carried out on a computer with Intel i7-10700 CPU, and there is no GPU acceleration on this machine, which ensures the fairness of the inference time comparison. The results are shown in \cref{table_infer_time_bsnet}.

\begin{table}[h!]
\centering
\begin{tabular}{ c|c|c } 
 \hline
       & Linear Beltrami Solver (LBS) & Beltrami Solver Network (BSNet) \\ 
\hline
 mean(s) & 0.4622 & 0.0135 \\ 
 std(s) & 0.0393 & 0.0007 \\ 
 \hline
\end{tabular}
\caption{Inference time of LBS and BSNet}
\label{table_infer_time_bsnet}
\end{table}

\begin{figure*}[t]
	\centering
	\includegraphics[width=3in]{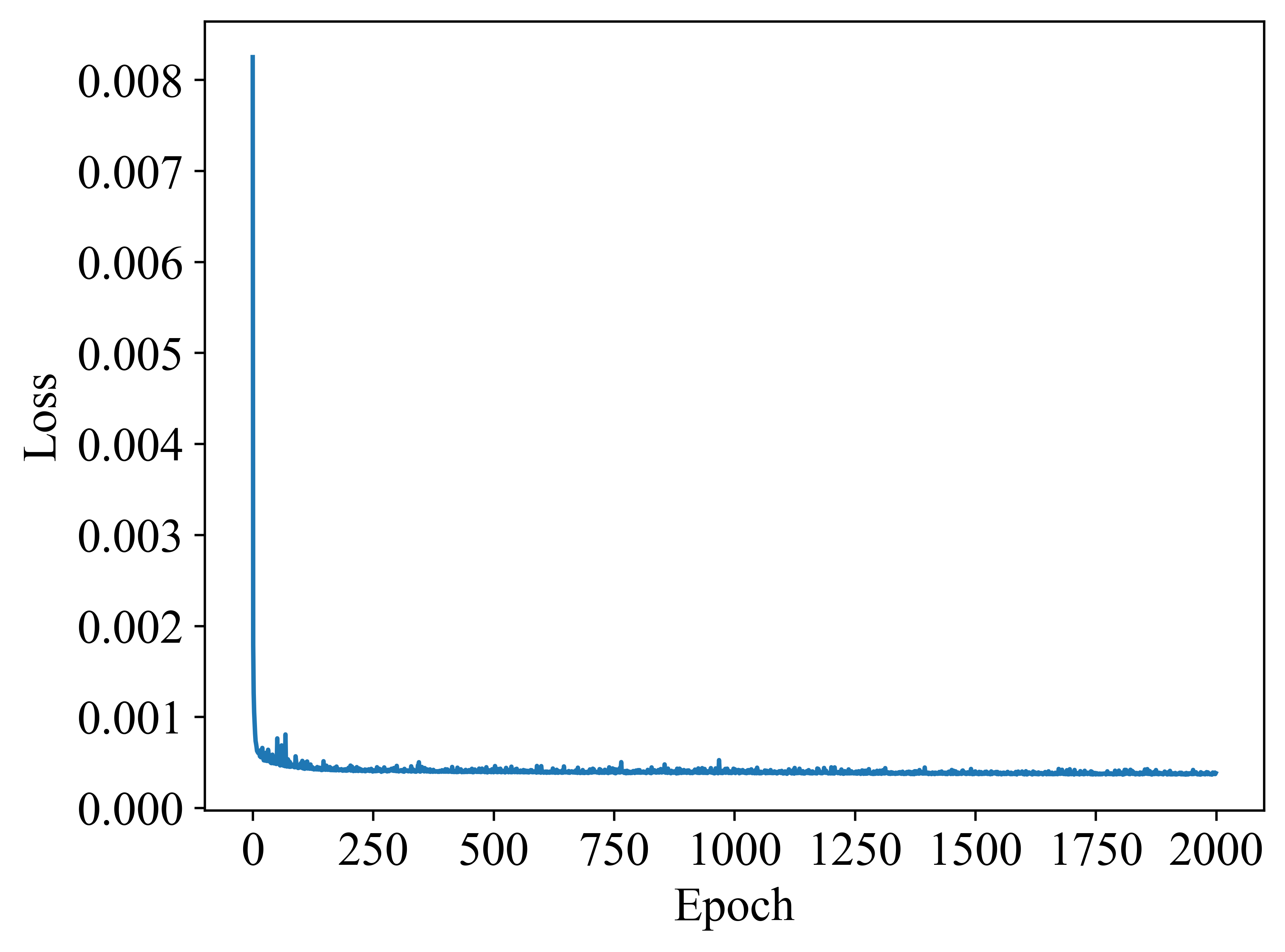}
	\caption{Training loss of BSNet}
	\label{bsnet_loss} 
\end{figure*}

The curve of the loss is shown in \cref{bsnet_loss}, from which we can be confident that the model has successfully converged during training.

\subsection{Image Registration} In this subsection, we examine the effectiveness of our proposed deep learning framework to solve the mapping problem. We test our framework to solve the diffeomorphic image registration problem with large deformations, which can be formulated as a mapping problem. A deep learning based registration network, called the QCRegNet, is trained as described in \cref{sec:QCRegNet}. We apply the QCRegNet to the image registration problems for underwater images and medical images.

    \subsubsection{Underwater Image Registration} \label{sec:uw_reg}
    Underwater image registration has attracted much attention in recent years with a wide variety of applications. Applications include the detection of minerals in oceans, the examination of biodiversity in marine environments, the study of the seabed, the investigation of crucial structures inside the sea and so on. Underwater image registration is considered to be a challenging task due to the degraded underwater images. The distortions of the underwater images can be caused by multiple factors, such as the refraction of light due to the presence of particles and the variance in density, the absorption of light in water, water turbulence, as well as the addition of noises. The underwater image pairs can be severely corrupted by image blurs, noises, and geometric distortions. Many existing image registration algorithms fail to give satisfactory results. We apply our proposed QCRegNet to solve the challenging underwater image registration problem. The results are reported in this subsection.

The QCRegNet is trained as follows. The training dataset is generated by deforming images extracted from ILSVRC2012 using mapping obtained by BSNet with random samples of BCs. Each random sample of BCs consists of two normalized $112\times 112$ images, which serve as the real and imaginary parts of the BC. The training set of image pairs is then obtained. QCRegNet takes the image pairs as its input and the estimator outputs a two-channel image representing the BC of the registration map. The network is trained by optimizing the loss function \cref{Estimator_loss}.

    \begin{figure}[!htp]
        \centering
        
        \subfloat[Fixed]{\includegraphics[width=.23\textwidth]{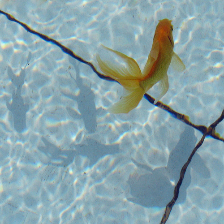}}\hspace{.1em}
        \subfloat[Hyperelastic]{\includegraphics[width=.23\textwidth]{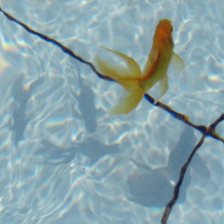}}\hspace{.1em}
        \subfloat[VoxelMorph]{\includegraphics[width=.23\textwidth]{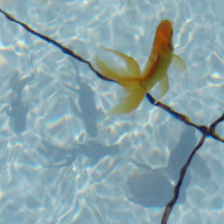}}\hspace{.1em}
        \subfloat[QCRegNet]{\includegraphics[width=.23\textwidth]{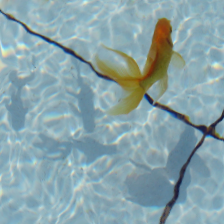}}\hspace{.1em}
        \subfloat[Moving]{\includegraphics[width=.23\textwidth]{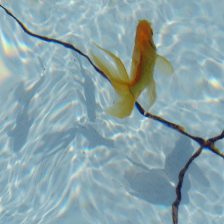}}\hspace{.1em}
        \subfloat[Hyperelastic Deform Field]{\includegraphics[width=.23\textwidth, height=.23\textwidth, trim={1.7em 1.7em 1.7em 1.7em}, clip]{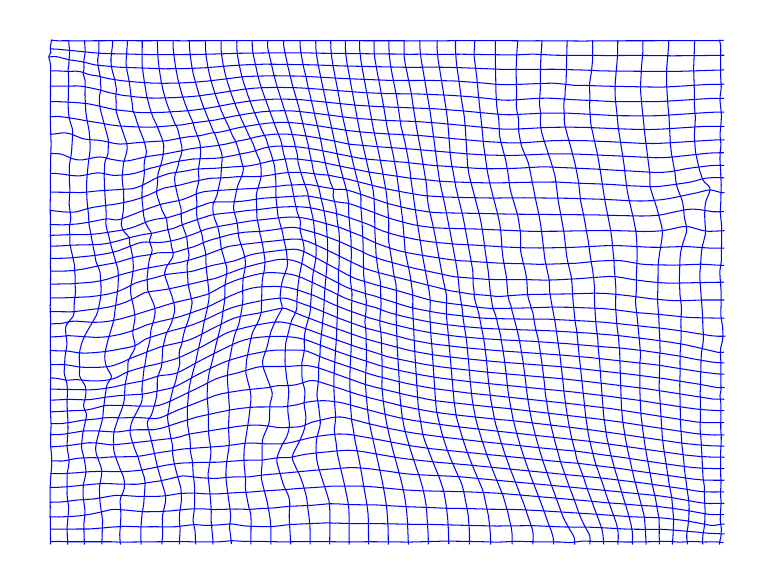}}\hspace{.1em}
        \subfloat[VoxelMorph Deform Field]{\includegraphics[width=.23\textwidth, height=.23\textwidth, trim={1.7em 1.7em 1.7em 1.7em}, clip]{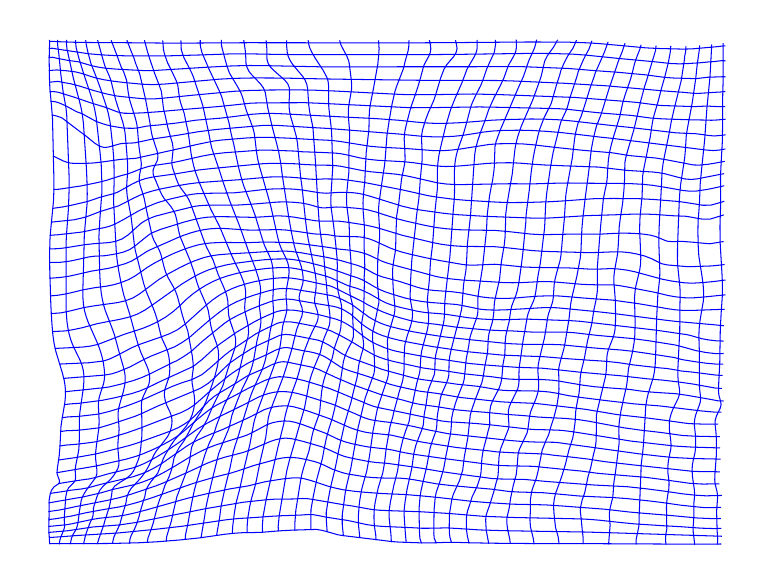}}\hspace{.1em}
        \subfloat[QCRegNet Deform Field]{\includegraphics[width=.23\textwidth, height=.23\textwidth, trim={1.7em 1.7em 1.7em 1.7em}, clip]{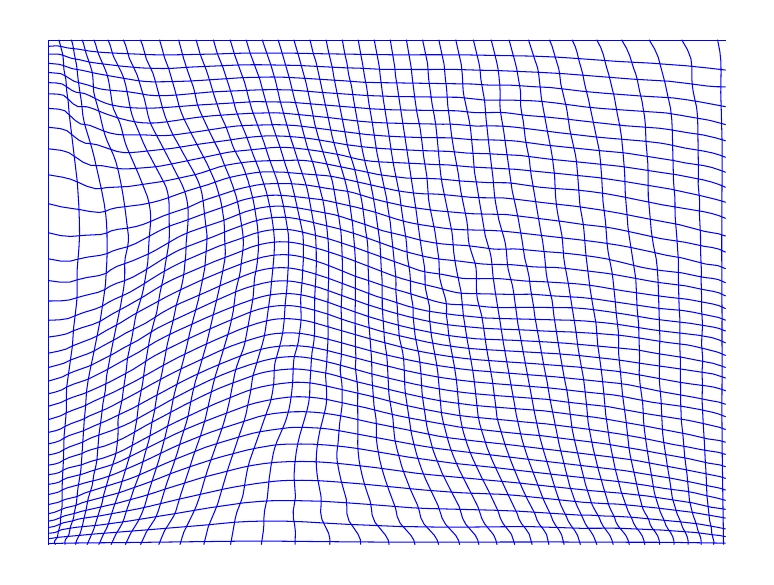}}\\
        \caption{Registration results of underwater images capturing a moving fish by Hyperelastic model, VoxelMorph \& QCRegNet.}\label{fig:1}
        
        \vspace{.5em}
        \subfloat[Fixed]{\includegraphics[width=.23\textwidth]{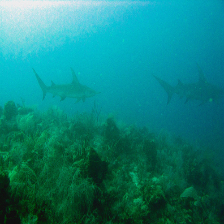}}\hspace{.1em}
        \subfloat[Hyperelastic]{\includegraphics[width=.23\textwidth]{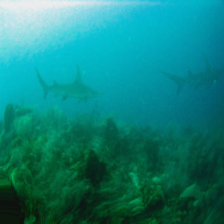}}\hspace{.1em}
        \subfloat[VoxelMorph]{\includegraphics[width=.23\textwidth]{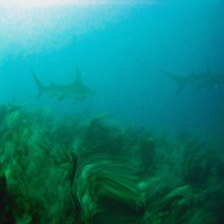}}\hspace{.1em}
        \subfloat[QCRegNet]{\includegraphics[width=.23\textwidth]{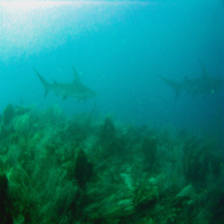}}\hspace{.1em}
    
        \subfloat[Moving]{\includegraphics[width=.23\textwidth]{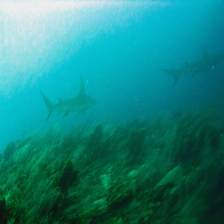}}\hspace{.1em}
        \subfloat[Hyperelastic Deform Field]{\includegraphics[width=.23\textwidth, height=.23\textwidth, trim={1.7em 1.7em 1.7em 1.7em}, clip]{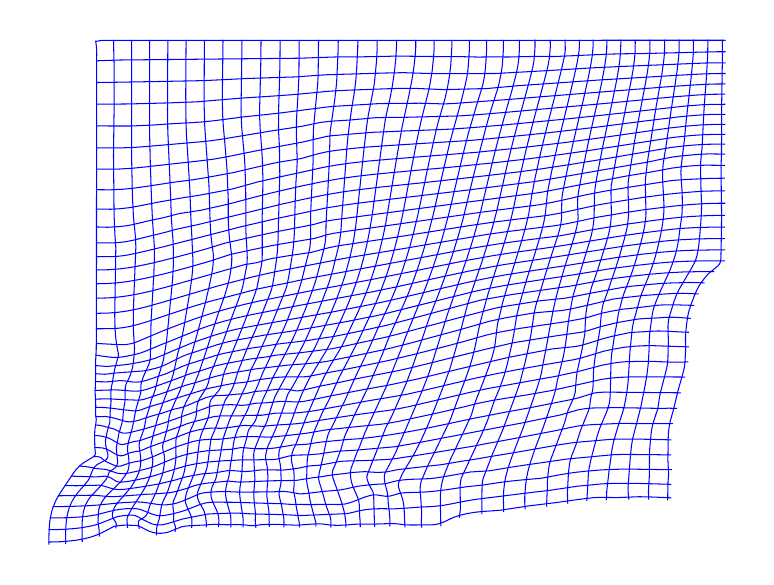}}\hspace{.1em}
        \subfloat[VoxelMorph Deform Field]{\includegraphics[width=.23\textwidth, height=.23\textwidth, trim={1.7em 1.7em 1.7em 1.7em}, clip]{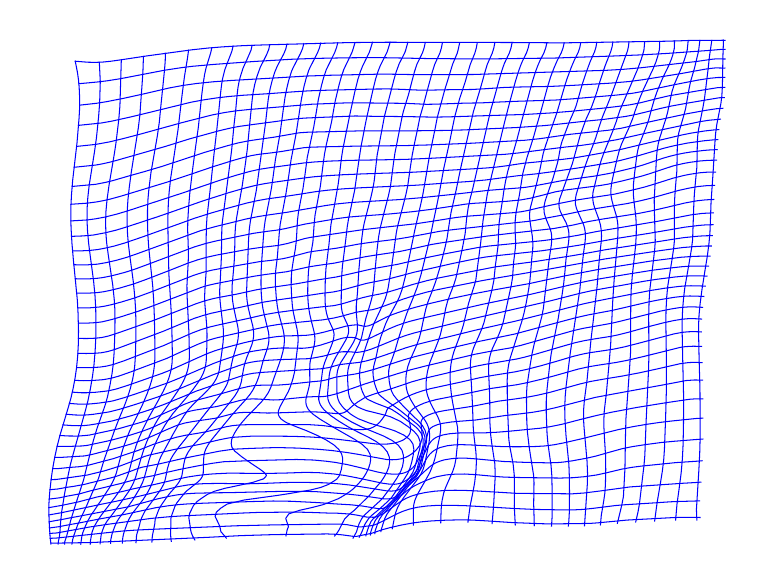}}\hspace{.1em}
        \subfloat[QCRegNet Deform Field]{\includegraphics[width=.23\textwidth, height=.23\textwidth, trim={1.7em 1.7em 1.7em 1.7em}, clip]{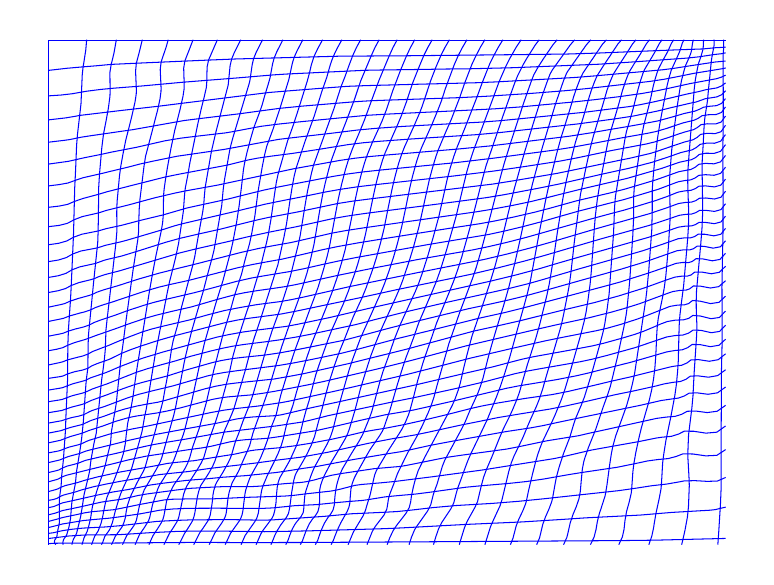}}\\
        \caption{Registration results of underwater images capturing a shark by Hyperelastic model, VoxelMorph \& QCRegNet.}\label{fig:2}
    \end{figure}   
    
    \begin{figure} 
        \centering
        \subfloat[Fixed]{\includegraphics[width=.23\textwidth]{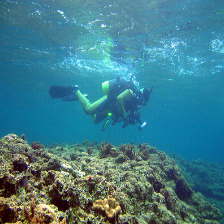}}\hspace{.1em}
        \subfloat[Hyperelastic]{\includegraphics[width=.23\textwidth]{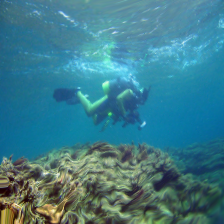}}\hspace{.1em}
        \subfloat[VoxelMorph]{\includegraphics[width=.23\textwidth]{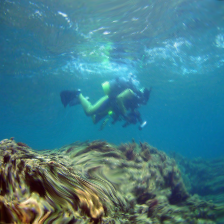}}\hspace{.1em}
        \subfloat[QCRegNet]{\includegraphics[width=.23\textwidth]{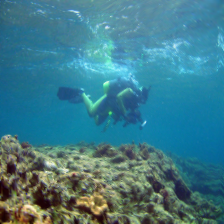}}\hspace{.1em}
    
        \subfloat[Moving]{\includegraphics[width=.23\textwidth]{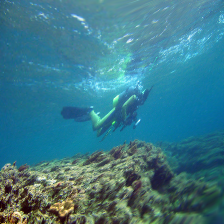}}\hspace{.1em}
        \subfloat[Hyperelastic Deform Field]{\includegraphics[width=.23\textwidth, height=.23\textwidth, trim={1.7em 1.7em 1.7em 1.7em}, clip]{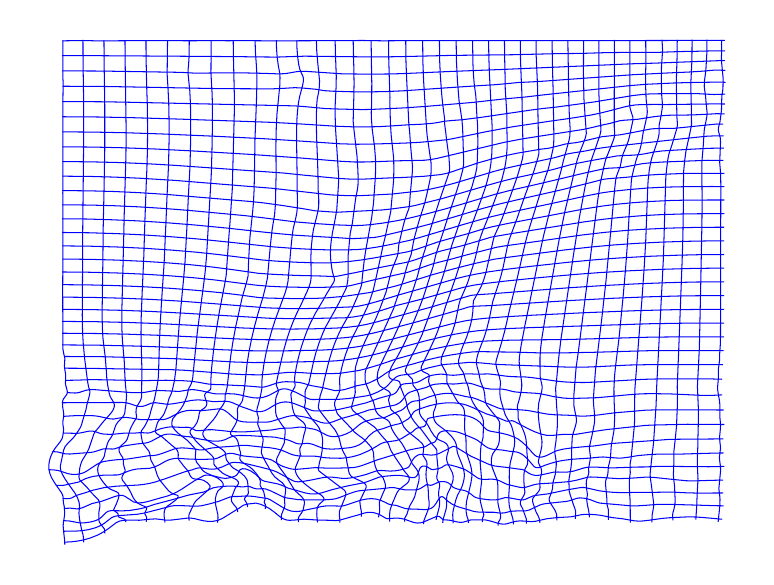}}\hspace{.1em}
        \subfloat[VoxelMorph Deform Field]{\includegraphics[width=.23\textwidth, height=.23\textwidth, trim={1.7em 1.7em 1.7em 1.7em}, clip]{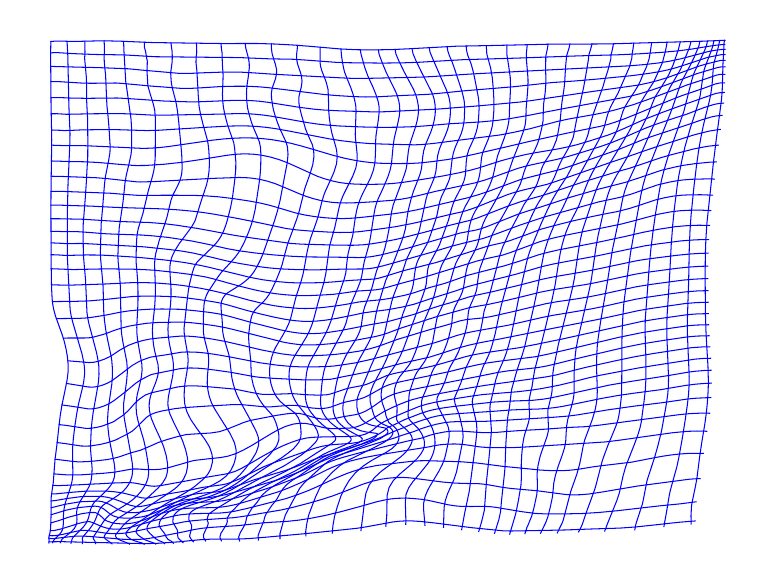}}\hspace{.1em}
        \subfloat[QCRegNet Deform Field]{\includegraphics[width=.23\textwidth, height=.23\textwidth, trim={1.7em 1.7em 1.7em 1.7em}, clip]{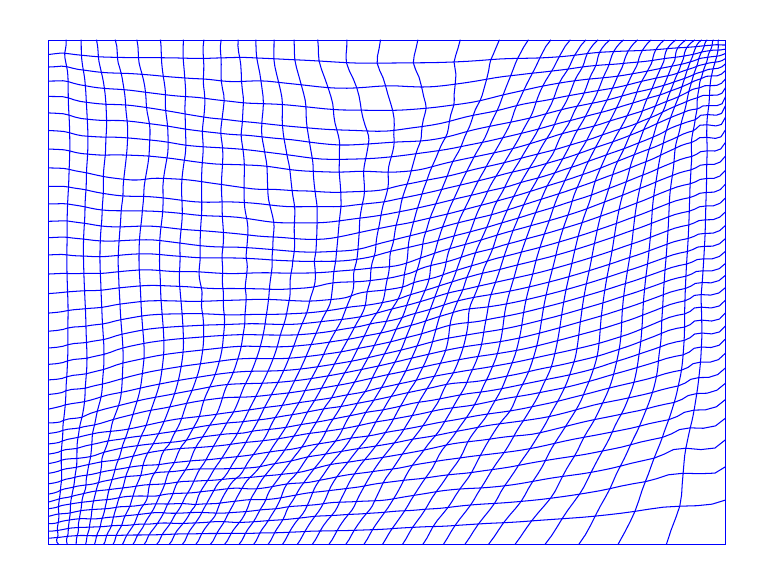}}\\
        \caption{Registration results of underwater images capturing a diver by Hyperelastic model, VoxelMorph \& QCRegNet.}\label{fig:3}
        \vspace{.5em}
        
        \subfloat[Fixed]{\includegraphics[width=.23\textwidth]{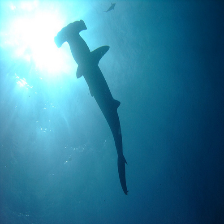}}\hspace{.1em}
        \subfloat[Hyperelastic]{\includegraphics[width=.23\textwidth]{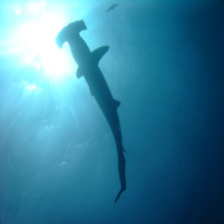}}\hspace{.1em}
        \subfloat[VoxelMorph]{\includegraphics[width=.23\textwidth]{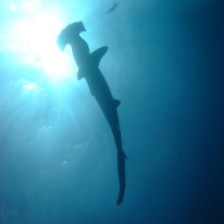}}\hspace{.1em}
        \subfloat[QCRegNet]{\includegraphics[width=.23\textwidth]{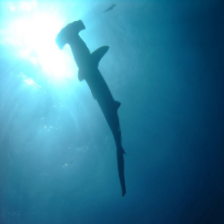}}\hspace{.1em}
    
        \subfloat[Moving]{\includegraphics[width=.23\textwidth]{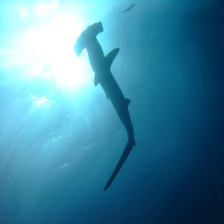}}\hspace{.1em}
        \subfloat[Hyperelastic Deform Field]{\includegraphics[width=.23\textwidth, height=.23\textwidth, trim={1.7em 1.7em 1.7em 1.7em}, clip]{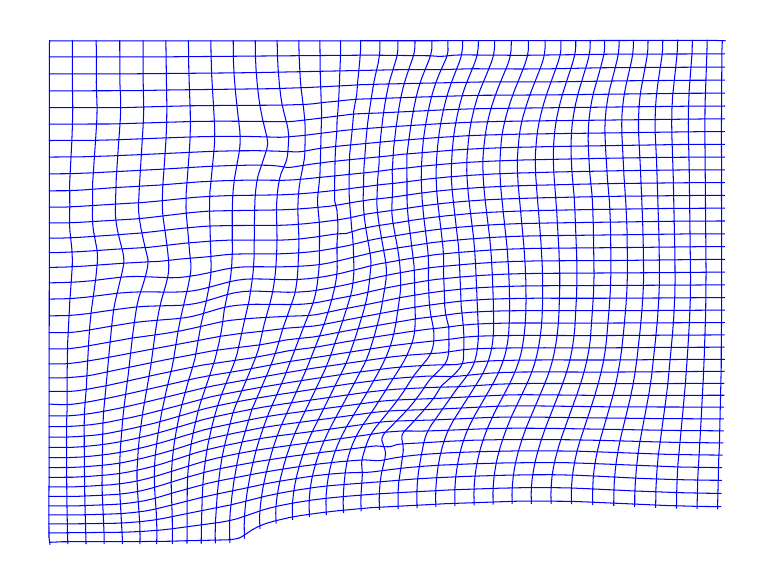}}\hspace{.1em}
        \subfloat[VoxelMorph Deform Field]{\includegraphics[width=.23\textwidth, height=.23\textwidth, trim={1.7em 1.7em 1.7em 1.7em}, clip]{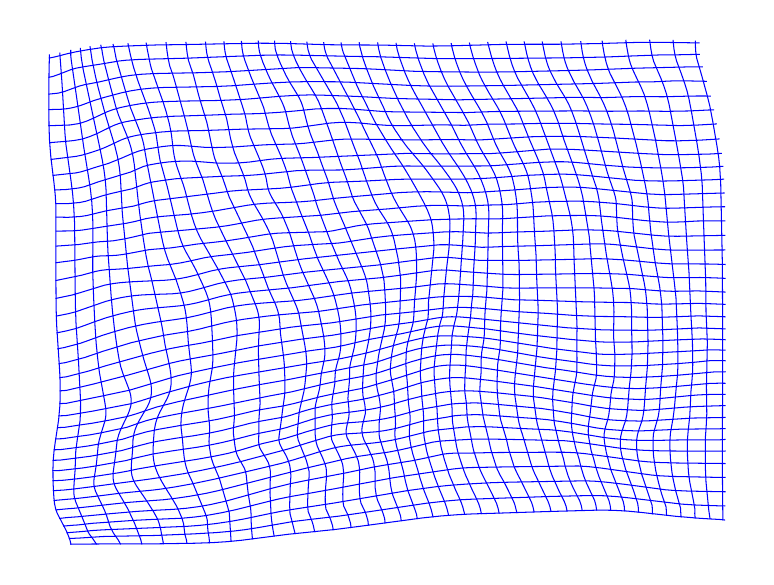}}\hspace{.1em}
        \subfloat[QCRegNet Deform Field]{\includegraphics[width=.23\textwidth, height=.23\textwidth, trim={1.7em 1.7em 1.7em 1.7em}, clip]{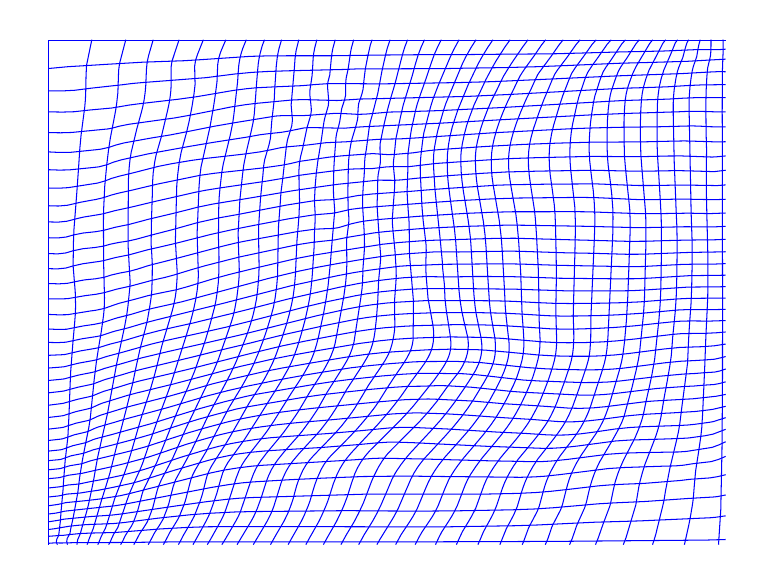}}\\
        \caption{Registration results of underwater images capturing another shark by Hyperelastic model, VoxelMorph \& QCRegNet.}\label{fig:4}
        
    \end{figure}
    
     \begin{figure} 
        \centering
        \subfloat[Fixed]{\includegraphics[width=.23\textwidth]{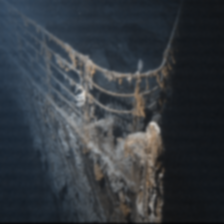}}\hspace{.1em}
        \subfloat[Hyperelastic]{\includegraphics[width=.23\textwidth]{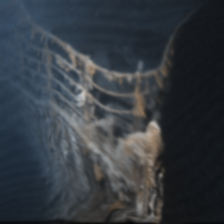}}\hspace{.1em}
        \subfloat[VoxelMorph]{\includegraphics[width=.23\textwidth]{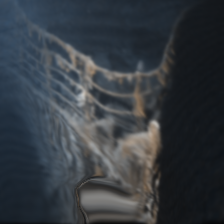}}\hspace{.1em}
        \subfloat[QCRegNet]{\includegraphics[width=.23\textwidth]{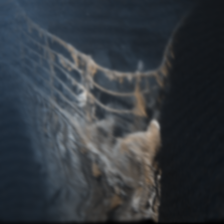}}\hspace{.1em}
    
        \subfloat[Moving]{\includegraphics[width=.23\textwidth]{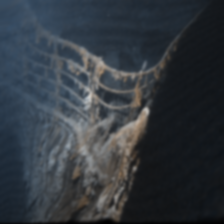}}\hspace{.1em}
        \subfloat[Hyperelastic Deform Field]{\includegraphics[width=.23\textwidth, height=.23\textwidth, trim={1.7em 1.7em 1.7em 1.7em}, clip]{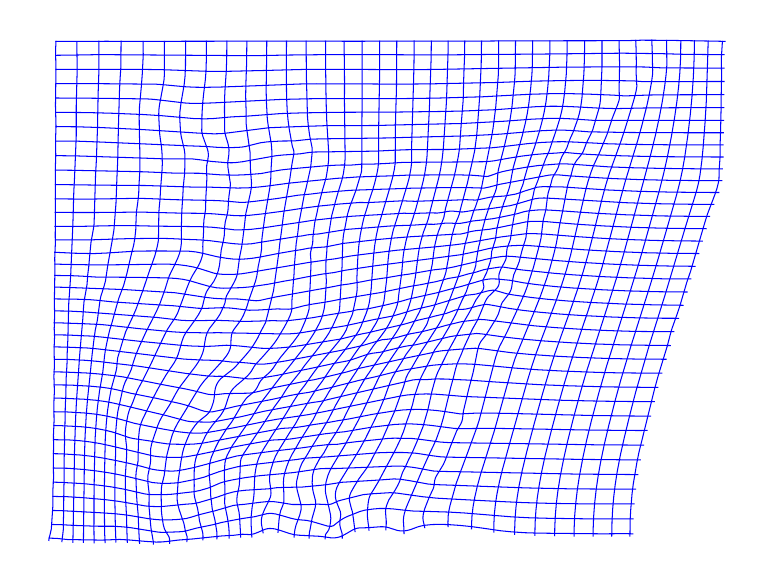}}\hspace{.1em}
        \subfloat[VoxelMorph Deform Field]{\includegraphics[width=.23\textwidth, height=.23\textwidth, trim={1.7em 1.7em 1.7em 1.7em}, clip]{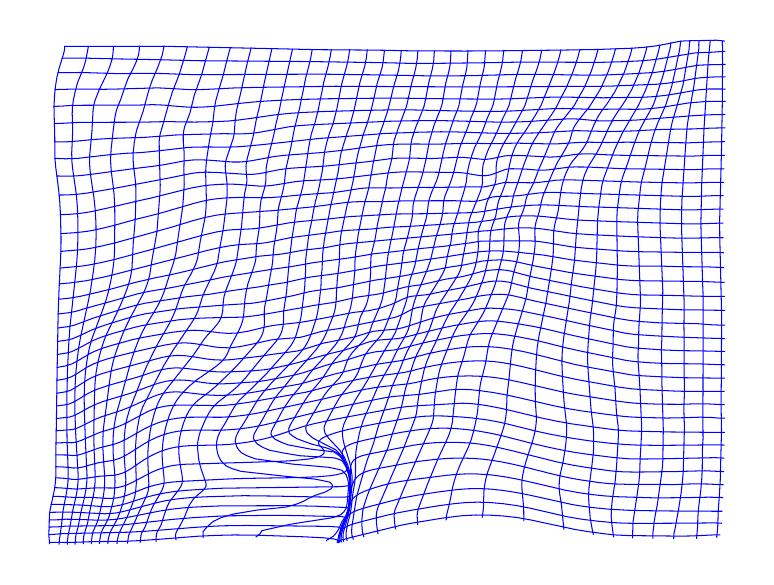}}\hspace{.1em}
        \subfloat[QCRegNet Deform Field]{\includegraphics[width=.23\textwidth, height=.23\textwidth, trim={1.7em 1.7em 1.7em 1.7em}, clip]{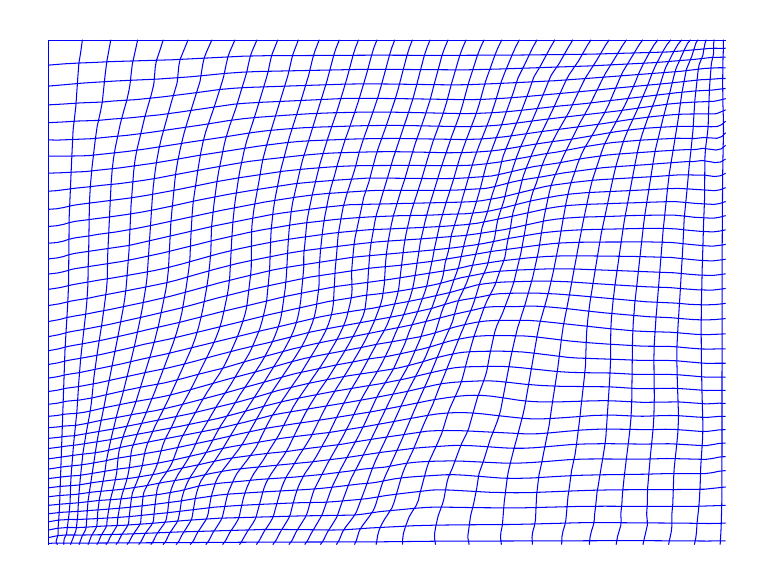}}\\
        \caption{Registration results of underwater images capturing a ship by Hyperelastic model, VoxelMorph \& QCRegNet.}\label{fig:5}
        
        \vspace{.5em}
        \subfloat[Fixed]{\includegraphics[width=.23\textwidth]{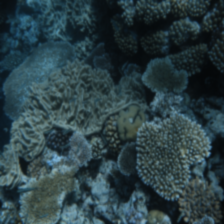}}\hspace{.1em}
        \subfloat[Hyperelastic]{\includegraphics[width=.23\textwidth]{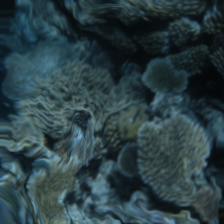}}\hspace{.1em}
        \subfloat[VoxelMorph]{\includegraphics[width=.23\textwidth]{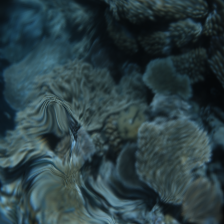}}\hspace{.1em}
        \subfloat[QCRegNet]{\includegraphics[width=.23\textwidth]{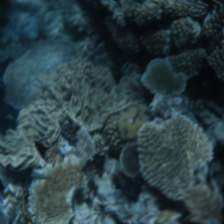}}\hspace{.1em}
    
        \subfloat[Moving]{\includegraphics[width=.23\textwidth]{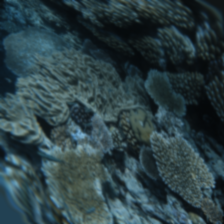}}\hspace{.1em}
        \subfloat[Hyperelastic Deform Field]{\includegraphics[width=.23\textwidth, height=.23\textwidth, trim={1.7em 1.7em 1.7em 1.7em}, clip]{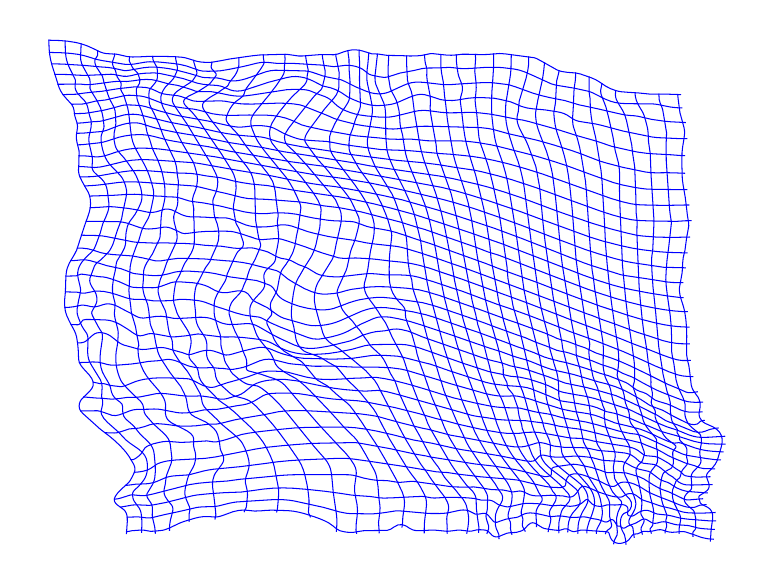}}\hspace{.1em}
        \subfloat[VoxelMorph Deform Field]{\includegraphics[width=.23\textwidth, height=.23\textwidth, trim={1.7em 1.7em 1.7em 1.7em}, clip]{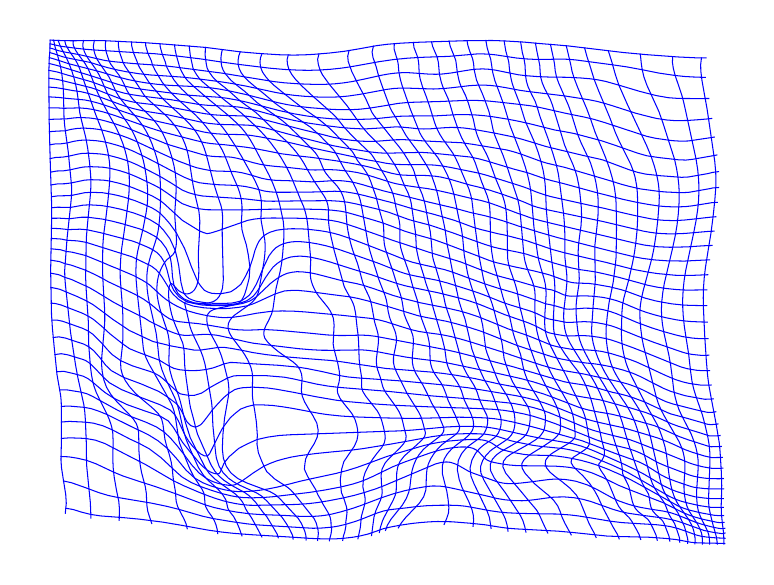}}\hspace{.1em}
        \subfloat[QCRegNet Deform Field]{\includegraphics[width=.23\textwidth, height=.23\textwidth, trim={1.7em 1.7em 1.7em 1.7em}, clip]{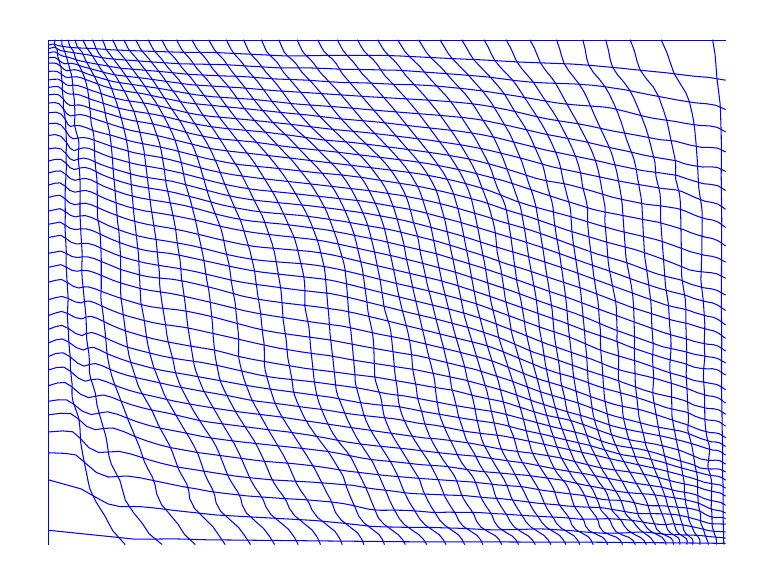}}\\
        \caption{Registration results of underwater images capturing the coral by Hyperelastic model, VoxelMorph \& QCRegNet.}\label{fig:6}   
    \end{figure}

\cref{fig:1} - \cref{fig:6} show the registration results of numerous underwater images obtained by different state-of-the-art image registration models. Besides QCRegNet, we also compute the registration results using the deep-learning based model VoxelMorph \cite{balakrishnan2019voxelmorph} and the optimization based hyperelastic model \cite{burger2013hyperelastic}. In each figure, the fixed and moving images are shown in (a) and (e) respectively. The moving image is to be deformed to the fixed image via the registration map. (b) shows the result obtained by the hyperelastic model. (c) shows the result obtained by the VoxelMorph. (d) shows the result obtained by our proposed QCRegNet. 

It can be easily observed from \cref{fig:1} - \cref{fig:6}  that the registration results obtained by the hyperelastic model are not accurate since some essential structures in the moving image cannot be matched to the corresponding structures in the fixed image. It can be visualized by comparing the registered image with the fixed image. The registration maps are plotted in (f), which are visualized as the deformed grid from the regular finite difference grid on the image domain. On the other hand, VoxelMorph can give a rough registration between the moving and fixed image. However, inaccuracies can be evidently observed by comparing the registered and the fixed images. Also, the registration maps by VoxelMorph as shown in (g) are non-bijective and foldings can be seen at multiple positions. The occurrence of foldings is more severe in those cases when images are blurry and largely deformed. On the contrary, the registration results obtained by QCRegNet outperform the other two methods. The registered images closely resemble the fixed images. The registration maps as shown in (h) are always bijective, which is one of the key features of our proposed framework. 

\cref{table_nj_sj} shows $N_J$ and $S_J$ of the registration maps obtained by different methods. The quantitative measurements demonstrate the effectiveness of our proposed model over the other methods. In particular, QCRegNet shows satisfactory registration results without foldings ($N_J=0$ and $S_J=0$), even for the very challenging scenarios when blurry images with large geometric distortions are considered.  

\begin{table}
    \centering
    \begin{tabular}{@{\extracolsep{1em}} c  *{3}{>{\centering\arraybackslash}p{1cm}} >{\centering\arraybackslash}p{2.3cm} *{2}{>{\centering\arraybackslash}p{1cm}}}
    \toprule
         & \multicolumn{2}{c}{Hyperelastic} & \multicolumn{2}{c}{VoxelMorph} & \multicolumn{2}{c}{QCRegNet}\\
         \cmidrule{2-3} \cmidrule{4-5} \cmidrule{6-7}
         & $N_J$ & $S_J$ & $N_J$ & $S_J$ & $N_J$ & $S_J$\\
         \midrule
         \cref{fig:1} & 0 & 0 & 0 & 0 & 0 & 0 \\
         \cref{fig:2} & 0 & 0 & 2 & 479.7763 & 0 & 0 \\
         \cref{fig:3} & 30 & 1.6608 & 186 & 5693054.9218 & 0 & 0 \\
         \cref{fig:4} & 0 & 0 & 0 & 0 & 0 & 0 \\
         \cref{fig:5} & 0 & 0 & 20 & 13066.9024 & 0 & 0 \\
         \cref{fig:6} & 187 & 25.7112 & 7 & 13042.7050 & 0 & 0 \\
         
    \bottomrule
    \end{tabular}
    \caption{Comparison of $N_J$ and $S_J$ of mappings generated by different algorithms}
    \label{table_nj_sj}
\end{table}

    \subsubsection{Medical Image Registration} 
        \begin{figure}[h]
        \centering
        
        \begin{tabular}{m{0.22\textwidth}m{0.22\textwidth}m{0.22\textwidth}m{0.22\textwidth}}
        
        \includegraphics[width=0.23\textwidth]{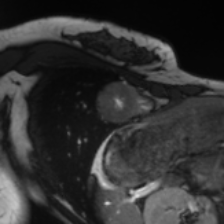}\vspace{.2em}
        \includegraphics[width=0.23\textwidth]{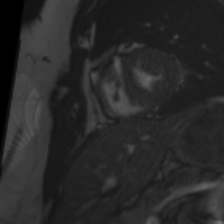}\vspace{.2em}
        \includegraphics[width=0.23\textwidth]{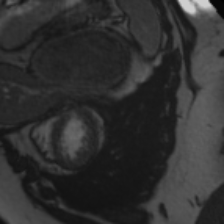}\vspace{.2em}
        \includegraphics[width=0.23\textwidth]{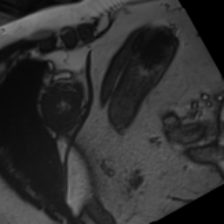}
        \subcaption{Fixed}
        &
        
        \includegraphics[width=0.23\textwidth]{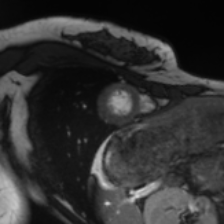}\vspace{.2em}
        \includegraphics[width=0.23\textwidth]{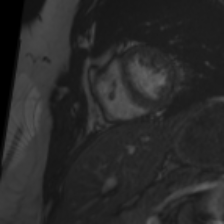}\vspace{.2em}
        \includegraphics[width=0.23\textwidth]{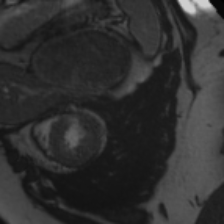}\vspace{.2em}
        \includegraphics[width=0.23\textwidth]{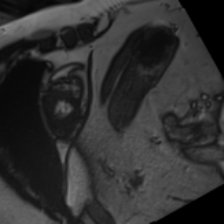}
        \subcaption{Moving}
        &
        
        \includegraphics[width=0.23\textwidth]{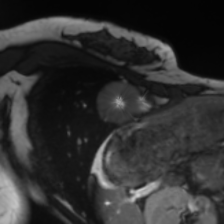}\vspace{.2em}
        \includegraphics[width=0.23\textwidth]{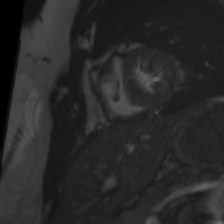}\vspace{.2em}
        \includegraphics[width=0.23\textwidth]{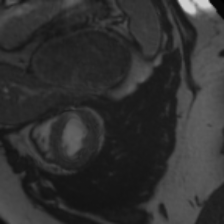}\vspace{.2em}
        \includegraphics[width=0.23\textwidth]{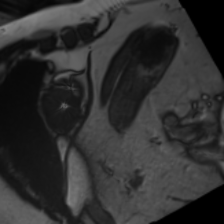}
        \subcaption{QCRegNet}
        &
        
        \includegraphics[width=0.23\textwidth, height=0.23\textwidth, trim={1.7em 1.7em 1.7em 1.7em}, clip]{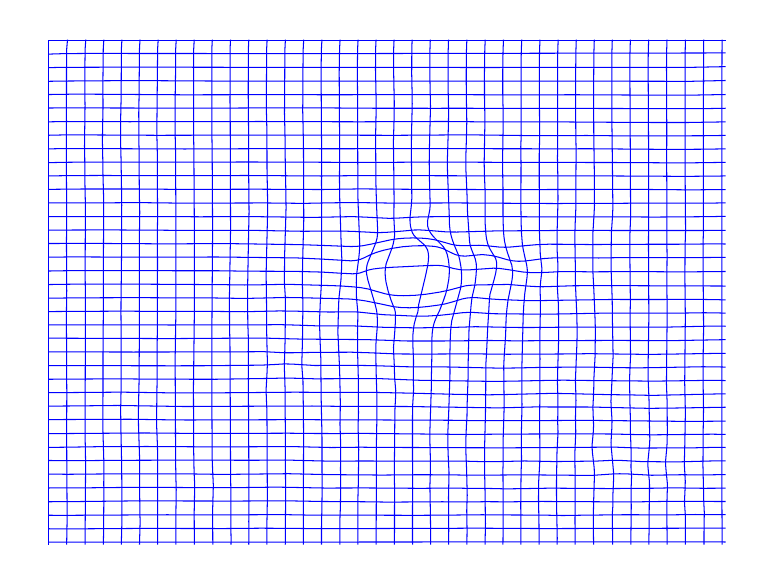}\vspace{.2em}
        \includegraphics[width=0.23\textwidth, height=0.23\textwidth, trim={1.7em 1.7em 1.7em 1.7em}, clip]{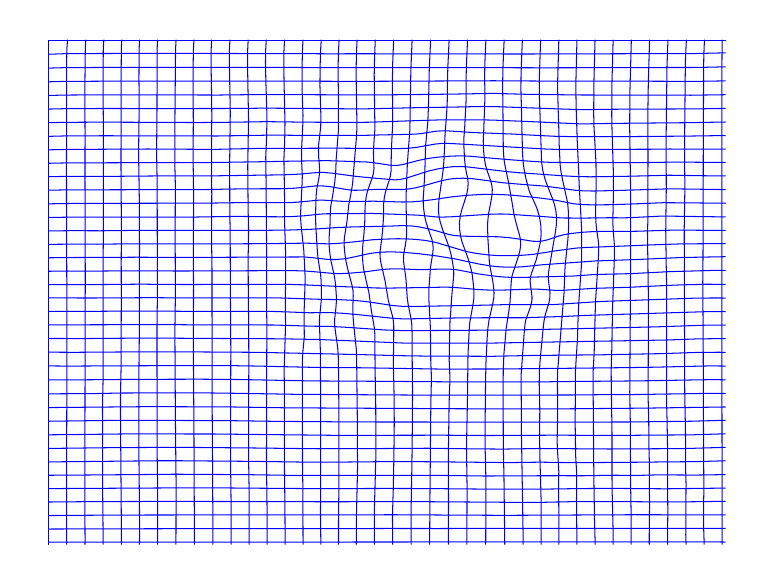}\vspace{.2em}
        \includegraphics[width=0.23\textwidth, height=0.23\textwidth, trim={1.7em 1.7em 1.7em 1.7em}, clip]{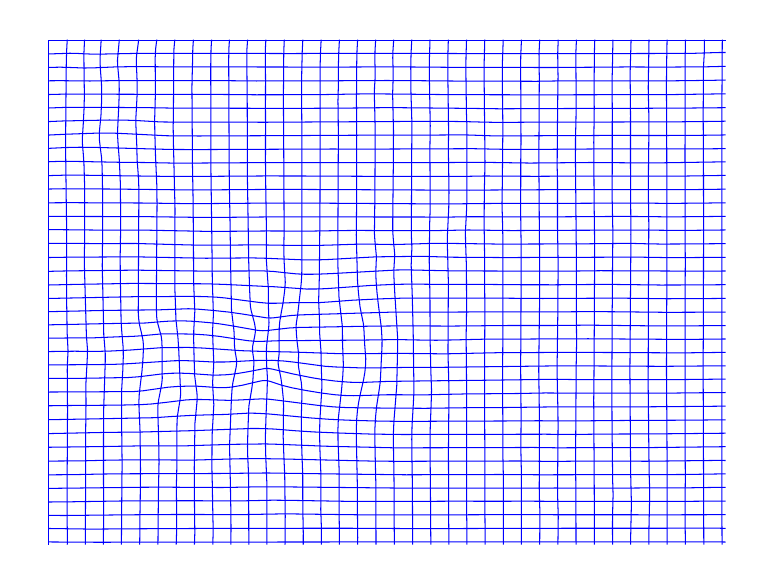}\vspace{.2em}
        \includegraphics[width=0.23\textwidth, height=0.23\textwidth, trim={1.7em 1.7em 1.7em 1.7em}, clip]{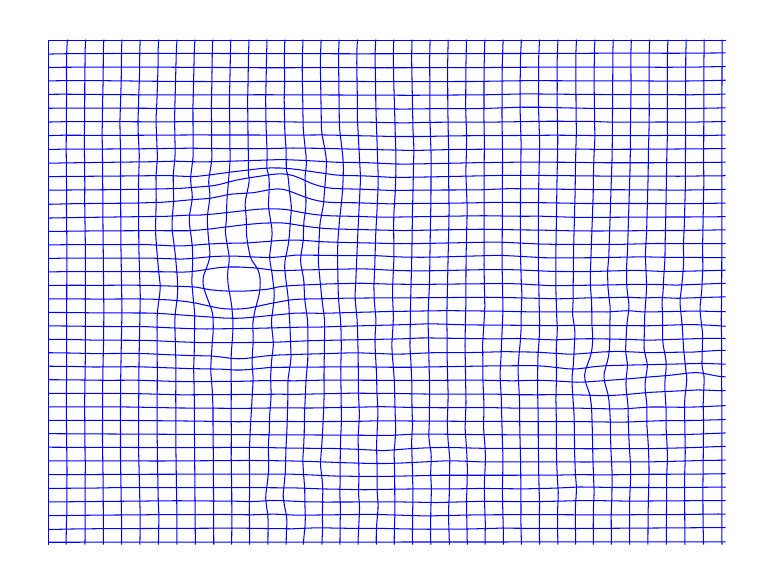}
        \subcaption{QCRegNet}
        
        \end{tabular}

        \caption{Registration results of the cardiac images using the QCRegNet.}
        \label{fig:medreg}
        
    \end{figure}
 
    We also apply the proposed QCRegNet for medical image registration. Medical image registration plays an increasingly important role in many clinical applications ranging from computer assisted diagnosis to computer aided therapy and surgery. Accurate and efficient registration of medical images is of the utmost importance. In this subsection, we examine the effectiveness of our QCRegNet on the image registration problem of cardiac images. 
    
    The cardiac images are extracted from the ACDC dataset \cite{bernard2018deep}. These are MRI data acquired from real clinical examination. Each scan records a patient's heartbeat, including the diastolic and systolic phases. In our experiments, 100 cardiac MRI recordings are used as the training data and 50 are used as the testing data. Data augmentation is carried out to obtain more data to train our model. The procedure of data augmentation is accomplished by randomly selecting frame pairs, and rotating and cutting patches from them. 
    
    \cref{fig:medreg} shows the image registration results of some cardiac image pairs. The fixed cardiac images are shown in the first column (a). The moving images are shown in the second column (b). The registered images using the QCRegNet are shown in the third column (c). It is observed that the registered images closely resemble the fixed target images, demonstrating the effectiveness of the QCRegNet. The registration maps are shown in the last column (d). As shown in the figures, all registration maps are bijective without unnatural foldings.

\subsubsection{Convergence and inference Time of QCRegNet}

One of the superiority of our model against traditional methods is that compared with traditional methods, the inference time of our model is trivial. To see this, we apply the three methods to some image pairs, and compute the mean and standard deviation of inference time. This experiment is carried out on a computer with Intel i7-10700 CPU, and there is no GPU acceleration on this machine, which ensures the fairness of the inference time comparison. The results are shown in \cref{table_infer_time}.

\begin{table}[h!]
\centering
\begin{tabular}{ c|c|c|c } 
 \hline
       & Hyperelastic & VoxelMorph & QCRegNet \\ 
\hline
 mean(s) & 30.1764 & 0.0251 & 0.0559 \\ 
 std(s) & 19.0847 & 0.0027 & 0.0035 \\ 
 \hline
\end{tabular}
\caption{Inference time of Hyperelastic model, VoxelMorph, and QCRegNet}
\label{table_infer_time}
\end{table}

From \cref{table_infer_time}, we notice that hyperelastic registration model takes much more time to solve the registration problem than deep learning based methods. It is expected as the hyperelastic registration model requires solving an optimization problem for each image pair. The standard deviation of the computational time for hyperelastic registration model is also much larger than that of the two deep learning based methods. The main reason is that for different image pairs, the difficulty of searching for an acceptable solution varies. However, for deep learning methods, once the network is successfully trained, the computational cost for every image pair is fixed and the registration can be done in real time. From the table, it is observed that the computational time for QCRegNet is comparable to VoxelMorph, while QCRegNet can often obtain more satisfactory registration results in challenging scenarios. 

\begin{figure*}[t]
	\centering
	\includegraphics[width=3in]{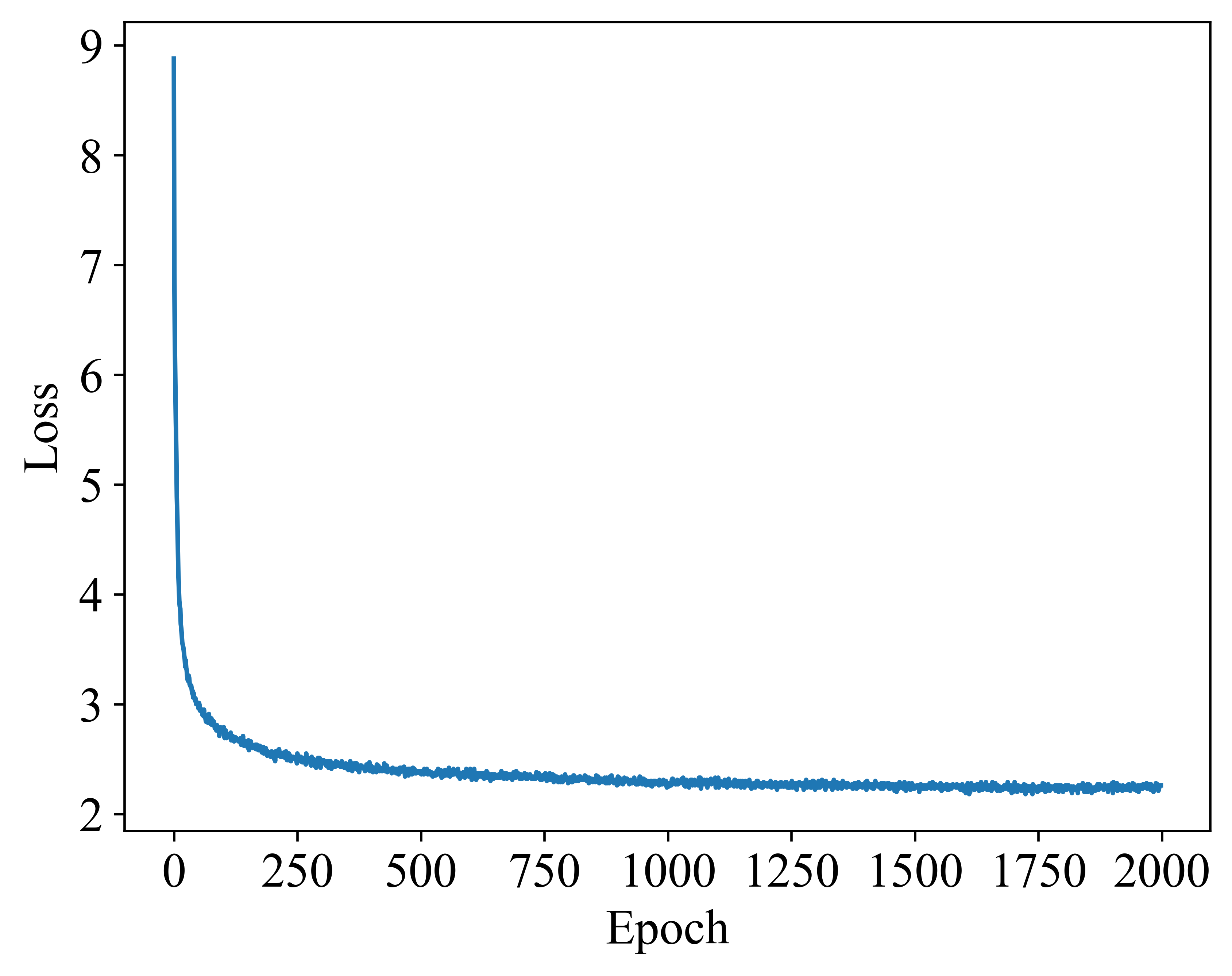}
	\caption{Training loss of QCRegNet}
	\label{qcregnet_loss} 
\end{figure*}

The plot of the training loss is shown in \cref{qcregnet_loss}, from which we can be confident that the model has successfully converged during training.

\section{Conclusion}\label{sec:conclusion}
In this work, we proposed a learning based framework for solving mapping problems. QC Teichm\"uller theories are integrated into the framework to control the geometric properties of the mappings. The main idea of our proposed framework is to train a deep neural network to learn the Beltrami coefficient (BC) as a latent representation, which represents the mapping and describes the geometric distortions or quasiconformality under the mapping. The integration of quasiconformality into the deep neural network enhances the interpretability of the network. By constraining the magnitude of the BC feature, the geometric distortion and bijectivity of the mapping can easily be controlled. As such, the diffeomorphism optimization problem (DOP) can be easily solved using our proposed framework. In order to incorporate the BC into the network, the Beltrami Solver Network (BSNet) is developed in this work. The BSNet computes the corresponding quasiconformal map associated to a prescribed BC. We examine the efficacy of this framework to the diffeomorphic image registration problem. The Quasiconformal Registration Network (QCRegNet) is trained. The experimental results show the feasibility and effectiveness of the framework. The registration results outperform other state-of-the-art approaches, even in challenging cases.

\bibliographystyle{siam_template/siamplain}
\bibliography{mylib}

\begin{thebibliography}{10}

\bibitem{AshburnerJohn2007Afdi}
{\sc J.~Ashburner}, {\em A fast diffeomorphic image registration algorithm},
  NeuroImage (Orlando, Fla.), 38 (2007), pp.~95--113.

\bibitem{balakrishnan2019voxelmorph}
{\sc G.~Balakrishnan, A.~Zhao, M.~R. Sabuncu, J.~Guttag, and A.~V. Dalca}, {\em
  Voxelmorph: a learning framework for deformable medical image registration},
  IEEE transactions on medical imaging, 38 (2019), pp.~1788--1800.

\bibitem{beg2005computing}
{\sc M.~F. Beg, M.~I. Miller, A.~Trouv{\'e}, and L.~Younes}, {\em Computing
  large deformation metric mappings via geodesic flows of diffeomorphisms},
  International journal of computer vision, 61 (2005), pp.~139--157.

\bibitem{bernard2018deep}
{\sc O.~Bernard, A.~Lalande, C.~Zotti, F.~Cervenansky, X.~Yang, P.-A. Heng,
  I.~Cetin, K.~Lekadir, O.~Camara, M.~A.~G. Ballester, et~al.}, {\em Deep
  learning techniques for automatic mri cardiac multi-structures segmentation
  and diagnosis: is the problem solved?}, IEEE transactions on medical imaging,
  37 (2018), pp.~2514--2525.

\bibitem{burger2013hyperelastic}
{\sc M.~Burger, J.~Modersitzki, and L.~Ruthotto}, {\em A hyperelastic
  regularization energy for image registration}, SIAM Journal on Scientific
  Computing, 35 (2013), pp.~B132--B148.

\bibitem{chak2018subsampled}
{\sc W.~H. Chak, C.~P. Lau, and L.~M. Lui}, {\em Subsampled turbulence removal
  network}, arXiv preprint arXiv:1807.04418,  (2018).

\bibitem{CHAN2016177}
{\sc H.~L. Chan, H.~Li, and L.~M. Lui}, {\em Quasi-conformal statistical shape
  analysis of hippocampal surfaces for alzheimer's disease analysis},
  Neurocomputing, 175 (2016), pp.~177--187,
  \url{https://doi.org/https://doi.org/10.1016/j.neucom.2015.10.047},
  \url{https://www.sciencedirect.com/science/article/pii/S0925231215015064}.

\bibitem{chan2018topology}
{\sc H.-L. Chan, S.~Yan, L.-M. Lui, and X.-C. Tai}, {\em Topology-preserving
  image segmentation by beltrami representation of shapes}, Journal of
  Mathematical Imaging and Vision, 60 (2018), pp.~401--421.

\bibitem{chan2020quasi}
{\sc H.-L. Chan, H.-M. Yuen, C.-T. Au, K.~C.-C. Chan, A.~M. Li, and L.-M. Lui},
  {\em Quasi-conformal geometry based local deformation analysis of lateral
  cephalogram for childhood osa classification}, arXiv preprint
  arXiv:2006.11408,  (2020).

\bibitem{chen2019image}
{\sc K.~Chen, L.~M. Lui, and J.~Modersitzki}, {\em Image and surface
  registration}, in Handbook of Numerical Analysis, vol.~20, Elsevier, 2019,
  pp.~579--611.

\bibitem{choi2020tooth}
{\sc G.~P. Choi, H.~L. Chan, R.~Yong, S.~Ranjitkar, A.~Brook, G.~Townsend,
  K.~Chen, and L.~M. Lui}, {\em Tooth morphometry using quasi-conformal
  theory}, Pattern Recognition, 99 (2020), p.~107064.

\bibitem{choi2020shape}
{\sc G.~P. Choi, D.~Qiu, and L.~M. Lui}, {\em Shape analysis via inconsistent
  surface registration}, Proceedings of the Royal Society A, 476 (2020),
  p.~20200147.

\bibitem{choi2015flash}
{\sc P.~T. Choi, K.~C. Lam, and L.~M. Lui}, {\em Flash: Fast landmark aligned
  spherical harmonic parameterization for genus-0 closed brain surfaces}, SIAM
  Journal on Imaging Sciences, 8 (2015), pp.~67--94.

\bibitem{de2019deep}
{\sc B.~D. de~Vos, F.~F. Berendsen, M.~A. Viergever, H.~Sokooti, M.~Staring,
  and I.~I{\v{s}}gum}, {\em A deep learning framework for unsupervised affine
  and deformable image registration}, Medical image analysis, 52 (2019),
  pp.~128--143.

\bibitem{glocker2011deformable}
{\sc B.~Glocker, A.~Sotiras, N.~Komodakis, and N.~Paragios}, {\em Deformable
  medical image registration: setting the state of the art with discrete
  methods}, Annual review of biomedical engineering, 13 (2011), pp.~219--244.

\bibitem{guo2016deep}
{\sc Y.~Guo, Y.~Liu, A.~Oerlemans, S.~Lao, S.~Wu, and M.~S. Lew}, {\em Deep
  learning for visual understanding: A review}, Neurocomputing, 187 (2016),
  pp.~27--48.

\bibitem{horn1981determining}
{\sc B.~K. Horn and B.~G. Schunck}, {\em Determining optical flow}, Artificial
  intelligence, 17 (1981), pp.~185--203.

\bibitem{ibrahim2016improved}
{\sc M.~Ibrahim, K.~Chen, and L.~Rada}, {\em An improved model for joint
  segmentation and registration based on linear curvature smoother}, Journal of
  Algorithms \& Computational Technology, 10 (2016), pp.~314--324.

\bibitem{NIPS2015_33ceb07b}
{\sc M.~Jaderberg, K.~Simonyan, A.~Zisserman, and k.~kavukcuoglu}, {\em Spatial
  transformer networks}, in Advances in Neural Information Processing Systems,
  C.~Cortes, N.~Lawrence, D.~Lee, M.~Sugiyama, and R.~Garnett, eds., vol.~28,
  Curran Associates, Inc., 2015,
  \url{https://proceedings.neurips.cc/paper/2015/file/33ceb07bf4eeb3da587e268d663aba1a-Paper.pdf}.

\bibitem{joshi2000landmark}
{\sc S.~C. Joshi and M.~I. Miller}, {\em Landmark matching via large
  deformation diffeomorphisms}, IEEE transactions on image processing, 9
  (2000), pp.~1357--1370.

\bibitem{lam2014genus}
{\sc K.~C. Lam, X.~Gu, and L.~M. Lui}, {\em Genus-one surface registration via
  teichm{\"u}ller extremal mapping}, in International Conference on Medical
  Image Computing and Computer-Assisted Intervention, Springer, 2014,
  pp.~25--32.

\bibitem{lam2015landmark}
{\sc K.~C. Lam, X.~Gu, and L.~M. Lui}, {\em Landmark constrained genus-one
  surface teichm{\"u}ller map applied to surface registration in medical
  imaging}, Medical image analysis, 25 (2015), pp.~45--55.

\bibitem{lam2014landmark}
{\sc K.~C. Lam and L.~M. Lui}, {\em Landmark-and intensity-based registration
  with large deformations via quasi-conformal maps}, SIAM Journal on Imaging
  Sciences, 7 (2014), pp.~2364--2392.

\bibitem{lam2015quasi}
{\sc K.~C. Lam and L.~M. Lui}, {\em Quasi-conformal hybrid multi-modality image
  registration and its application to medical image fusion}, in International
  Symposium on Visual Computing, Springer, 2015, pp.~809--818.

\bibitem{law2022quasiconformal}
{\sc H.~Law, G.~P.~T. Choi, K.~C. Lam, and L.~M. Lui}, {\em Quasiconformal
  model with {CNN} features for large deformation image registration}, Inverse
  Prob. Imaging, 16 (2022), pp.~1019--1046.

\bibitem{le2011combined}
{\sc C.~Le~Guyader and L.~A. Vese}, {\em A combined segmentation and
  registration framework with a nonlinear elasticity smoother}, Computer Vision
  and Image Understanding, 115 (2011), pp.~1689--1709.

\bibitem{lu2021learning}
{\sc L.~Lu, P.~Jin, G.~Pang, Z.~Zhang, and G.~E. Karniadakis}, {\em Learning
  nonlinear operators via deeponet based on the universal approximation theorem
  of operators}, Nature Machine Intelligence, 3 (2021), pp.~218--229.

\bibitem{lui2013texture}
{\sc L.~M. Lui, K.~C. Lam, T.~W. Wong, and X.~Gu}, {\em Texture map and video
  compression using beltrami representation}, SIAM Journal on Imaging Sciences,
  6 (2013), pp.~1880--1902.

\bibitem{lui2014teichmuller}
{\sc L.~M. Lui, K.~C. Lam, S.-T. Yau, and X.~Gu}, {\em Teichmuller mapping
  (t-map) and its applications to landmark matching registration}, SIAM Journal
  on Imaging Sciences, 7 (2014), pp.~391--426.

\bibitem{lui2015splitting}
{\sc L.~M. Lui and T.~C. Ng}, {\em A splitting method for diffeomorphism
  optimization problem using beltrami coefficients}, Journal of Scientific
  Computing, 63 (2015), pp.~573--611.

\bibitem{lui2014geometric}
{\sc L.~M. Lui and C.~Wen}, {\em Geometric registration of high-genus
  surfaces}, SIAM Journal on Imaging Sciences, 7 (2014), pp.~337--365.

\bibitem{lui2010compression}
{\sc L.~M. Lui, T.~W. Wong, P.~Thompson, T.~Chan, X.~Gu, and S.-T. Yau}, {\em
  Compression of surface registrations using beltrami coefficients}, in 2010
  IEEE Computer Society Conference on Computer Vision and Pattern Recognition,
  IEEE, 2010, pp.~2839--2846.

\bibitem{lui2010shape}
{\sc L.~M. Lui, T.~W. Wong, P.~Thompson, T.~Chan, X.~Gu, and S.-T. Yau}, {\em
  Shape-based diffeomorphic registration on hippocampal surfaces using beltrami
  holomorphic flow}, in International Conference on Medical Image Computing and
  Computer-Assisted Intervention, Springer, 2010, pp.~323--330.

\bibitem{lui2012optimization}
{\sc L.~M. Lui, T.~W. Wong, W.~Zeng, X.~Gu, P.~M. Thompson, T.~F. Chan, and
  S.-T. Yau}, {\em Optimization of surface registrations using beltrami
  holomorphic flow}, Journal of scientific computing, 50 (2012), pp.~557--585.

\bibitem{lui2013shape}
{\sc L.~M. Lui, W.~Zeng, S.-T. Yau, and X.~Gu}, {\em Shape analysis of planar
  multiply-connected objects using conformal welding}, IEEE transactions on
  pattern analysis and machine intelligence, 36 (2013), pp.~1384--1401.

\bibitem{meng2021structure}
{\sc M.~Meng, Q.~Chen, and J.~Wu}, {\em Structure preservation adversarial
  network for visual domain adaptation}, Information Sciences, 579 (2021),
  pp.~266--280.

\bibitem{meng2016tempo}
{\sc T.~W. Meng, G.~P.-T. Choi, and L.~M. Lui}, {\em Tempo: Feature-endowed
  teichmuller extremal mappings of point clouds}, SIAM Journal on Imaging
  Sciences, 9 (2016), pp.~1922--1962.

\bibitem{NEURIPS2019_9015}
{\sc A.~Paszke, S.~Gross, F.~Massa, A.~Lerer, J.~Bradbury, G.~Chanan,
  T.~Killeen, Z.~Lin, N.~Gimelshein, L.~Antiga, A.~Desmaison, A.~Kopf, E.~Yang,
  Z.~DeVito, M.~Raison, A.~Tejani, S.~Chilamkurthy, B.~Steiner, L.~Fang,
  J.~Bai, and S.~Chintala}, {\em Pytorch: An imperative style, high-performance
  deep learning library}, in Advances in Neural Information Processing Systems
  32, H.~Wallach, H.~Larochelle, A.~Beygelzimer, F.~d\textquotesingle
  Alch\'{e}-Buc, E.~Fox, and R.~Garnett, eds., Curran Associates, Inc., 2019,
  pp.~8024--8035,
  \url{http://papers.neurips.cc/paper/9015-pytorch-an-imperative-style-high-performance-deep-learning-library.pdf}.

\bibitem{qiu2019computing}
{\sc D.~Qiu, K.-C. Lam, and L.-M. Lui}, {\em Computing quasi-conformal folds},
  SIAM Journal on Imaging Sciences, 12 (2019), pp.~1392--1424.

\bibitem{qiu2020inconsistent}
{\sc D.~Qiu and L.~M. Lui}, {\em Inconsistent surface registration via
  optimization of mapping distortions}, Journal of Scientific Computing, 83
  (2020), pp.~1--31.

\bibitem{raissi2019physics}
{\sc M.~Raissi, P.~Perdikaris, and G.~E. Karniadakis}, {\em Physics-informed
  neural networks: A deep learning framework for solving forward and inverse
  problems involving nonlinear partial differential equations}, Journal of
  Computational physics, 378 (2019), pp.~686--707.

\bibitem{unet_bib}
{\sc O.~Ronneberger, P.~Fischer, and T.~Brox}, {\em U-net: Convolutional
  networks for biomedical image segmentation}, in Medical Image Computing and
  Computer-Assisted Intervention -- MICCAI 2015, N.~Navab, J.~Hornegger, W.~M.
  Wells, and A.~F. Frangi, eds., Cham, 2015, Springer International Publishing,
  pp.~234--241.

\bibitem{ILSVRC15}
{\sc O.~Russakovsky, J.~Deng, H.~Su, J.~Krause, S.~Satheesh, S.~Ma, Z.~Huang,
  A.~Karpathy, A.~Khosla, M.~Bernstein, A.~C. Berg, and L.~Fei-Fei}, {\em
  {ImageNet Large Scale Visual Recognition Challenge}}, International Journal
  of Computer Vision (IJCV), 115 (2015), pp.~211--252,
  \url{https://doi.org/10.1007/s11263-015-0816-y}.

\bibitem{sdika2010combining}
{\sc M.~Sdika}, {\em Combining atlas based segmentation and intensity
  classification with nearest neighbor transform and accuracy weighted vote},
  Medical Image Analysis, 14 (2010), pp.~219--226.

\bibitem{siu2020image}
{\sc C.~Y. Siu, H.~L. Chan, and R.~L. Ming~Lui}, {\em Image segmentation with
  partial convexity shape prior using discrete conformality structures}, SIAM
  Journal on Imaging Sciences, 13 (2020), pp.~2105--2139.

\bibitem{rmsprop}
{\sc T.~Tieleman, G.~Hinton, et~al.}, {\em Lecture 6.5-rmsprop: Divide the
  gradient by a running average of its recent magnitude}, COURSERA: Neural
  networks for machine learning, 4 (2012), pp.~26--31.

\bibitem{vercauteren2009diffeomorphic}
{\sc T.~Vercauteren, X.~Pennec, A.~Perchant, and N.~Ayache}, {\em Diffeomorphic
  demons: Efficient non-parametric image registration}, NeuroImage, 45 (2009),
  pp.~S61--S72.

\bibitem{voulodimos2018deep}
{\sc A.~Voulodimos, N.~Doulamis, A.~Doulamis, and E.~Protopapadakis}, {\em Deep
  learning for computer vision: A brief review}, Computational intelligence and
  neuroscience, 2018 (2018).

\bibitem{warfield1999nonlinear}
{\sc S.~Warfield, A.~Robatino, J.~Dengler, F.~Jolesz, and R.~Kikinis}, {\em
  Nonlinear registration and template driven segmentation}, Brain warping, 4
  (1999), pp.~67--84.

\bibitem{wen2015landmark}
{\sc C.~Wen, D.~Wang, L.~Shi, W.~C. Chu, J.~C. Cheng, and L.~M. Lui}, {\em
  Landmark constrained registration of high-genus surfaces applied to
  vestibular system morphometry}, Computerized Medical Imaging and Graphics, 44
  (2015), pp.~1--12.

\bibitem{yezzi2001variational}
{\sc A.~Yezzi, L.~Zollei, and T.~Kapur}, {\em A variational framework for joint
  segmentation and registration}, in Proceedings IEEE Workshop on Mathematical
  Methods in Biomedical Image Analysis (MMBIA 2001), IEEE, 2001, pp.~44--51.

\bibitem{yung2018efficient}
{\sc C.~P. Yung, G.~P. Choi, K.~Chen, and L.~M. Lui}, {\em Efficient
  feature-based image registration by mapping sparsified surfaces}, Journal of
  Visual Communication and Image Representation, 55 (2018), pp.~561--571.

\bibitem{zeng2012computing}
{\sc W.~Zeng, L.~M. Lui, F.~Luo, T.~F.-C. Chan, S.-T. Yau, and D.~X. Gu}, {\em
  Computing quasiconformal maps using an auxiliary metric and discrete
  curvature flow}, Numerische Mathematik, 121 (2012), pp.~671--703.

\bibitem{zeng2014surface}
{\sc W.~Zeng, L.~Ming~Lui, and X.~Gu}, {\em Surface registration by
  optimization in constrained diffeomorphism space}, in Proceedings of the IEEE
  Conference on Computer Vision and Pattern Recognition, 2014, pp.~4169--4176.

\bibitem{DaopingZhang2021Taci}
{\sc D.~Zhang, X.~cheng Tai, and L.~M. Lui}, {\em Topology- and
  convexity-preserving image segmentation based on image registration}, Applied
  Mathematical Modelling, 100 (2021), p.~218.

\bibitem{zhang2021topology_3d}
{\sc D.~Zhang and L.~M. Lui}, {\em Topology-preserving 3d image segmentation
  based on hyperelastic regularization}, Journal of Scientific Computing, 87
  (2021), pp.~1--33.

\bibitem{zhang2019new}
{\sc D.~Zhang, A.~Theljani, and K.~Chen}, {\em On a new diffeomorphic
  multi-modality image registration model and its convergent gauss-newton
  solver}, Journal of Mathematical Research with Applications, 39 (2019),
  pp.~633--656.

\bibitem{zhang2014automatic}
{\sc M.~Zhang, F.~Li, X.~Wang, Z.~Wu, S.-Q. Xin, L.-M. Lui, L.~Shi, D.~Wang,
  and Y.~He}, {\em Automatic registration of vestibular systems with exact
  landmark correspondence}, Graphical models, 76 (2014), pp.~532--541.

\bibitem{zhao2019object}
{\sc Z.-Q. Zhao, P.~Zheng, S.-t. Xu, and X.~Wu}, {\em Object detection with
  deep learning: A review}, IEEE transactions on neural networks and learning
  systems, 30 (2019), pp.~3212--3232.

\end{thebibliography}
\end{document}